\documentclass[preprint,12pt]{elsarticle}

\usepackage{amssymb}
\usepackage{amsmath}
\usepackage{bm}
\usepackage{algorithmicx}
\usepackage{algpseudocode}
\usepackage{algorithm}
\usepackage{subfig}
\usepackage{booktabs}
\usepackage{multirow}
\usepackage{newtxtext}
\usepackage{makecell}
\usepackage{url}
\usepackage[colorlinks=false]{hyperref}
\usepackage{geometry}
\usepackage{color}
\usepackage{mathtools}
\DeclarePairedDelimiter\abs{\lvert}{\rvert}

\DeclarePairedDelimiter\floor{\lfloor}{\rfloor}

\begin{document}

\begin{frontmatter}



\title{Scalable Batch Bayesian Optimization Via Subspace Acquisition Functions}


\author{Dawei Zhan\corref{cor}}
\ead[cor]{zhandawei@swjtu.edu.cn}
\cortext[cor]{Corresponding author}

\author{Zhaoxi Zeng}
\author{Shuoxiao Wei}
\author{Ping Wu} 

\affiliation[label1]{organization={School of Computing and Artificial Intelligence, Southwest Jiaotong University},
            city={Chengdu},
            country={China}}

\begin{abstract}
Extending Bayesian optimization to batch evaluation can enable the designer to make the most use of parallel computing technology. However, most of current batch approaches do not scale well with the batch size.  That is, their optimization efficiencies often deteriorate as the batch size increases.  To address this issue, we propose a simple and efficient approach to extend Bayesian optimization to large-scale batch evaluation in this work. Different from existing batch approaches, the idea of the new approach is to draw a batch of axis-aligned subspaces of the original problem and select one point from each subspace using existing acquisition functions. Numerical experiments show that our proposed approach speedups the convergence significantly when compared with the sequential Bayesian optimization algorithm, and  performs very competitively when compared with ten batch Bayesian optimization algorithms.  The implementation of our proposed approach is available at \url{https://github.com/zhandawei/SubSpace_Acquisition_Functions}.
\end{abstract}




\end{frontmatter}




\section{Introduction}
Bayesian  optimization~\cite{Mockus1994}, also known as Efficient Global Optimization (EGO)~\cite{Jones_1998},  is a powerful tool for  solving expensive black-box optimization problems. Unlike gradient-based algorithms and evolutionary algorithms, Bayesian optimization employs a statistical model to approximate the original objective function and uses an acquisition function to determine where to sample in its search process.  Bayesian optimization has gained lots of success in machine learning~\cite{Snoek_2012}, robot control~\cite{Lizotte_2007}, and engineering design~\cite{Lam_2018}.  The developments of Bayesian optimization have been reviewed in~\cite{Shahriari_2016,Wang_2023,Garnett_2023}.

%

The standard Bayesian optimization algorithm selects and evaluates new  points one by one.  The sequential selection and evaluation process is inefficient when parallel computing resources are available.  It should be noted that the evaluation of the expensive simulations and asynchronous batch evaluations  can also be beneficial from parallel computing techniques. In this work, we focus on synchronous parallel computing.  Studies on asynchronous batch approaches can be found in~\cite{Janusevskis_2012,Alvi_2019,Ath_2021}.  Developing batch Bayesian optimization algorithms, which are able to evaluate a batch of samples simultaneously, is in great needs. The main reason Bayesian optimization operates sequentially is that optimizing the acquisition function can often deliver only one sample point. Therefore, the key to develop batch Bayesian optimization algorithms is to design batch acquisition functions. Currently,  popular acquisition functions used in Bayesian optimization are the Expected Improvement (EI)~\cite{Mockus1994}, the Probability of Improvement (PI)~\cite{Kushner_1964}, the Upper Confidence Bound (UCB)~\cite{Cox_1997}, the Knowledge Gradient (KG)~\cite{Frazier_2008}, Entropy Search (ES)~\cite{Hennig_2012}, Predictive Entropy Search (PES)~\cite{HL_2014}, Max-value Entropy Search (MES)~\cite{WangZi_2017}, Thompson Sampling (TS)~\cite{Kandasamy_2018} and so on.   There are also works focusing on the optimization methods of the acquisition functions~\cite{Wilson_2017,Grosnit_2021,Gramacy_2022,Wycoff_2026}. Among different acquisition functions, the EI criterion is arguably the most widely used. Therefore, we mainly focus on the parallel extensions of the EI criterion  in this work. We also extend our proposed approach to the PI and UCB acquisition functions in Section~\ref{section_experiment}.

The most direct extension of the EI function for batch evaluation is the multi-point expected improvement, which is often called $q$EI where $q$ is the number of batch points~\cite{Schonlau_1997}. The $q$EI criterion measures the expected improvement when a batch of samples are considered. Although the $q$EI  has a solid theoretical foundation, its exact computation is time-consuming when the batch size $q$ is large.  Works have been done to accelerate the calculation of $q$EI~\cite{Ginsbourger_2010,Chevalier_2013,Marmin_2015,Wang_2020}. However, current approaches for computing $q$EI are only applicable when the batch size is smaller than ten. Although  one can use the Fast Multi-Point Expected Improvement (F$q$EI)~\cite{Zhan_2023}  to reduce the computational cost or use the $q$LogEI~\cite{Ament_2023} to reduce the numerical optimization difficulty,  optimizing the multi-point acquisition function is still challenging when the batch size is large. The reason is that the dimension of the $q$EI criterion is  $d\times q$, where $d$ is the dimension of the  optimization problem and $q$ is the batch size. As the batch size $q$ increases,  the volume of the search space of the acquisition function increases exponentially~\cite{Ament_2023,Zhan_2023}.  As a result, the $q$EI and its variants are often not applicable for large batch size.

To avoid the curse of dimensionality of the $q$EI criterion, several heuristic batch approaches have been proposed.  The Kriging believer (KB) and Constant Liar (CL) approaches~\cite{Ginsbourger_2010} get the first query point using the standard EI criterion, assign the objective function value of the acquisition point using fake values, and then use the fake point to update the Gaussian process model to get the second query point. This process is repeated until a batch of $q$ query points are obtained. The KB approach uses the Gaussian process prediction as the fake value while the CL approach uses  the minimum, the maximum, or the mean value of the sample points as the fake value~\cite{Ginsbourger_2010}. The KB and CL approaches are able to locate a batch of points within one iteration without evaluating their objective values. However, as the batch size increases, the accuracy of the Gaussian process model decreases gradually as more and more fake points are included. This may mislead the search direction and further affect the optimization efficiency. The shortcomings of the KB approach have also been discussed and addressed in~\cite{Alvi_2019,Ath_2021}.

It has been observed that the EI function decreases dramatically around the updating point but slightly at places far away from the updating point after the Gaussian process model is updated using a new point~\cite{Gonzalez_2016}. Therefore, we can design an artificial function to simulate the effect that the updating point bring to the EI function and use it to select a batch of points that are similar to the points the sequential EI selects. This idea has been used in the Local Penalization (LP)~\cite{Gonzalez_2016}, the Expected Improvement and Mutual Information (EIMI)~\cite{LiZheng_2016}, and the Pseudo Expected Improvement (PEI)~\cite{Zhan2017}. The LP approach uses one minus the probability that the studying point belongs to the balls of already selected points as the artificial function~\cite{Gonzalez_2016}, the EIMI approach uses the mutual information between the studying point and the already selected points as the artificial function~\cite{LiZheng_2016}, and the PEI criterion uses one minus the correlation between the studying point and the already selected points as the artificial function~\cite{Zhan2017}. Through sequentially optimizing these modified EI functions, these approaches can select a batch of samples one by one, and then evaluate them in parallel.  However, due to the use of artificial functions, the simulated EI functions will be less and less accurate, and as a result the qualities of the selected samples will also be lower and lower as the batch size increases.

There are also approaches try to select multiple query points through multiple surrogate models. Besides Gaussian process models, the  multiple surrogate efficient global optimization  algorithm~\cite{Viana_2013} also considers the radial basis function models, neural network models and support vector machine models.  In comparison, the Multi-Scale Multi-Recommendations (MSMR) approach~\cite{Joy_2020} only considers Gaussian process models, and trains different Gaussian process models using different length-scales in the kernel functions.   However, this kind of approaches are difficult to scale as the computational cost for training a large number surrogate models is often  high.

Another idea for delivering a batch of samples within one cycle is to use multi-objective optimization method~\cite{Bischl_2014,Gupta_2018,Binois_2025,Carciaghi_2025}. These approaches consider multiple acquisition functions at the same time, and transform the infill selection problem into a multi-objective optimization problem. By solving the multi-objective infill selection problem, a set of Pareto optimal points can be obtained and a batch of samples can be then selected from the Pareto optimal points. The multi-objective optimization based efficient global optimization~\cite{Feng_2015} treats the two parts of the EI function as two objectives and solves the two-objective optimization problem using the multi-objective evolutionary algorithm based on decomposition~\cite{Zhang_2007}. In the Multi-objective ACquisition Ensemble (MACE)  approach~\cite{Lyu_2018}, the EI, PI and LCB are chosen as the three objectives and the multi-objective optimization based on differential evolution~\cite{Robic_2005} is utilized to solve the three-objective optimization problem. The adaptive batch acquisition functions via multi-objective optimization approach~\cite{Chen_2023}  considers multiple acquisition functions but selects only two of them to solve based on an objective reduction method. The Pareto optimization for exploitation and exploration approach~\cite{Jiang_2025} uses the Gaussian process mean and variance as two objectives and select the query points one by one through updating the Gaussian process variance function after one query point is selected.  Recently, a non-dominated sorting memetic algorithm which combines NSGA-II and gradient descent algorithm is introduced to improve the solution quantity of the bi-objective infill selection problem~\cite{Carciaghi_2025}. 
For the multi-objective optimization approaches, the number of batch samples is limited to the population size of the adopted multi-objective evolutionary algorithms. Also, identifying the most promising points for expensive evaluations from obtained Pareto front is a challenging task. 

Besides batch extensions of the expected improvement function, other batch approaches such as the batch  lower confidence bound~\cite{Desautels_2014}, multi-point knowledge gradient~\cite{Wu_2016}, parallel predictive entropy search~\cite{Shah_2015}, Thompson sampling~\cite{Kandasamy_2018,Eriksson_2019} and determinantal point processes~\cite{Nava_2022,Nezami_2025} have also been frequently studied.

As the rapid developments of  high-performance computing and cloud computing, distributing the expensive evaluations on large-scale computing clusters can further reduce the wall-clock time and support fast decision making. Therefore, designing batch criteria for large batch size to make the most use of  the  computing resources and further accelerate the optimization process of Bayesian optimization is a meaningful task. In this work, we propose a new approach to generate a large number of query points to fill this research gap. The idea is to locate a batch of samples in multiple different axis-aligned subspaces. The proposed approach is simple yet efficient, and it can be applied to a wide range of acquisition functions. The results show that our proposed approach has significantly better performance over current batch approaches.

The reminder of this paper is organized as following. Section~\ref{section_background} introduces the basic background knowledge about Bayesian optimization. Section~\ref{section_proposed} introduces our proposed subspace acquisition functions in detail. Numerical experiments are given in Section~\ref{section_experiment}. Finally, conclusions about our work are given in Section~\ref{section_conclusion}.

\section{Backgrounds}
In this work, we try to solve an expensive, single-objective, bound-constrained, and black-box optimization problem:

\begin{equation}
	\bm{x}^{\star} \in {\underset{\bm{x} \in\mathcal{X} }{\arg\min}} f(\bm{x})
\end{equation}
where $\mathcal{X} = \{ \bm{x} \in \mathbb{R}^d: a_i \le x_i \le b_i,~ i= 1,2,\cdots,d\} $ is the design space, and  $a_i$ and $b_i$ are the lower bound and upper bound of the $i$th coordinate. The objective function $f$ is assumed to be black-box, which often prohibits the use of gradient-based methods. The objective function $f$ is also assumed to be expensive to evaluate, which often prohibits the direct use of evolutionary algorithms. This kind of optimization problems widely exist in machine learning and engineering design~\cite{Snoek_2012,Lizotte_2007,Lam_2018}. Currently,  Bayesian optimization algorithms are the mainstream approaches for solving these expensive optimization problems~\cite{Wang_2023}. In the following, we give a brief introduction about Bayesian optimization. More details about Bayesian optimization can be found in the book~\cite{Garnett_2023}.

\label{section_background}
\subsection{Gaussian Process Model}
Gaussian process models~\cite{Rasmussen_2006} are the most frequently used statistical models in Bayesian optimization. They are employed to approximate the expensive objective function based on the observed samples. The Gaussian process model treats the unknown objective function as a realization of a Gaussian process~\cite{Rasmussen_2006}
\begin{equation}
	f(\bm{x}) \sim N\left(m(\bm{x}), \kappa(\bm{x},\bm{x}')\right)
\end{equation}
where $m$ is the mean function of the random process and $\kappa$ is the covariance function, also known as the kernel function, between any two points $\bm{x}$ and $\bm{x}'$.  Under the prior distribution, the objective values of every combination of points follow a multivariate normal distribution~\cite{Rasmussen_2006}. Commonly used mean functions of the Gaussian process include constant values and polynomial functions. Commonly used covariance functions include the squared exponential function and the Mat\'{e}rn function~\cite{Rasmussen_2006}.

Assume we have gotten a set of $n$ samples 
\begin{equation*}
	\bm{X} = [\bm{x}^{(1)},\bm{x}^{(2)},\cdots,\bm{x}^{(n)}]
\end{equation*}
and their objective values
\begin{equation*}
	f(\bm{X}) = [f(\bm{x}^{(1)}),f(\bm{x}^{(2)}),\cdots,f(\bm{x}^{(n)})].
\end{equation*}
Conditioned on the observed points, the posterior probability distribution of an unknown point $\bm{x}$ can be computed using Bayes' rule~\cite{Rasmussen_2006}
\begin{equation}
	f(\bm{x}) \mid f(\bm{X}) \sim N\left(\mu(\bm{x}),\sigma^2(\bm{x})\right)
\end{equation}
where 
\begin{equation}
	\label{eq_mean}
	\mu(\bm{x}) = \kappa(\bm{x},\bm{X}) \kappa(\bm{X},\bm{X})^{-1} \left(f(\bm{X}) - m(\bm{X})\right) + m(\bm{x}) \\
\end{equation}
and 
\begin{equation}
	\label{eq_variance}
	\sigma^2(\bm{x})  = \kappa(\bm{x},\bm{x}) - \kappa(\bm{x},\bm{X})\kappa(\bm{X},\bm{X})^{-1}\kappa(\bm{X},\bm{x}).
\end{equation}
The hyperparameters in the prior distribution can be estimated by maximizing the likelihood function of the $n$ observed samples. More details about Gaussian process models can be found in the book~\cite{Rasmussen_2006}.

\begin{figure}
	\centering
	\subfloat[GP approximation]{\includegraphics[width=0.45\linewidth]{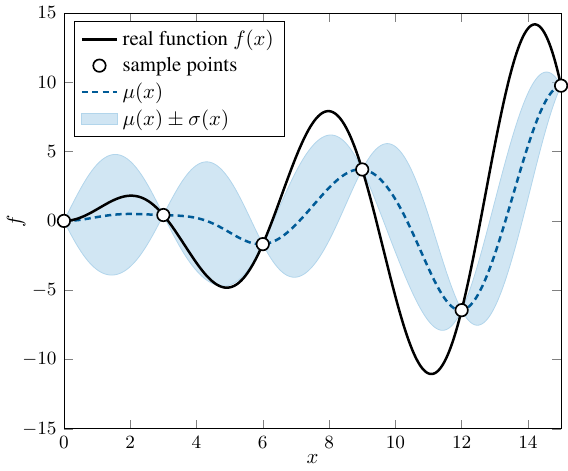}} \hfil
	\subfloat[EI function]{\includegraphics[width=0.45\linewidth]{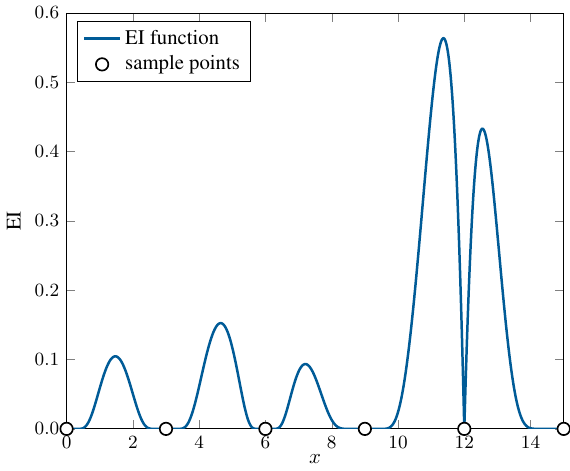}} \hfil
	\caption{Gaussian process approximation and the corresponding EI function of $f(x) = x\sin(x)$.  }
	\label{fig_GP}
\end{figure}

Commonly used kernel functions such as the squared exponential function and the Mat\'{e}rn function are stationary, which means the kernel functions depend only on $\bm{x} -\bm{x}'$ and are invariant to translations in input space.  The GP model with stationary kernels often assumes the function to be modeled is stationary, and a better way to model a non-stationary function is to use non-stationary kernels, see for example~\cite{Snoek_2014,Heinonen_2016,Garnett_2023}.  We consider a one-dimensional non-stationary function $f(x) = x \sin (x)$ in Fig.~\ref{fig_GP}(a). We call this function non-stationary as its variance grows as $x$ increases.  Here, we use GP with squared exponential kernel to approximate this function using six uniformly distributed samples in.  We can see from the figure that the objective value of any point $x$ can be interpreted as a random value with mean $\mu(x)$ and variance $\sigma^2(x)$. The variance vanishes at already observed points and rises up at places far away from the observed points.

\subsection{Expected Improvement}
The second major component of Bayesian optimization is the acquisition functions. After training the Gaussian process model, the following step  is to decide where to sample next in order to locate the global optimum of the original function as quickly as possible.  This is done by maximizing a well-designed acquisition function in Bayesian optimization. To take a balance between global search and local search, the acquisition function should on one hand select points around current best observation and on the other hand select points around sparsely sampled areas. Among different kinds of acquisition functions, the expected improvement (EI) function is arguably the most widely used because of its mathematical tractability and good performance~\cite{Frazier_2018,Zhan_2020}. Therefore, we focus on the EI acquisition function in this work, and discuss the application of our proposed approach to PI and UCB in Section~\ref{section_SSPI}.

Assume the current minimum objective value among $n$ evaluated samples is $ f_{\min}$ and the corresponding best solution is  $\bm{x}_{\min}$. According to the Gaussian process model, the objective value of an unknown point $\bm{x}$ can be interpreted as a random value following the normal distribution~\cite{Jones_1998}
\begin{equation}
	\label{eq_norm}
	F(\bm{x}) \sim N\left(\mu(\bm{x}),\sigma^2(\bm{x})\right)
\end{equation}
where $\mu(\bm{x})$ and $\sigma^2(\bm{x})$ is the Gaussian process mean and variance in (\ref{eq_mean}) and (\ref{eq_variance}) respectively. The EI function is the expectation of the improvement by comparing the random value $F(\bm{x})$ with $f_{\min}$~\cite{Jones_1998}
\begin{equation}
	\text{EI}(\bm{x}) =  \mathbb{E}[\max(f_{\min}-F(\bm{x}), 0)]
\end{equation}
where $f_{\min}$ is current minimum objective value among all evaluated samples. Integrating this equation with the normal distribution in (\ref{eq_norm}), we can get the closed-form expression~\cite{Jones_1998}
\begin{equation}
	\label{eq_EI}
	\text{EI}(\bm{x}) = \left(f_{\min} - \mu(\bm{x}) \right)\Phi\left(\frac{f_{\min} - \mu(\bm{x}) }{\sigma(\bm{x})}\right)  + \sigma(\bm{x})\phi\left(\frac{f_{\min} - \mu(\bm{x}) }{\sigma(\bm{x})}\right)
\end{equation}
where $\Phi$ and $\phi$ are the standard normal cumulative and density functions respectively. As can be seen from the formula, the EI function is a nonlinear combination of the Gaussian process mean and variance.

The EI function of the one-dimensional example demonstrated in Fig.~\ref{fig_GP}(a) is shown in Fig.~\ref{fig_GP}(b). We can see that the EI function is zero at sampled points because sampling at these already evaluated points brings no improvement.  We can also see that the EI function is high at places where the Gaussian process mean is low and the variance is high.

\subsection{Multi-Point Expected Improvement}
The Multi-Point Expected Improvement~\cite{Ginsbourger_2010}, or $q$EI, is the batch extension of the single-point EI function. Instead of measuring the potential improvement of one single point, the $q$EI measures the potential improvement of  $q$ points $[\bm{x}_1,\bm{x}_2,\cdots,\bm{x}_q]$~\cite{Ginsbourger_2010}
\begin{equation}
	\label{eq_qEI}
	q\text{EI}\left([\bm{x}_1,\bm{x}_2,\cdots,\bm{x}_q]\right) =  \mathbb{E}\left[\max\left(f_{\min}- \min_{i=1}^q F(\bm{x}_i), 0\right)\right]
\end{equation}
where $F(\bm{x}_i)$ is the random objective value of the $i$th query points.  Although each random objective value $F(\bm{x}_i)$ follows a Gaussian distribution, the  minimum of them $ \min_{i=1}^q F(\bm{x}_i)$ no longer follows a Gaussian distribution. As a result, directing integrating this equation gives no closed-form expression~\cite{Ginsbourger_2010}.   The fast computation method of $q$EI has been developed in~\cite{Ginsbourger_2010,Chevalier_2013,Marmin_2015}. The formula can be applied to any $q$, but the computation is often restricted to $q=10$ due to the high computational cost of evaluating the multivariate CDF and PDF functions. When the batch size is greater than ten, the  $q$EI is often computed through Monte Carlo integration.  It can also be seen from (\ref{eq_qEI}) that the dimension of $q$EI is $d\times q$, which is $q$ times larger than the single-point EI function.  Therefore, it is also challenging to optimize the $q$EI function when the dimension of the problem $d$ or the batch size $q$ is large.

\begin{figure}
	\centering
	\includegraphics[width=0.6\linewidth]{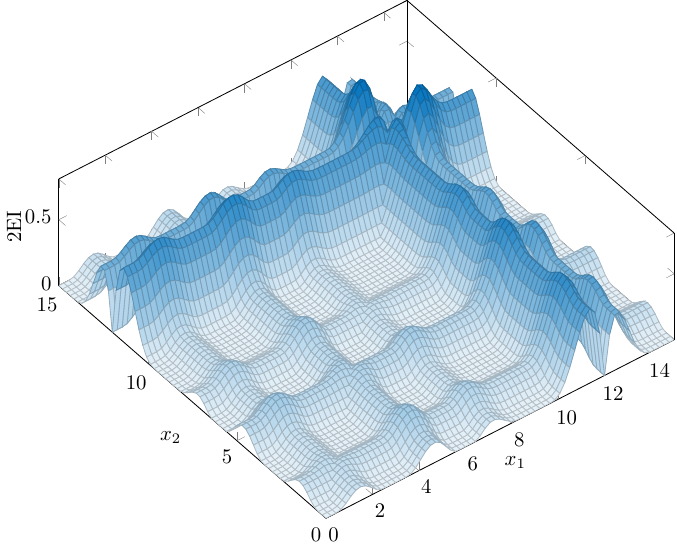}
	\caption{The landscape of the $q$EI function of  $f(x) = x\sin(x)$ when $q=2$. }
	\label{fig_qEI}
\end{figure}

The $2$EI function of the above one-dimensional example is illustrated in Fig.~\ref{fig_qEI}, where the $x_1$-axis and $x_2$-axis present the two query points and the $z$-axis is the corresponding $2$EI value. We can see that the landscape of the $2$EI function is symmetric along its diagonal, which means the $2$EI function is invariant to  the order of the query points. 

\subsection{Related Works}
Our work is related to the subspace embedding approach explored in high-dimensional BO~\cite{Wang_2016,Nayebi_2019,Wang_2026}. Both the random subspace embedding approach and our subspace acquisition approach select query points in low-dimensional space. But there also exists some key differences between them.  The subspaces employed in  REMBO~\cite{Wang_2016} and HeSBO~\cite{Nayebi_2019}  are self-defined and fixed subspaces while the subspaces in our algorithm are randomly selected axis-aligned subspaces.  For expensive evaluation,  the random embedding approach uses randomly generated matrices to map the points from the lower-dimensional space to the original high-dimensional space. However, the point mapped back to the high-dimensional space may fall out of the boundaries of the design space. While our approach simply uses the coordinates of the best solution to fill the un-selected variables, therefore the query point in the high-dimensional space will never fall out of boundaries.   The second difference is that the aim of  the random embedding approach is to extend BO to high-dimensional space while the major goal of our approach is to select a batch of different query points for parallel evaluations.  

Another related work to ours is the ECI (Expected Coordinate Improvement)~\cite{Zhan_2024}, which perturbs one coordinate at a time to generate query points. The difference between our approach and the ECI approach is that our approach perturbs  multiple coordinates to generate one candidate point while the ECI perturbs only one coordinate~\cite{Zhan_2024}. Another important difference between our approach and ECI is that we search in multiple different subspaces to generate a batch of query points while ECI searches in one coordinate at a time to generate one query point within one iteration. The strategies for selecting the subspaces or coordinates are also different. Our approach uses random selection to select a batch of different subspaces while the ECI approach selects the coordinate to optimize based on the ECI values of the coordinates. Finally, the aim of our work is to extend BO to batch evaluation while the aim of ECI~\cite{Zhan_2024} is to extend BO to high-dimensional space.

\section{Proposed Subspace Approach}
\label{section_proposed}

\subsection{SubSpace Acquisition Functions}
We propose the subspace versions of existing acquisition functions, which measure the improvements of a candidate point in specific axis-aligned subspaces.  This idea can be used to a wide range of acquisition functions. In the following, we illustrate this idea using the EI acquisition function, and discuss the subspace versions of the PI and LCB acquisition functions in Section~\ref{section_SSPI}.

Consider the original $d$-dimensional design space
\begin{equation*}
	\mathcal{X} = \{ [x_1,x_2,\cdots,x_d]\} \subseteq \mathbb{R}^d,
\end{equation*}
then a $k$-dimensional  ($1 \le k \le d$) axis-aligned subspace can be expressed as 
\begin{equation*}
	\mathcal{S} = \{[s_1,s_2,\cdots,s_k]\} \subseteq \mathbb{R}^k
\end{equation*}
where $s_i \in \{x_1,x_2,\cdots,x_d\}$ for $i=1,2,\cdots,k$ and  $s_1 \ne s_2 \ne \cdots, \ne s_k$.  Assume the current best solution is  $\bm{x}_{\min} = [x_{\min,1},x_{\min,2},\cdots,x_{\min,d}]$. The standard EI function measures the potential improvement a point can get  below the current best solution $\bm{x}_{\min}$. From another point of view, the EI function can be seen as the measurement of the potential improvement as we move the current best solution $\bm{x}_{\min}$ in the original design space. Following this, we define the SubSpace Expected Improvement (SSEI) as the amount of expected improvement as we move the current best solution $\bm{x}_{\min}$ in an axis-aligned subspace 

\begin{equation}
	\label{eq_ESSI}
	\text{SSEI}(\bm{s})  = \left(f_{\min} - \mu(\bm{z}) \right)\Phi\left(\frac{f_{\min} - \mu(\bm{z}) }{\sigma(\bm{z})}\right) + \sigma(\bm{z})\phi\left(\frac{f_{\min} - \mu(\bm{z}) }{\sigma(\bm{z})}\right)
\end{equation}
where $\bm{s} = [s_{1},s_{2},\cdots,s_{k}]$ are the coordinates to be optimized, and  
\begin{equation*}
	\bm{z} = [x_{\min,1},\cdots,s_{1},\cdots,x_{\min,i}\cdots,s_{k},\cdots,x_{\min,d}]
\end{equation*}
is the $d$-dimensional input of the Gaussian process model, in which the $s_1,s_2,\cdots,s_k$ are the coordinates to be optimized and the remaining coordinates are fixed at the values of current best solution $\bm{x}_{\min} = [x_{\min,1},x_{\min,2},\cdots,x_{\min,d}]$. It should be noted that, the SSEI is not a new acquisition function, it is simply the subspace version of the standard EI function.

\begin{figure}[h!]
	\centering
	\includegraphics[width=1.0\linewidth]{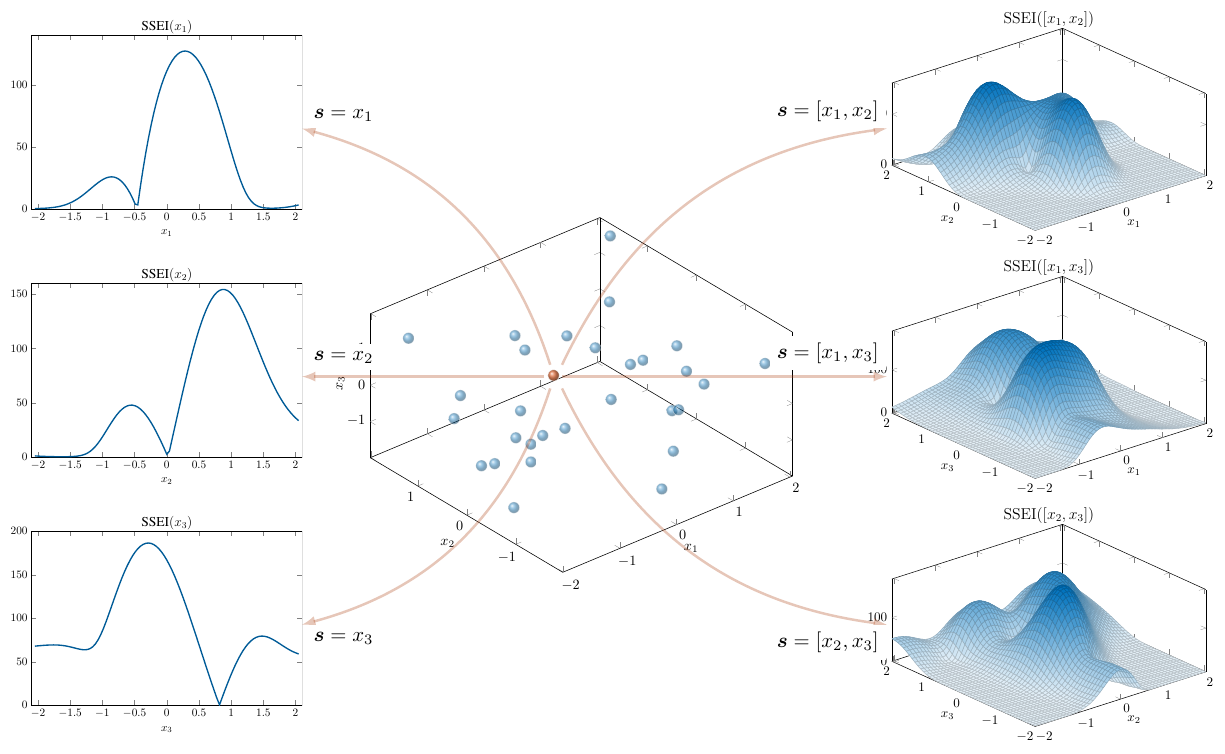}
	\caption{Six SSEI functions of a 3-dimensional problem. The figure in the middle shows the scatter plots of the thirty samples in the original 3-dimensional space.  The red filled point is the current best solution among the thirty samples. On the left, we show the SSEI functions as we move the current best solution along subspaces $\bm{s} = x_1$, $\bm{s} = x_2$ and $\bm{s} = x_3$, respectively. On the right, we show the SSEI functions as we move the current best solution in subspaces $\bm{s} = [x_1,x_2]$, $\bm{s} = [x_1,x_3]$ and $\bm{s} = [x_2,x_3]$, respectively.  }
	\label{fig_SSEI}
\end{figure}

The SSEI functions of the 3-dimensional Rosenbrock problem are demonstrated in Fig.~\ref{fig_SSEI}. We draw thirty random points and train a Gaussian process model using these points. The left figures are the SSEI functions in three $1$-dimensional subspaces while the right figures are the SSEI functions in three $2$-dimensional subspaces.  From another point of view, the left and right figures can also be treated as the $1$-dimensional and $2$-dimensional slices of the original expected improvement function.    After optimizing these SSEI functions, we can then get a batch of six query points for parallel evaluations.

The dimension of the subspaces $k$ can vary from $1$ to $d$. When $k=1$, the available axis-aligned subspaces are $\{[x_1], [x_2],\cdots, [x_d]\}$. Similarly, as $k=2$, the available axis-aligned subspaces are $\{[x_1,x_2], [x_1,x_3],\cdots,[x_{d-1},x_d]\}$. This goes on until $k$ reaches to $d$, in which case there is only one available axis-aligned subspace $\{[x_1,x_2,\cdots,x_d]\}$. The total number of subspaces of a $d$-dimensional space can be calculated as

\begin{equation}
	N  = \binom{d}{1} + \binom{d }{2}+ \cdots +\binom{d }{d}  = 2^d - 1
\end{equation}
where $\binom{d}{i}$ donates the number of combinations of selecting $i$ different variables from $d$ total variables.

When the required batch size is smaller than or equal to the number of subspaces $q \ge N$, we  simply select one query point within one subspace. However, when the required batch size is larger than the number of subspaces  $q < N $, which might happen for low-dimensional problems,  the available subspaces will be insufficient for selecting the $q$ query points if we still select one query point in one subspace. To tackle this issue, we propose the Subspace Multi-Point Expected Improvement (SS$q$EI) for selecting multiple query points within one subspace.

Assume the subspace is $\bm{s} = [s_1,s_2,\cdots,s_k]$, and the number of query points we need to select from this subspace is $q$, we define the SS$q$EI as the overall potential improvement as we move the $q$ points from the current best solution along the axis-aligned subspace
\begin{equation}
	\label{eq_SSqEI}
	\text{SS}q\text{EI}\left([\bm{s}_1,\bm{s}_2,\cdots,\bm{s}_q]\right) =  \mathbb{E}\left[\max\left(f_{\min}- \min_{i=1}^q F(\bm{z}_i), 0\right)\right]
\end{equation}
where 
\begin{equation*}
	\bm{z}_i= [x_{\min,1},\cdots,s_{i,1},\cdots,s_{i,k},\cdots,x_{\min,d}]
\end{equation*}is the $i$th candidate point, whose coordinates $s_{i,1}, s_{i,2},\cdots,s_{i,k}$ are variables in the subspace that needs to be optimized and the remaining coordinates are fixed at the current best solution.

\subsection{Computational Framework}

Based on the proposed SSEI and SS$q$EI acquisition functions, we introduce a novel batch Bayesian optimization algorithm. In each iteration, we locate a batch of $q$ points in $q$ different subspaces using the SSEI acquisition function if the batch size is smaller than or equal to the number of subspaces $q\le  N $. Otherwise, we locate the batch of $q$ points within the $N $ subspaces using the SS$q$EI acquisition function. After getting the $q$ query points,  we then evaluate them in parallel.  The computational framework of the proposed approach is given in Algorithm~\ref{algorithm_ESSI}.

\begin{algorithm}
	\caption{Computational Framework of the Proposed Batch Bayesian optimization}
	\label{algorithm_ESSI}
	\begin{algorithmic}[1]	
		\Require{$d=$ dimension of the problem, $q = $ batch size, $n_{\text{init}} = $ number of initial samples,
			$n_{\max} = $ maximum number of  objective  evaluations.}
		\Ensure best found solution $(\bm{x}_{\min},f_{\min})$.		
		\State  Randomly generate $n_\text{init}$ samples  $\mathcal{D} = (\bm{X},f(\bm{X}))$, set $n = n_\text{init}$, and $N = 2^d-1$.
		\While{$n < n_{\max}$} 
		\State Train a GP model using the current data set $\mathcal{D}$. \Comment{Train a GP model}
		\State $\mathcal{S} = \varnothing$  \Comment{Subspace selection}
		\While{$\abs{\mathcal{S} } < \min(q,N)$} 
		\State Draw a random integer $k$ from $1$ to $d$. \label{step_random_integer} 
		\State Draw a random $k$-dimensional axis-aligned subspace $\bm{s}$. \label{step_random_subspace} 
		\If{$\bm{s} \not\in  \mathcal{S}$} \label{step_reject_sampling_start} 
		\State	$\mathcal{S} = \{\mathcal{S},\bm{s}\}$.
		\EndIf  \label{step_reject_sampling_end} 
		\EndWhile 
		\If{$q \le N $ }  \Comment{Acquisition  optimization}  
		\For{$i = 1$ to $q$}  \Comment{Select one point within one subspace}
		\State \label{step_SSEI_optimization}
		\begin{equation*}
			\bm{p}^{(i)} \in  {\underset {\bm{s}^{(i)} \in \mathcal{S}^{(i)}} {\arg\max}} ~\text{SSEI}(\bm{s}^{(i)}) 
		\end{equation*}
		\State  Get $\bm{x}^{(n+i)}$ by replacing coordinates in $\bm{x}_{\min}$ by $\bm{p}^{(i)}$. \label{step_SSEI_replacement}
		\EndFor
		\Else 
		\State Distribute $q$ points to $N $ subspaces as equally as possible $q_1,q_2,...q_{N}$ where $\sum_{i=1}^{N} q_i = q$. 
		\For{$i = 1$ to $N$}  \Comment{Select multiple points within one subspace}
		\State
		\begin{equation*}
			[\bm{p}_{1}^{(i)},\bm{p}_{2}^{(i)},\cdots,\bm{p}_{q_i}^{(i)}] \in  {\underset {\bm{s}_{j}^{(i)} \in \mathcal{S}^{(i)},j=1,\cdots,q_i} {\arg\max}} ~	\text{SS}q\text{EI}\left([\bm{s}_{1}^{(i)},\bm{s}_{2}^{(i)},\cdots,\bm{s}_{q_i}^{(i)}]\right)
		\end{equation*}
		\State  Get $q_i$ query points by replacing coordinates in $\bm{x}_{\min}$ by $\bm{p}_{j}^{(i)}, j=1,2\cdots,q_i$.
		\EndFor
		\EndIf
		\State   Evaluate the $q$ query points in parallel to get  $f(\bm{x}^{(n+1)} ) ,\cdots,f(\bm{x}^{(n+q)} )$. \Comment{ Evaluation}
		\State Update data set $\mathcal{D} = \{\mathcal{D}, (\bm{x}^{(n+1)}, f(\bm{x}^{(n+1)} )),\cdots,(\bm{x}^{(n+q)}, f(\bm{x}^{(n+q)} ))\}$,  and set $n = n+q$.
		\EndWhile
	\end{algorithmic}	
\end{algorithm}

The subspace selection and acquisition optimization are the two key steps for our proposed batch Bayesian optimization algorithm, therefore are further explained in the following in detail.

In subspace selection, we use the random approach to select axis-aligned subspaces. This random selection strategy avoids setting  the  dimension of the subspace $k$ and turns out to be very efficient.   Instead of directly drawing $q$ subspaces from all $N$ axis-aligned subspaces, we select the $q$ subspaces by first drawing a random integer $k$ and then drawing a random combination $\bm{s}$, as shown in Step~\ref{step_random_integer} and Step~\ref{step_random_subspace}.  The reason is that the direct selection strategy is very computationally expensive to list all $N$ subspaces when $d$ is large.  For example, when $d=20$, the number of total subspaces would be $N = 2^{20}-1 = 1,048,575$.  It will cost lots of time and memory to list all the subspaces in these cases. Our strategy is very fast but might select identical subspaces during the $q$ selections. We eliminate this by checking whether the candidate subspace has been selected after each selection. If the candidate subspace has been selected, then we can reject the selection and start a new drawing, as shown in Step~\ref{step_reject_sampling_start} to Step~\ref{step_reject_sampling_end}.

\begin{figure}
	\centering
	\includegraphics[width=0.6\linewidth]{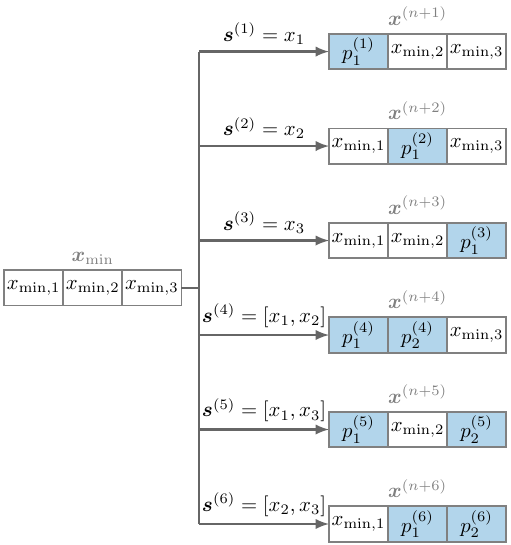}
	\caption{SSEI acquisition optimization results on the $3$-dimensional problem. The current best solution is $\bm{x}_{\min} = [x_{\min,1},x_{\min,2},x_{\min,3}]$, and the six subspaces we select are $\bm{s}^{(1)} = x_1$, $\bm{s}^{(2)} = x_2$, $\bm{s}^{(3)} = x_3$, $\bm{s}^{(4)} = [x_1,x_2]$, $\bm{s}^{(5)} = [x_1,x_3]$ and $\bm{s}^{(6)} = [x_2,x_3]$. After the optimization, we can get six query points.}
	\label{fig_Optimization_SSEI}
\end{figure}

In the acquisition optimization process, we use the SSEI to select $q$ query points in $q$ different subspaces when $q \le N$, and use the SS$q$EI to locate $q$ points within $N$ subspaces when $q > N$. In the first situation,  we optimize the variables of $i$th subspace while holding the other variables fixed at the current best solution $\bm{x}_{\min}$, as shown in Step~\ref{step_SSEI_optimization}. After the optimization, we replace the variables in $\bm{x}_{\min}$ by the optimized ones to get the $i$th query point, as shown in Step~\ref{step_SSEI_replacement}. We demonstrate the acquisition optimization process using SSEI on the $3$-dimensional example in Fig.~\ref{fig_Optimization_SSEI}.  In this example, we want to select $q=6$ query points. Since the total number of subspaces $N = 2^3-1 =7$ is greater than  $q=6$, we can select one query point within one subspace.

\begin{figure}
	\centering
	\includegraphics[width=0.55\linewidth]{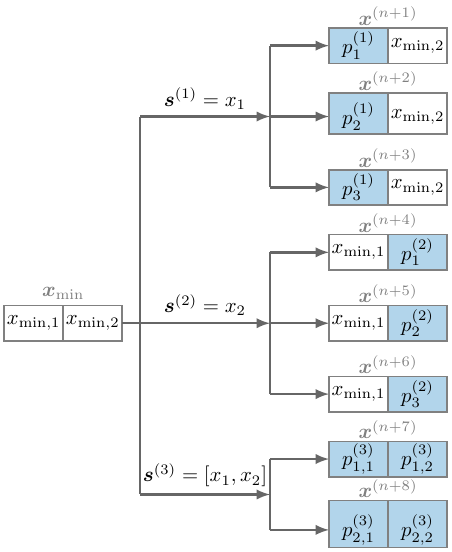}
	\caption{SS$q$EI acquisition optimization results on a $2$-dimensional problem. The current best solution is $\bm{x}_{\min} = [x_{\min,1},x_{\min,2}]$, and the three subspaces  are $\bm{s}^{(1)} = x_1$, $\bm{s}^{(2)} = x_2$, $\bm{s}^{(3)} = [x_1,x_2]$.  We select $q_1=3$ points in the first subspace, $q_2 = 3$ points in the second subspace and $q_3=2$ points in the last subspace using the SS$q$EI acquisition functions. After the optimization, we can get eight query points.}
	\label{fig_Optimization_SSqEI}
\end{figure}

In the second situation where the required number of points $q$ is greater than the number of subspaces, we have to select multiple points within one subspace using the SS$q$EI acquisition function. First, we need to divide the $q$ query points into $N$ subspaces as equally as possible.  If $q$ is a multiple of $N$, we simply select $q/N$ points in each subspace. Otherwise, we first distribute $\floor{q/N}$ points into each subspace, then assign one additional point to the first $(q \bmod{N} )$ subspaces. For example, if the dimension of the problem $d=2$, the total number of subspaces would be $N=2^2-1 = 3$. If we want to query $q=8$ points, we can  distribute the $8$ points into $3$ subspaces using above approach and get $q_1=3$, $q_2 = 3$ and $q_3 = 2$ for each subspace.  The acquisition optimization process of this example is illustrated in Fig.~\ref{fig_Optimization_SSqEI}. In this figure, we select $3$ query points in the first subspace, $3$ query points in the second subspace and $2$ query points in the last subspace using the SS$q$EI functions. After the acquisition optimizations, we replace the coordinates of the current best solutions by the corresponding optimized variables to get the full query points. 

Since the dimension of the subspace $k$ is randomly drawn from 1 to $d$
\begin{equation*}
	K \sim U(1,d)
\end{equation*}
the expectation of $k$ is then $\mathbb{E}(K) = \frac{1+d}{2}$. Therefore, on average the dimension of our SSEI and SS$q$EI functions is near half of the dimension of their counterparts. This also makes our subspace acquisition functions easier to solve compared with the original acquisition functions.

\section{Numerical Experiments}

\label{section_experiment}
\subsection{Experiment Settings}
We select the GPMean, SumSquares, Rosenbrock, DixonPrice, Ackley, Rastrigin, Griewank, Levy, Michalewicz and Schwefel problems to test the compared algorithms. These ten synthetic test problems are widely used benchmark problems in Bayesian optimization.  The first GPMean function is sampled from a Gaussian process. It can eliminate model misspecification and allow a direct comparison of different acquisition functions~\cite{Hennig_2012,Hernandez_2014}. The dimensions of  the test problems are set to be $d= 2, 4, 8, 10, 20$ and $30$ to test the compared algorithms' performances in both low-dimensional and high-dimensional spaces. A total of $60$ test problems are used in the experiments. The formulas and properties of the selected test problems are introduced in Appendix~\ref{section_test_problems} in detail.

The Latin hypercube sampling is used to generate the initial samples for the compared algorithms. The number of initial samples is set to be $n_{\text{init}} = 2d$ where $d$ is the dimension of the problem.  The number of acquisition samples   is set to be $256$ for $2$-dimensional and $4$-dimensional problems,  $512$ for $8$-dimensional and $10$-dimensional problems, and $1024$ for $20$-dimensional and $30$-dimensional problems.

We implement our proposed SSEI and SS$q$EI approaches in BoTorch~\cite{Balandat_2020}.  We use the constant mean  and the square exponential kernel  for modeling the Gaussian process models. We use BoTorch's default L-BFGS-B optimizer with $20$ restarts and $1024$ initial samples to optimize different acquisition functions. All experiments are run $30$ times to deliver reliable results. The BoTorch implementation of our proposed algorithm is available at \url{https://github.com/zhandawei/SubSpace_Acquisition_Functions}.

\subsection{Comparing SSEI with EI when $q=1$ }

The first problem we want to investigate is how well our proposed subspace approach performs compared with the sequential EI approach when only one query point is selected in each cycle. Both the SSEI and standard EI utilize the same Gaussian process model and same acquisition optimizer. The only difference between them is that the standard EI function searches in the whole design space while our SSEI function searches in  randomly selected subspaces for querying a new point. 

\begin{figure}
	\centering
	\subfloat[2-D Michalewicz]{\includegraphics[width=0.33\linewidth]{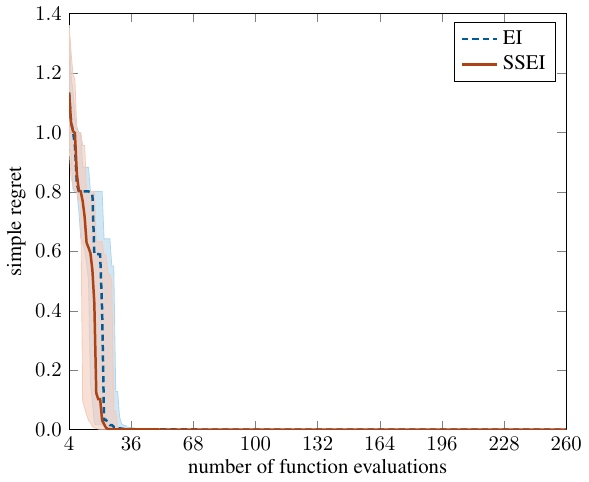}} \hfil
	\subfloat[4-D Michalewicz]{\includegraphics[width=0.33\linewidth]{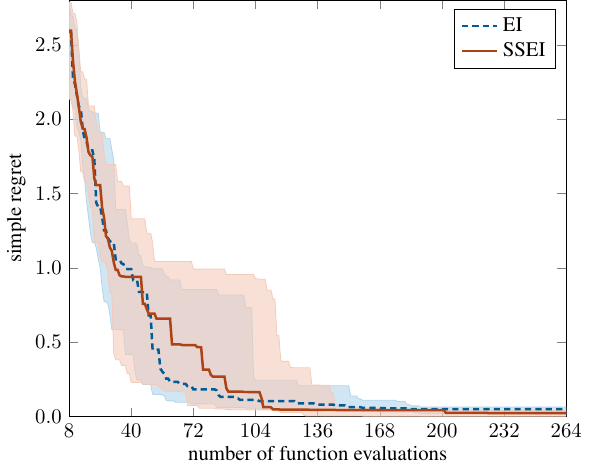}} \hfil
	\subfloat[8-D Michalewicz]{\includegraphics[width=0.33\linewidth]{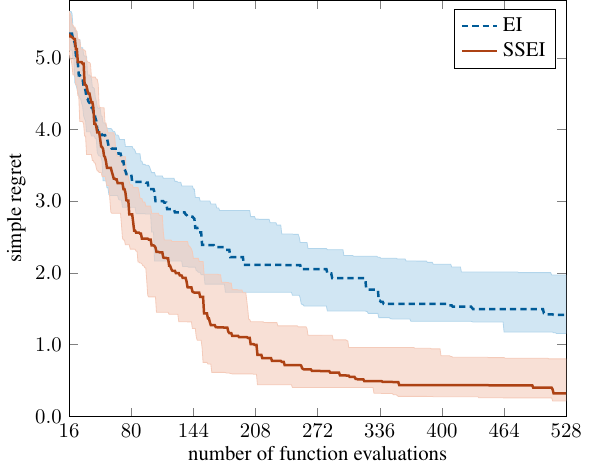}}   \\ 
	\subfloat[10-D Michalewicz]{\includegraphics[width=0.33\linewidth]{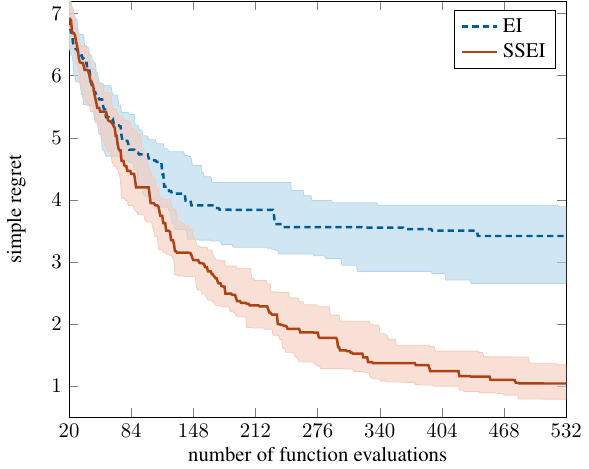}}\hfil
	\subfloat[20-D Michalewicz]{\includegraphics[width=0.33\linewidth]{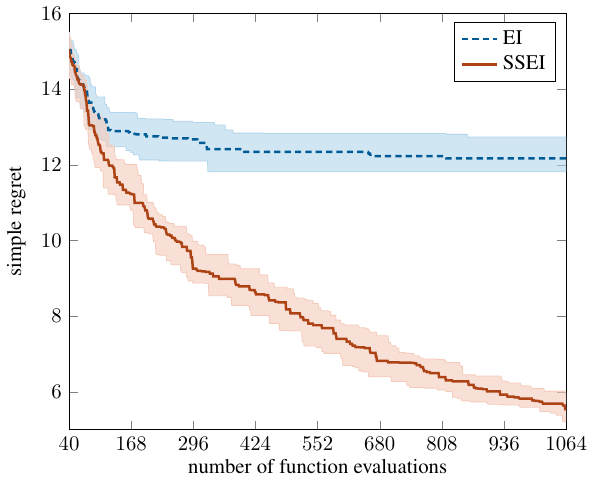}}\hfil
	\subfloat[30-D Michalewicz]{\includegraphics[width=0.33\linewidth]{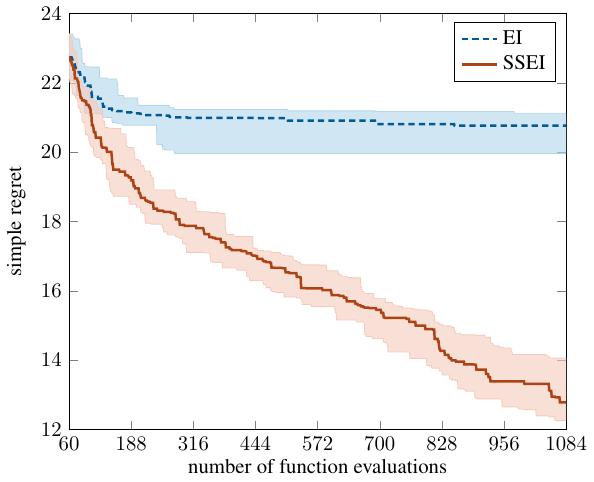}}\\
	\subfloat[2-D Schwefel]{\includegraphics[width=0.33\linewidth]{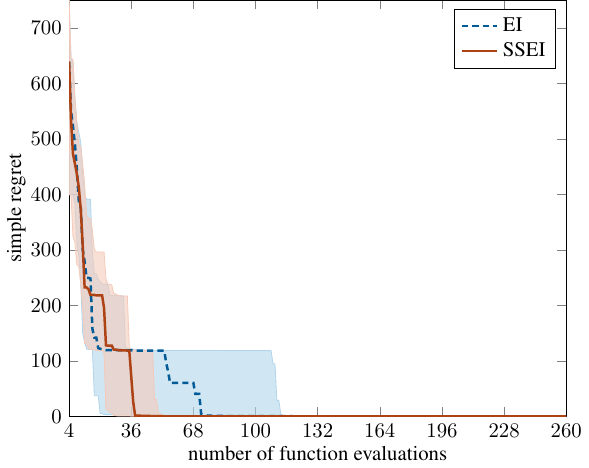}} \hfil
	\subfloat[4-D Schwefel]{\includegraphics[width=0.33\linewidth]{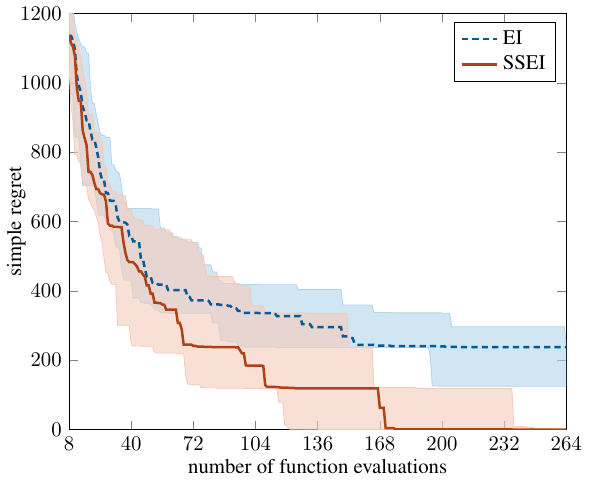}} \hfil
	\subfloat[8-D Schwefel]{\includegraphics[width=0.33\linewidth]{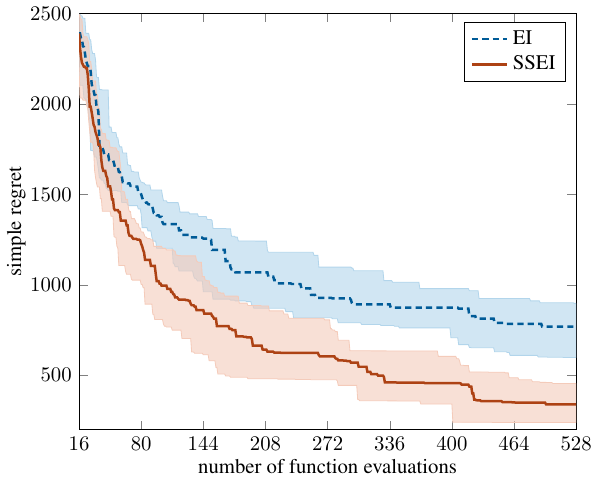}}   \\ 
	\subfloat[10-D Schwefel]{\includegraphics[width=0.33\linewidth]{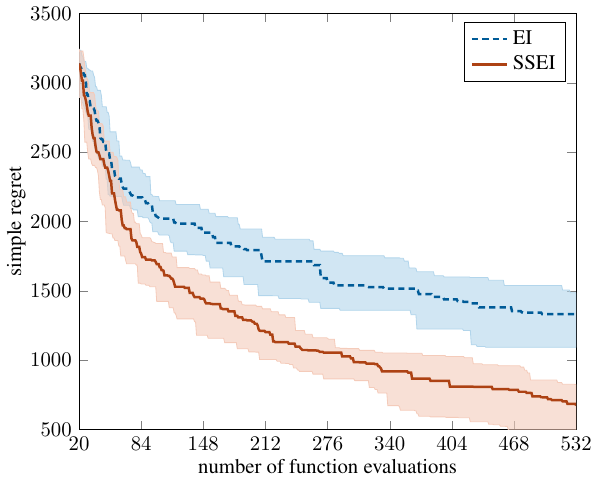}}\hfil
	\subfloat[20-D Schwefel]{\includegraphics[width=0.33\linewidth]{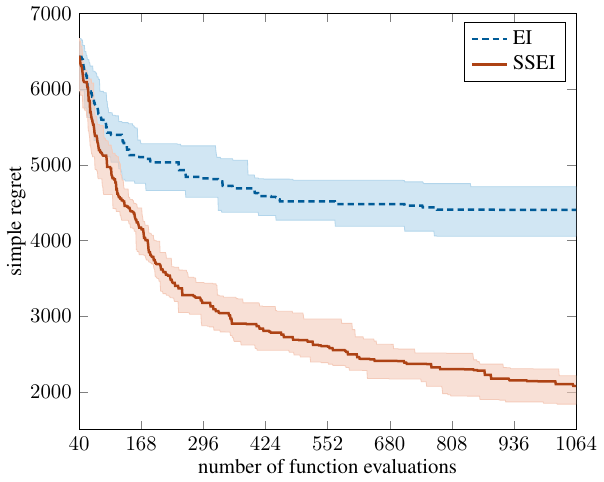}}\hfil
	\subfloat[30-D Schwefel]{\includegraphics[width=0.33\linewidth]{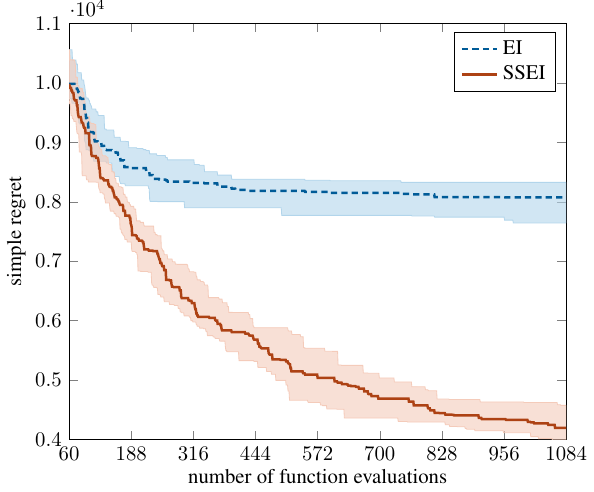}}\\
	\caption{Convergence curves of the SSEI and EI approaches on Michalewicz and Schwefel problems for $q=1$.}
	\label{fig_SSEI_EI}
\end{figure}

The convergence histories of the SSEI and the EI approaches on the complex Michalewicz and Schwefel problems with different dimensions are plotted in Fig.~\ref{fig_SSEI_EI}, where  the median, the first quartile, and the third quartile of the $30$ runs are shown.  We can see that, on the $2$-D and $4$-D Michalewicz and  Schwefel problems, the performances of SSEI and EI are very similar. The convergence speeds of these two approaches are very close and they find very similar optimization results at the end of iterations. As the dimension of the problem increases, the advantage of the proposed SSEI over the standard EI becomes more and more significant. The proposed SSEI finds much better optimization results than the standard EI on $8$-D and  higher-dimensional problems.  

An interesting phenomenon we can observed from Fig.~\ref{fig_SSEI_EI} is that the convergence speed of the EI approach is similar to the SSEI approach at the beginning of the iterations but is slower than SSEI as the iterations continue on high-dimensional problems.  To further explore this phenomenon, we record the EI values that the L-BFGS-B optimizer finds in the whole design space and in randomly selected subspaces respectively along the iterations. 

\begin{figure}
	\centering
	\subfloat[2-D Michalewicz]{\includegraphics[width=0.33\linewidth]{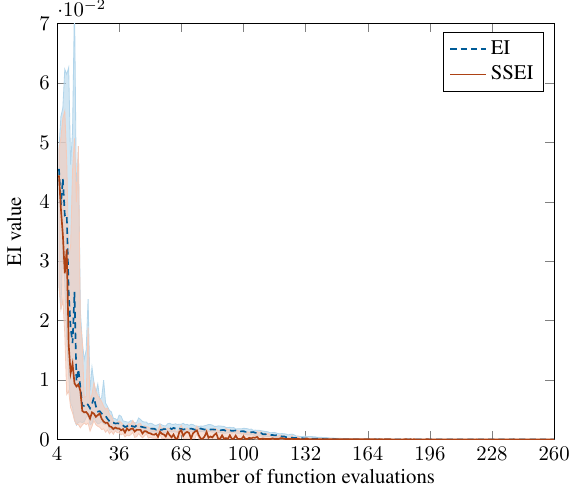}} \hfil
	\subfloat[4-D Michalewicz]{\includegraphics[width=0.33\linewidth]{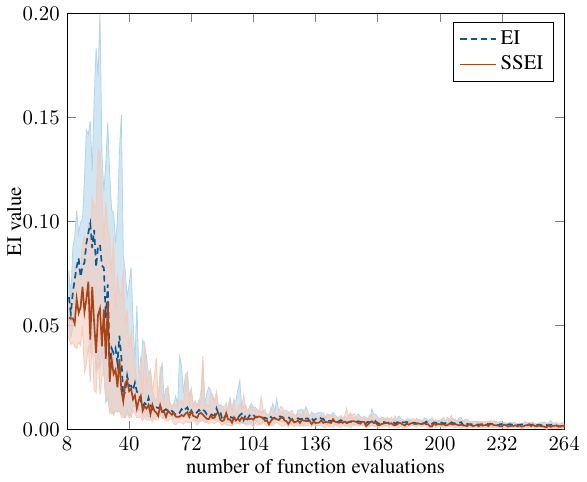}} \hfil
	\subfloat[8-D Michalewicz]{\includegraphics[width=0.33\linewidth]{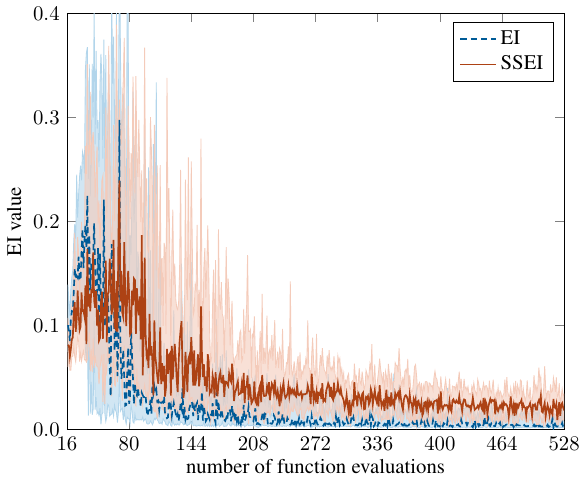}}   \\ 
	\subfloat[10-D Michalewicz]{\includegraphics[width=0.33\linewidth]{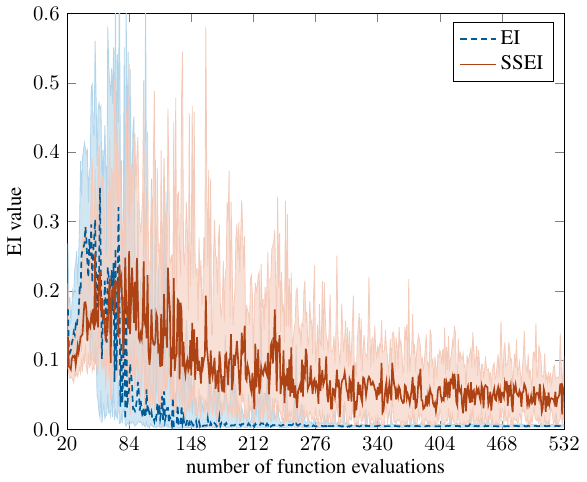}}\hfil
	\subfloat[20-D Michalewicz]{\includegraphics[width=0.33\linewidth]{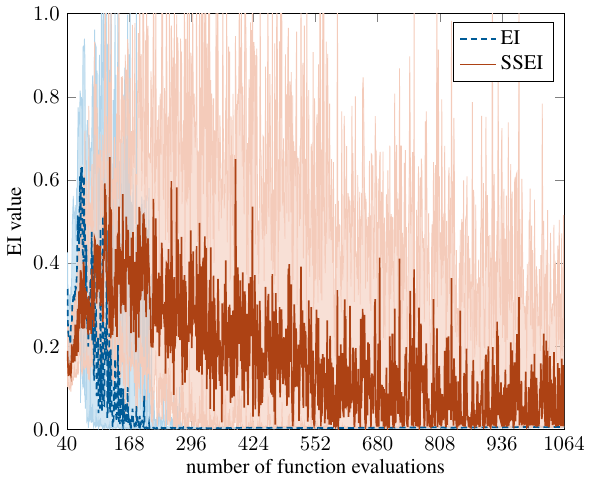}}\hfil
	\subfloat[30-D Michalewicz]{\includegraphics[width=0.33\linewidth]{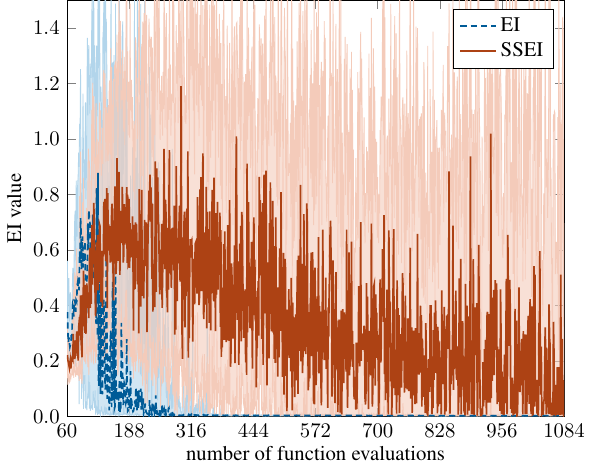}}\\
	\subfloat[2-D Schwefel]{\includegraphics[width=0.33\linewidth]{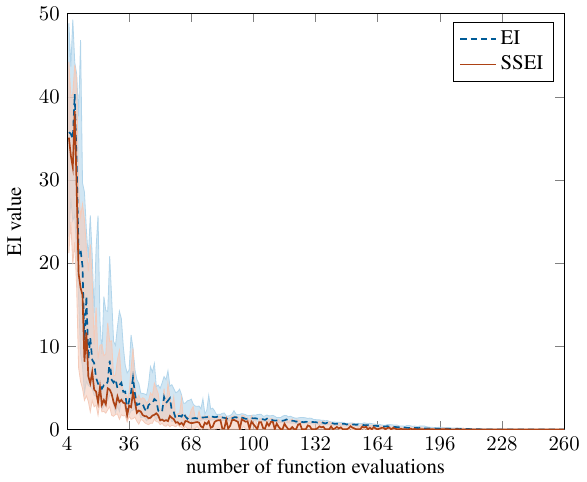}} \hfil
	\subfloat[4-D Schwefel]{\includegraphics[width=0.33\linewidth]{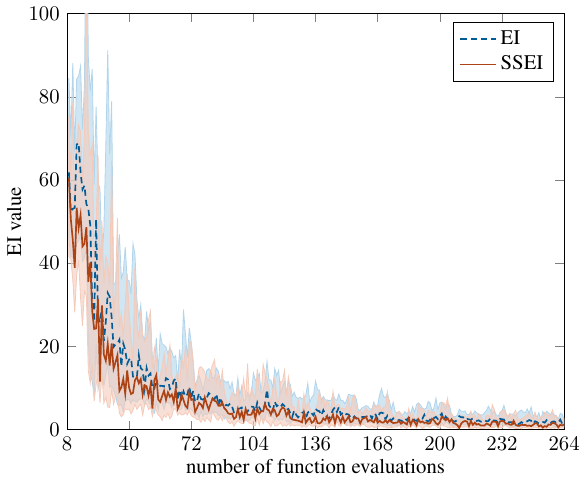}} \hfil
	\subfloat[8-D Schwefel]{\includegraphics[width=0.33\linewidth]{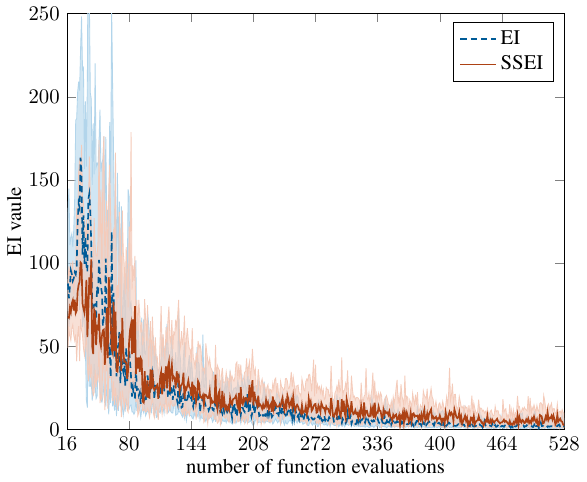}}   \\ 
	\subfloat[10-D Schwefel]{\includegraphics[width=0.33\linewidth]{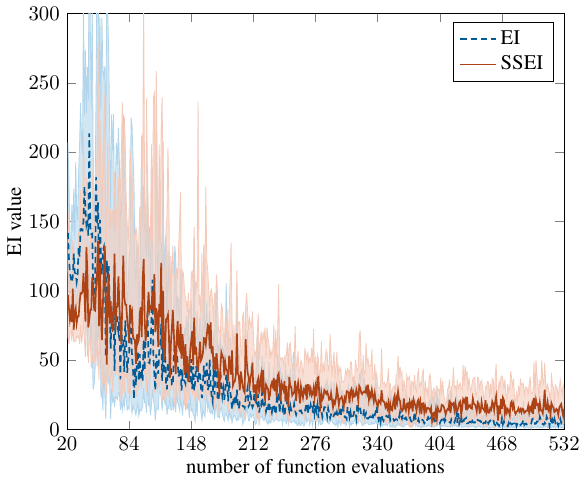}}\hfil
	\subfloat[20-D Schwefel]{\includegraphics[width=0.33\linewidth]{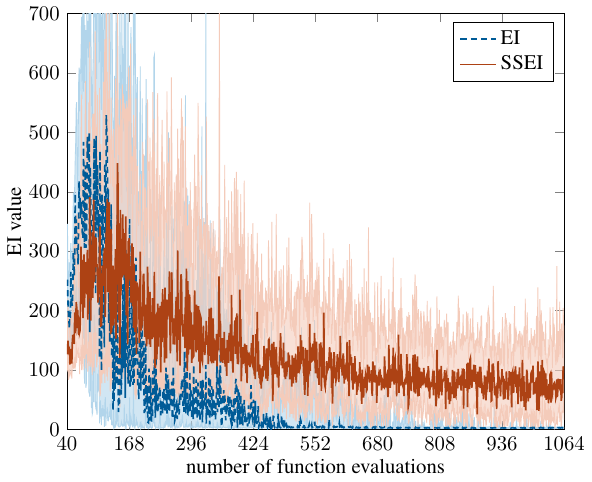}}\hfil
	\subfloat[30-D Schwefel]{\includegraphics[width=0.33\linewidth]{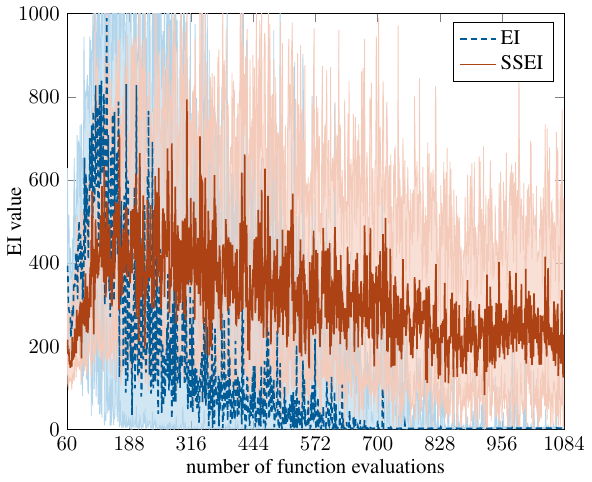}}\\
	\caption{Recorded EI values finds by the SSEI and EI approaches on Michalewicz and Schwefel problems.}
	\label{fig_record_EI}
\end{figure}

We plot the EI values found by the L-BFGS-B optimizer  in Fig.~\ref{fig_record_EI}, where the median, the first quartile and the third quartile of $30$ runs are shown. First, on $2$-D and $4$-D problems, the EI values finds by the L-BFGS-B in the whole design space and in subspaces are very close.  This explains why SSEI achieves similar performance compared with the standard EI on $2$-D and $4$-D problems.  On $8$-D and higher-dimensional problems, we can clearly see from Fig.~\ref{fig_record_EI} that the SSEI approach finds smaller EI values than the EI approach at the beginning of iterations, but significantly larger EI values at the middle and end of iterations. This can be explained by the exploration and exploitation trade-off.  At the beginning of iterations, exploration is needed for finding different promising areas.  Therefore, searching in the whole design space is more beneficial and is able to  find higher EI values than searching in subspaces.  However, as the iterations goes on, most of the promising areas have been located and exploitation is needed for finding better solutions. The EI approach still searches in the whole design space, thus fails to locate better solutions.  In comparison,  the proposed SSEI approach is able to locate solutions with higher EI values by introducing more local search through only changing a part of variables of the current best solution.   This explains why SSEI performs similarly to EI at the beginning but outperforms EI at the middle and end of iterations on high-dimensional problems. The experiment results show that, when the computational budget of acquisition functions is fixed, instead of spending it all in the whole design space, spending it in fewer dimensions can improve the solution quality especially on high-dimensional problems.

The detailed experiment results on sixty test problems are given in Table~\ref{table_EI_SSEI} in Appendix~\ref{section_results}, where the best results are highlighted in bold.  We conduct the Wilcoxon signed rank test to find out whether the results obtained by the proposed SSEI have significant difference from the results obtained by the EI. The significance level of the tests is set to $\alpha=0.05$. We use $+$, $-$ and $\approx$ to represent that our SSEI approach finds significantly better, significantly worse and similar results compared with the EI approach, respectively.  From the table we can draw very similar conclusion.  The performance of SSEI is similar to EI on $2$-D and $4$-D problems, and better than EI on high-dimensional problems. On low-dimensional problems, the acquisition optimizer is often able to fully search the whole design space.  However, as the dimension increases, it becomes more challenging for the  optimizer to find the global optimum of the acquisition function due to the curse of dimensionality.  Therefore, the trade-off between global exploration and local exploitation is needed. The EI approach searches in the whole design space all the time, thus often fails to locate optimal solutions at the middle and end of iterations when local exploitation is needed. In contrast,  by introducing more local search around the current best solution, the proposed SSEI is able to find better solutions at the middle and end of iterations, thus achieve better optimization results than the EI approach. These experiment results also indicate that we can improve the performance of Bayesian optimization in high-dimensional spaces by simply restricting the acquisition optimization in  randomly selected subspaces.

Introducing more local search to BO is helpful in high-dimensional spaces~\cite{Eriksson_2019,Rashidi_2024,Papenmeier_2025}. In the following, we compare our SSEI with the TuRBO~\cite{Eriksson_2019} which perturbs $\min(20,d)$ coordinates when selecting query points for expensive evaluation. In addition, we apply our random subspace selection strategy to TuRBO to randomly select a subspace with dimension of $K \sim U(1, d) $ to perturb, which is called SS-TuRBO in the following comparison.  The open-sourced code of TuRBO~\footnote{\url{https://github.com/uber-research/TuRBO}} is used in the comparison. The batch size is also set to $q=1$. The difference between SSEI and TuRBO is that our SSEI uses global GP models and selects the query point in a randomly selected subspace while the TuRBO uses local GP models with the trust region strategy and selects the query point using Thompson sampling by perturbing $\min(20,d)$ coordinates. The difference between SS-TuRBO and TuRBO is that the SS-TuRBO perturbs on average about half of the dimensions while TuRBO perturbs $\min(20,d)$ coordinates.

\begin{figure}
	\centering
	\subfloat[2-D Michalewicz]{\includegraphics[width=0.33\linewidth]{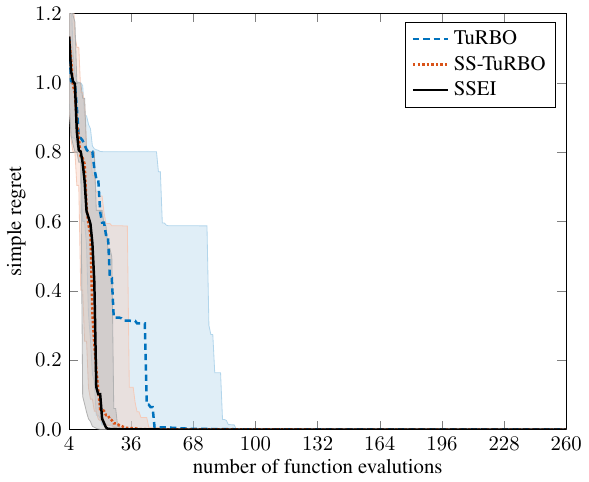}} \hfil
	\subfloat[4-D Michalewicz]{\includegraphics[width=0.33\linewidth]{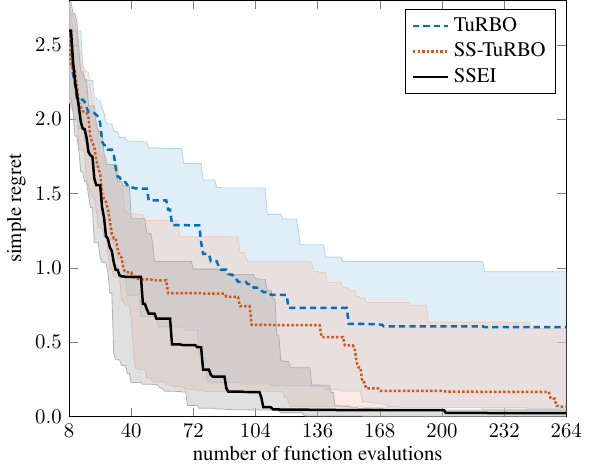}} \hfil
	\subfloat[8-D Michalewicz]{\includegraphics[width=0.33\linewidth]{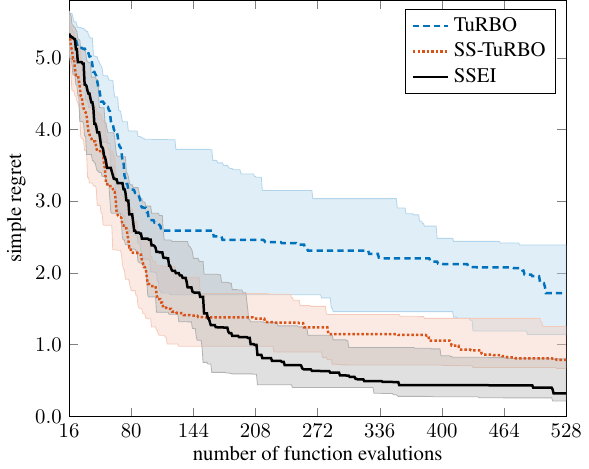}}   \\ 
	\subfloat[10-D Michalewicz]{\includegraphics[width=0.33\linewidth]{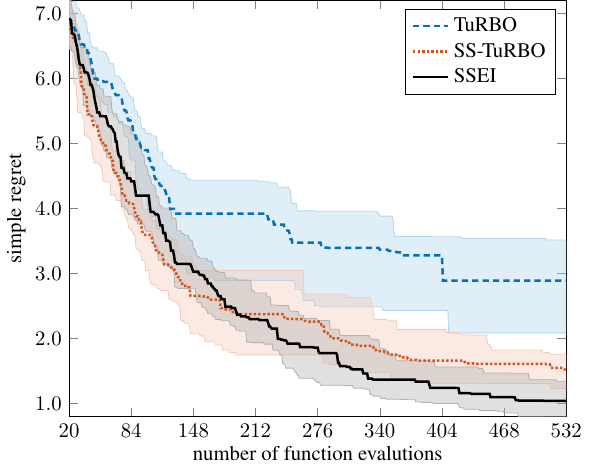}}\hfil
	\subfloat[20-D Michalewicz]{\includegraphics[width=0.33\linewidth]{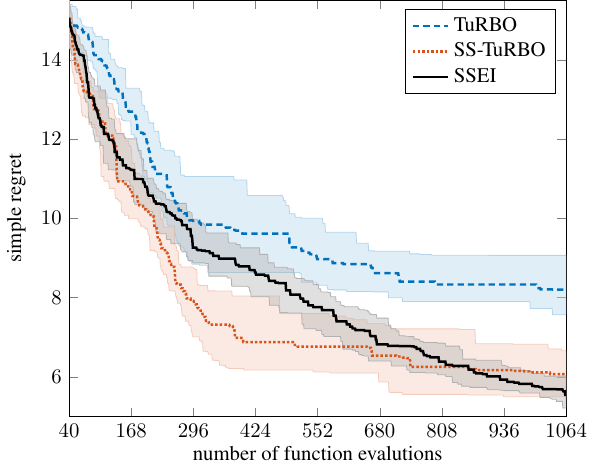}}\hfil
	\subfloat[30-D Michalewicz]{\includegraphics[width=0.33\linewidth]{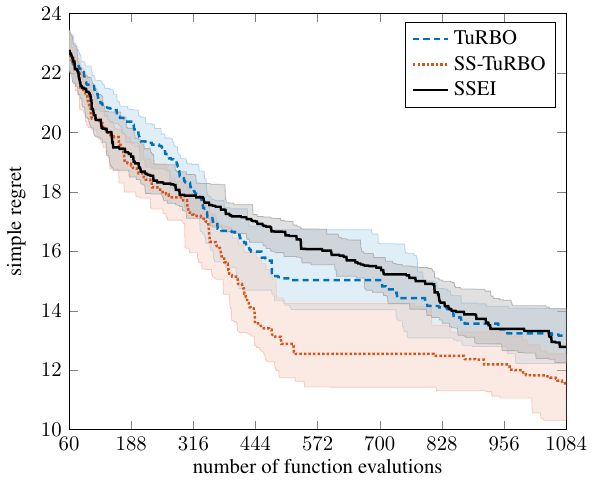}}\\
	\subfloat[2-D Schwefel]{\includegraphics[width=0.33\linewidth]{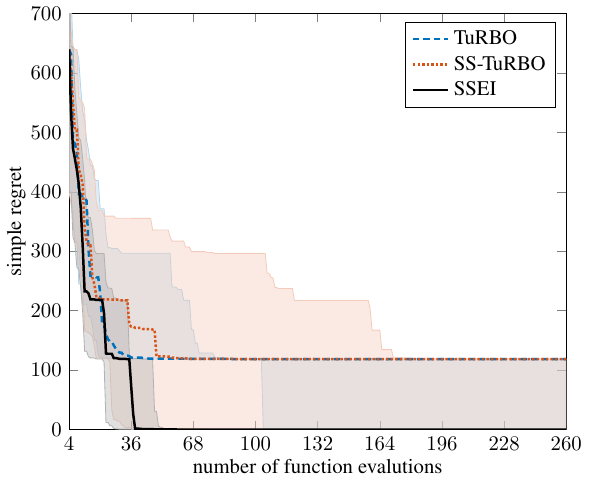}} \hfil
	\subfloat[4-D Schwefel]{\includegraphics[width=0.33\linewidth]{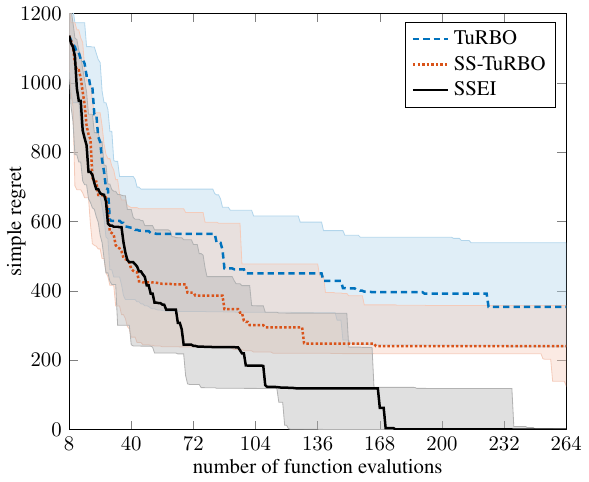}} \hfil
	\subfloat[8-D Schwefel]{\includegraphics[width=0.33\linewidth]{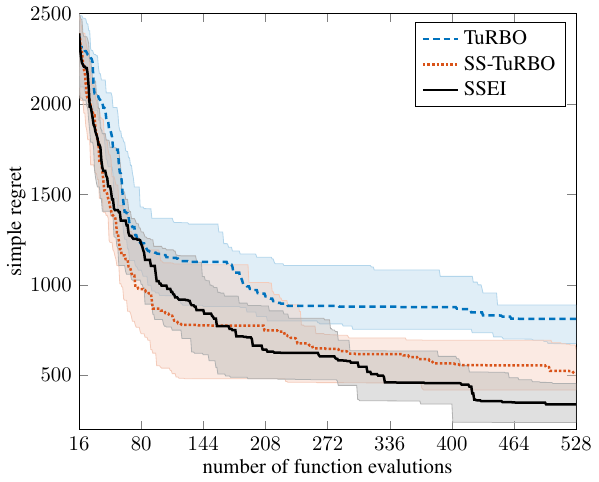}}   \\ 
	\subfloat[10-D Schwefel]{\includegraphics[width=0.33\linewidth]{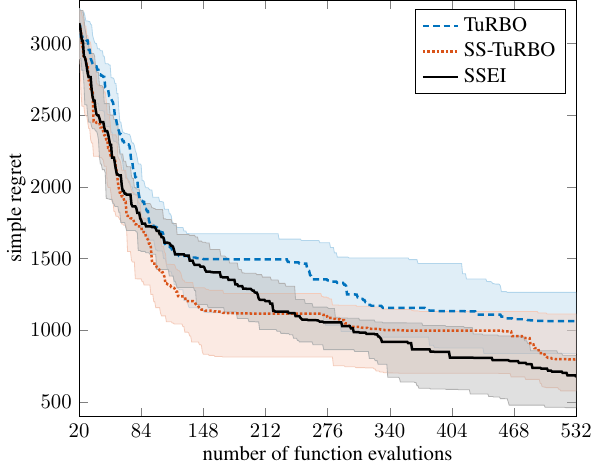}}\hfil
	\subfloat[20-D Schwefel]{\includegraphics[width=0.33\linewidth]{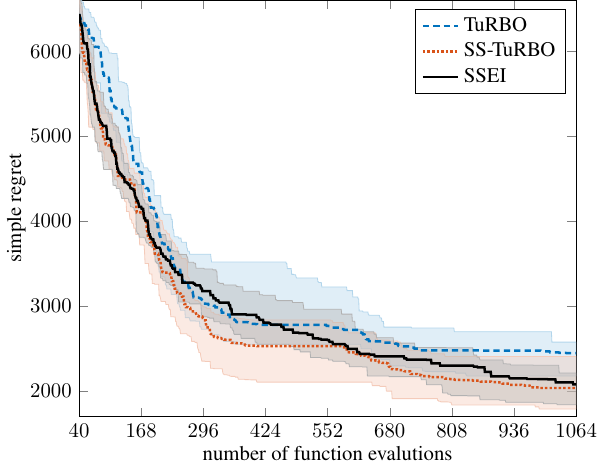}}\hfil
	\subfloat[30-D Schwefel]{\includegraphics[width=0.33\linewidth]{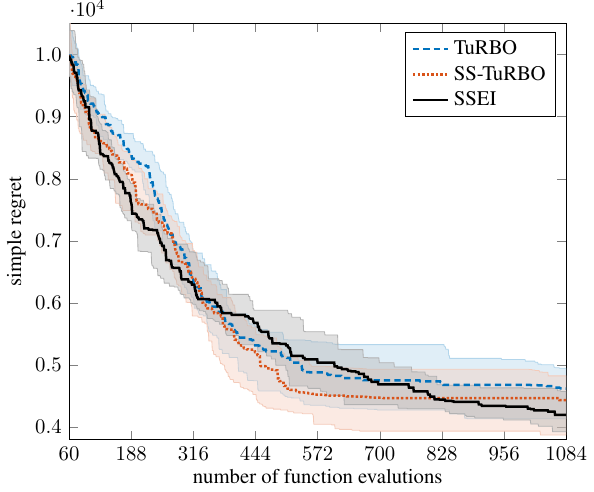}}\\
	\caption{Convergence histories of the TuRBO, SS-TuRBO and SSEI on the Michalewicz and Schwefel problems when only query point is selected in each iteration.}
	\label{fig_TuRBO}
\end{figure}

The convergence curves of the compared TuRBO, SS-TuRBO and SSEI on the Michalewicz and Schwefel problems are presented in Fig.~\ref{fig_TuRBO}, where the median, the first quartile, and the third quartile of the $30$ runs are plotted.  On problems with dimension $d \le 20$,  TuRBO perturbs all the dimensions to generate candidate points to choose from while SS-TuRBO perturbs on average near half of the dimensions to generate candidates. On 30-D problems, TuRBO perturbs $20$ dimensions while  SS-TuRBO perturbs near $15$ dimensions on average.  We can see clearly from the figures that SS-TuRBO converges significantly faster than TuRBO on both the Michalewicz and Schwefel problems across different dimensions. The Thompson sampling approach samples a discrete set of candidate points using GP posterior and selects the best solution from the candidates for expensive evaluation. Perturbing fewer dimensions to generate candidate points brings more local search around the current best solution, which seems to be more  beneficial not only on high-dimensional problems but also on low-dimension problems. The proposed SSEI and SS-TuRBO perbturb similar number of dimensions when selecting query points for expensive evaluation. The difference between them is that the SSEI approach uses global GP model and EI acquisition function to select query points while SS-TuRBO uses local GP model with trust region strategy and Thompson sampling to select query points. From the figures, we can find SSEI outperforms SS-TuRBO on 2-D, 4-D, 8-D and 10-D problems, and performs very similarly to SS-TuRBO on 20-D and 30-D problems. This indicates that global GP models seems to be more beneficial on low-dimensional problems and local GP models seem to have more advantage on high-dimensional problems.

\vspace{4mm}
\subsection{Study on  Batch Sizes}
The second problem we want to investigate is  how well our proposed subspace approach performs with respect to different batch sizes. We set the batch size $q$ to $1, 2, 4, 8, 16, 32, 64$ and $128$, respectively. The maximum number of axis-aligned subspaces is $N=2^2-1=3$ and $N=2^4-1=15$ for $2$-D and $4$-D problems. In these cases, the SS$q$EI in (\ref{eq_SSqEI}) is employed to locate multiple query points in each subspace when the required batch size $q$ is larger than the number of available subspaces $N$. For problems with $d \ge 8$, the number of available subspace is larger than $128$, therefore only the SSEI in (\ref{eq_ESSI}) is employed to select one query point in each subspace. 

The performance of a batch algorithm with respect to different batch sizes can be interpreted in two different ways: the performance against the number of function evaluations and  the performance against the number of iterations.  The first interpretation can show how well the batch algorithm performs in terms of  the computational budget while the second interpretation can show how well the batch algorithm performs in terms of  the wall-clock time.

\begin{figure}
	\centering
	\subfloat[2-D Michalewicz]{\includegraphics[width=0.33\linewidth]{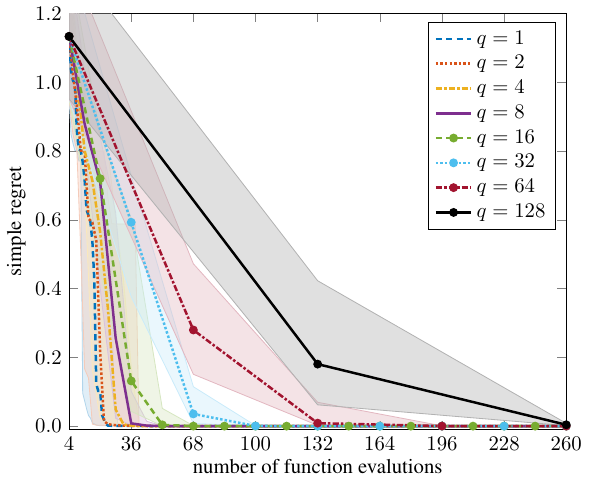}} \hfil
	\subfloat[4-D Michalewicz]{\includegraphics[width=0.33\linewidth]{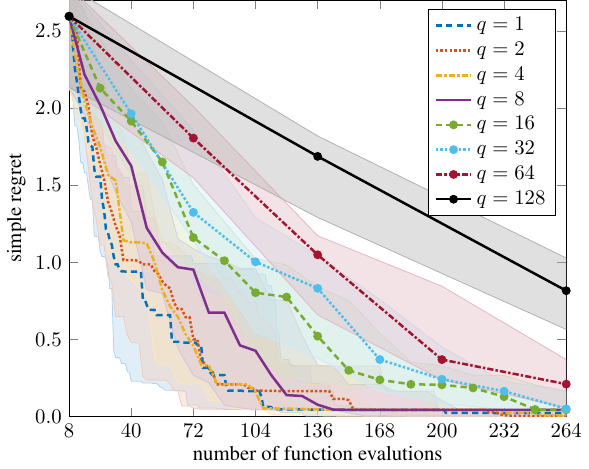}} \hfil
	\subfloat[8-D Michalewicz]{\includegraphics[width=0.33\linewidth]{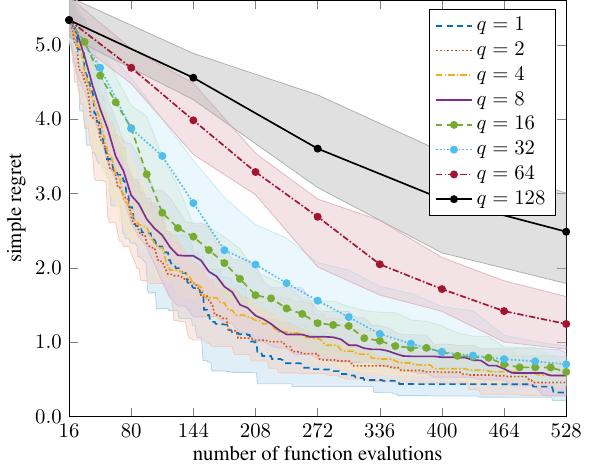}}   \\ 
	\subfloat[10-D Michalewicz]{\includegraphics[width=0.33\linewidth]{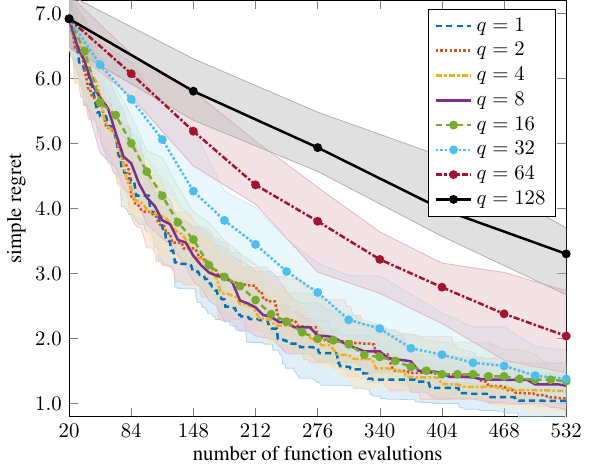}}\hfil
	\subfloat[20-D Michalewicz]{\includegraphics[width=0.33\linewidth]{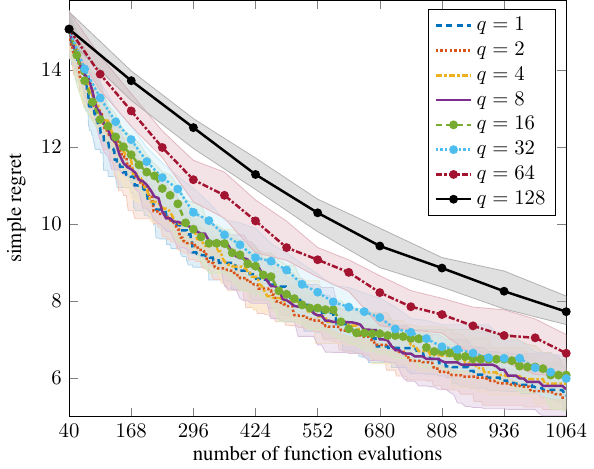}}\hfil
	\subfloat[30-D Michalewicz]{\includegraphics[width=0.33\linewidth]{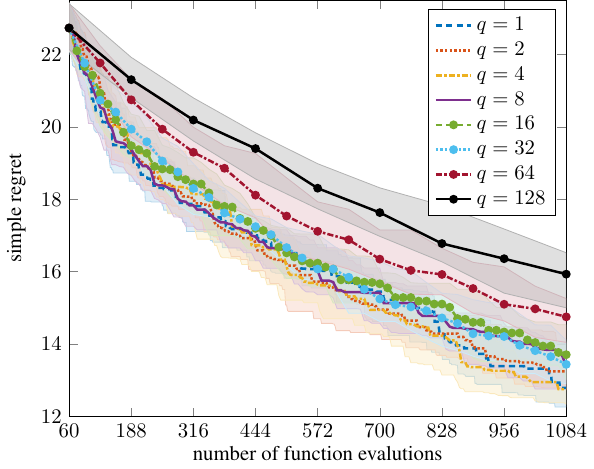}}\\
	\subfloat[2-D Schwefel]{\includegraphics[width=0.33\linewidth]{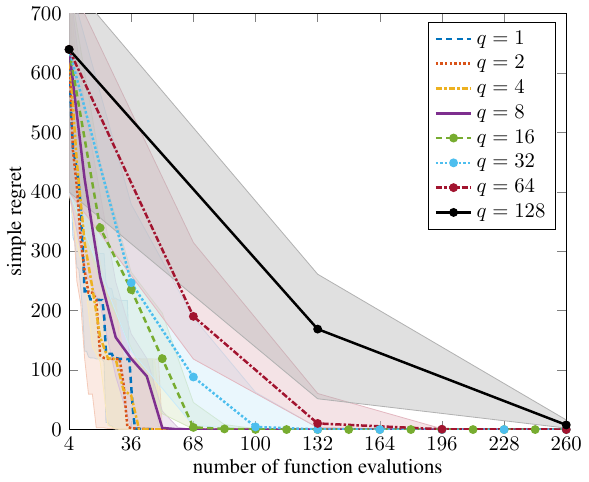}} \hfil
	\subfloat[4-D Schwefel]{\includegraphics[width=0.33\linewidth]{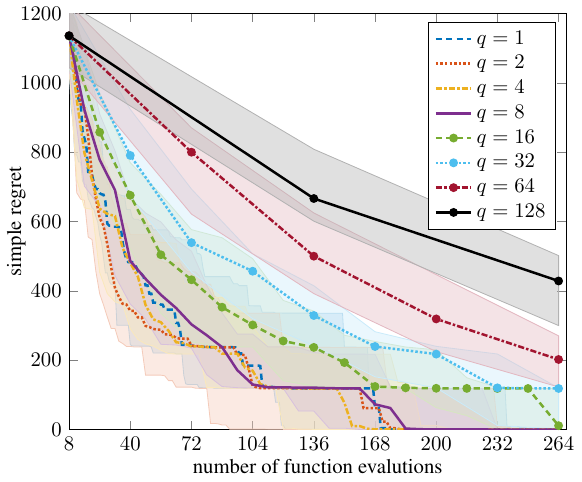}} \hfil
	\subfloat[8-D Schwefel]{\includegraphics[width=0.33\linewidth]{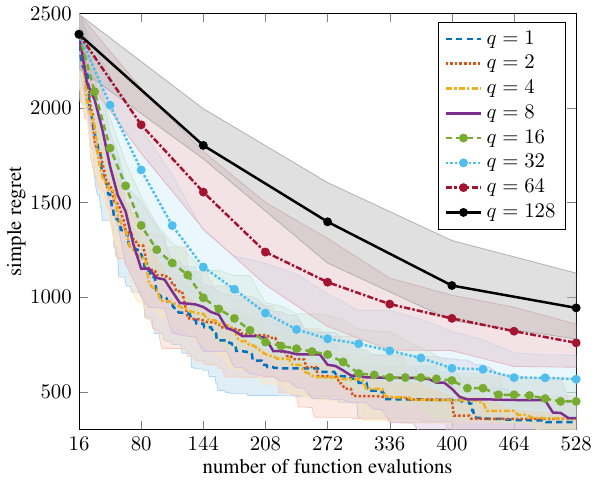}}   \\ 
	\subfloat[10-D Schwefel]{\includegraphics[width=0.33\linewidth]{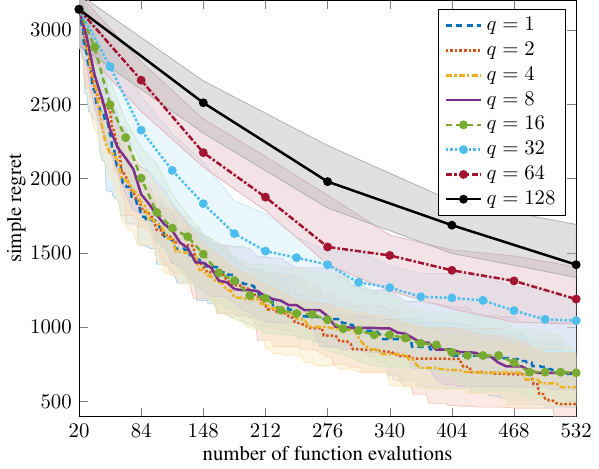}}\hfil
	\subfloat[20-D Schwefel]{\includegraphics[width=0.33\linewidth]{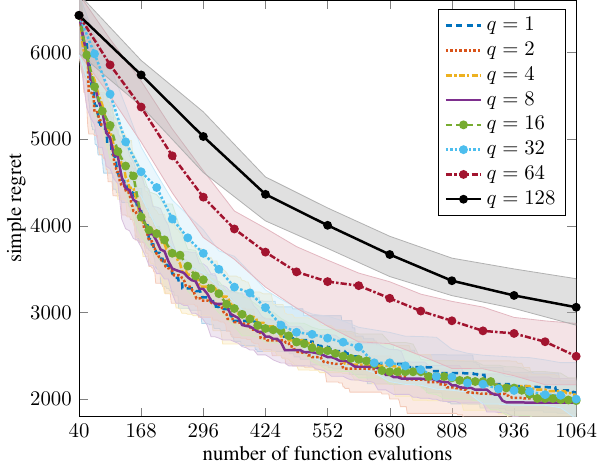}}\hfil
	\subfloat[30-D Schwefel]{\includegraphics[width=0.33\linewidth]{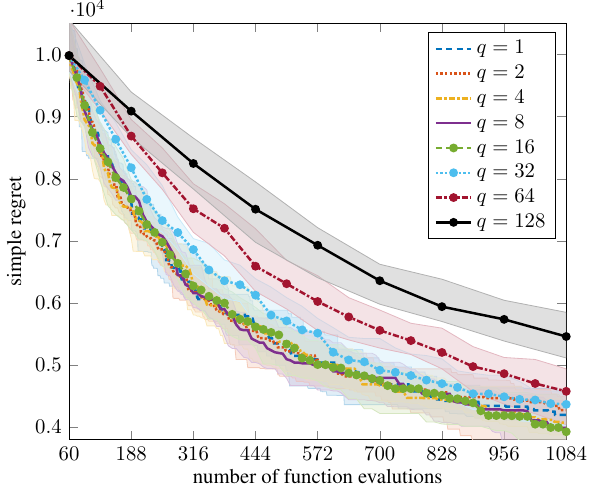}}\\
	\caption{Convergence histories of the proposed SSEI with different batch sizes on Michalewicz and Schwefel problems with respect to the number of function evaluations.}
	\label{fig_converge_evaluations}
\end{figure}

We show the convergence histories of our proposed SSEI approach with respect to the number of function evaluations in Fig.~\ref{fig_converge_evaluations}, where the median, the first quartile, and the third quartile of 30 runs are plotted. Often, the performance of the batch algorithm decreases as the batch size increases when the computational budget is fixed. The reason is that the GP model will be updated fewer times when more points are queried within one iteration. Therefore, the model accuracy is often less accurate compared with situations where fewer points are queried within one iteration.   From Fig.~\ref{fig_converge_evaluations} we can see that, the performance of our proposed SSEI decreases very slightly as the batch size increases from $1$ to $32$. The converge speeds of SSEI are pretty similar, and the optimization results obtained by SSEI  are very close for all $q \le 32$.  The experiment results indicate that our proposed SSEI is able to achieve nearly linear speedup when the batch size is under $32$. However, as the batch size further grows from $32$ to $128$, the performance of SSEI  deteriorates gradually. As can be seen in Fig.~\ref{fig_converge_evaluations}, the SSEI approach converges slower and finds worse results when the batch size increase from $32$ to $128$.  The detailed experiment results of SSEI using different batch sizes are given in Table~\ref{table_SSEI} in Appendix~\ref{section_results},  from which we can draw very similar conclusion that the proposed SSEI is able to achieve nearly linear speedup as the batch size increases from $1$ to $32$, but the performance of SSEI declines as the batch size further grows from $32$ to $128$. 

\begin{figure}
	\centering
	\subfloat[2-D Michalewicz]{\includegraphics[width=0.33\linewidth]{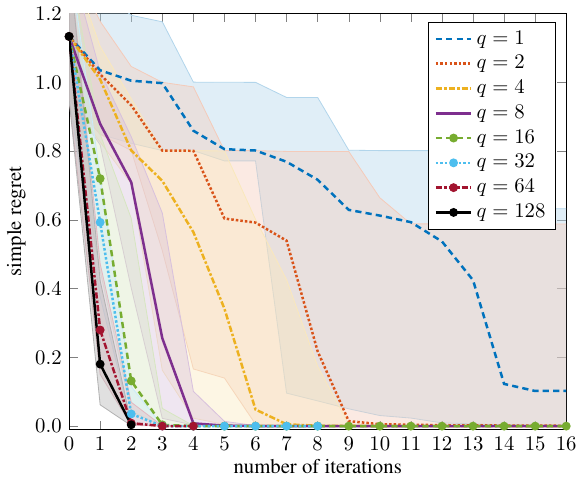}} \hfil
	\subfloat[4-D Michalewicz]{\includegraphics[width=0.33\linewidth]{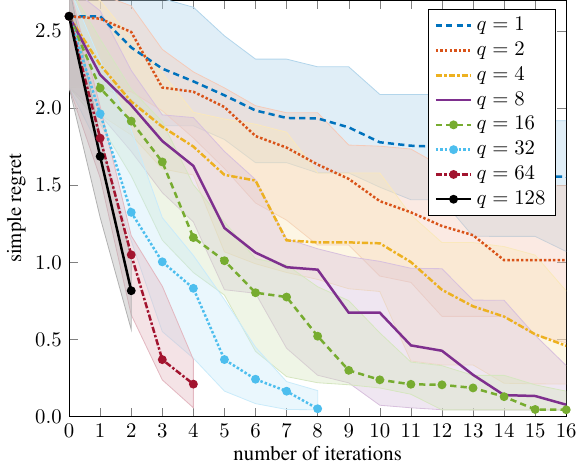}} \hfil
	\subfloat[8-D Michalewicz]{\includegraphics[width=0.33\linewidth]{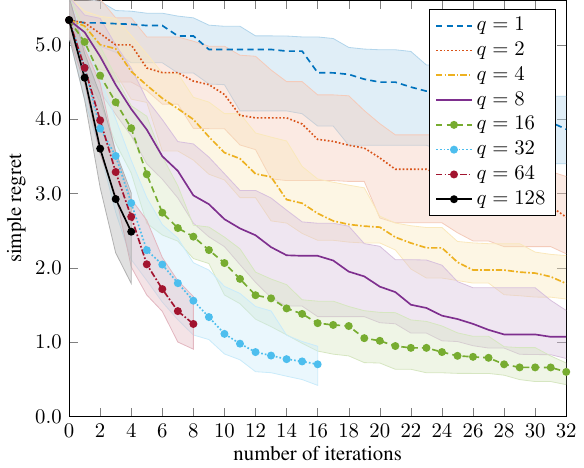}}   \\ 
	\subfloat[10-D Michalewicz]{\includegraphics[width=0.33\linewidth]{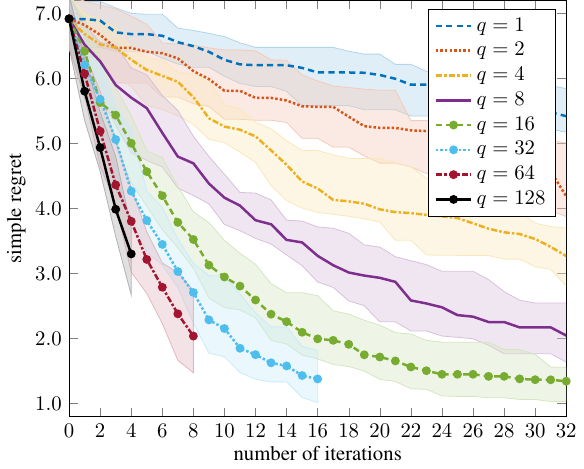}}\hfil
	\subfloat[20-D Michalewicz]{\includegraphics[width=0.33\linewidth]{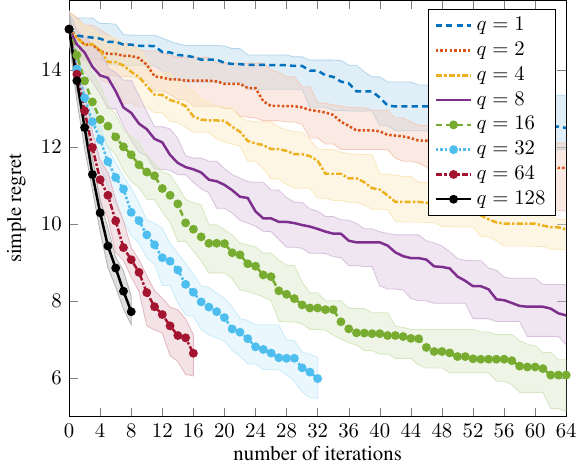}}\hfil
	\subfloat[30-D Michalewicz]{\includegraphics[width=0.33\linewidth]{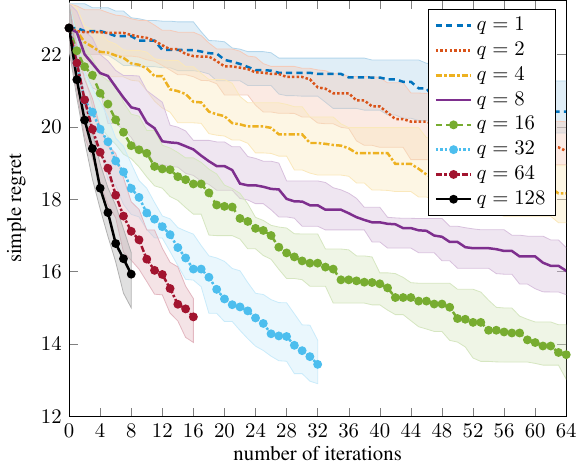}}\\
	\subfloat[2-D Schwefel]{\includegraphics[width=0.33\linewidth]{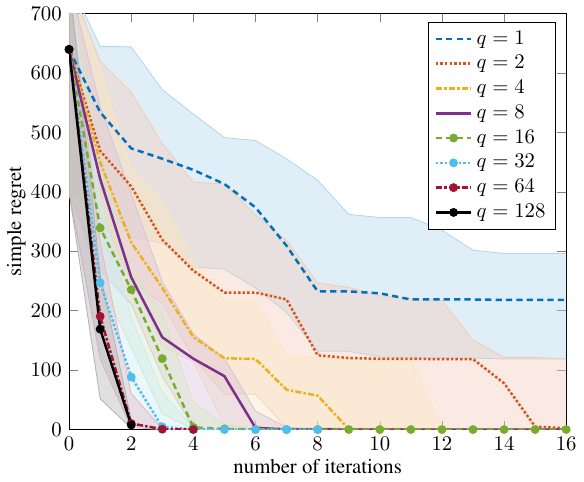}} \hfil
	\subfloat[4-D Schwefel]{\includegraphics[width=0.33\linewidth]{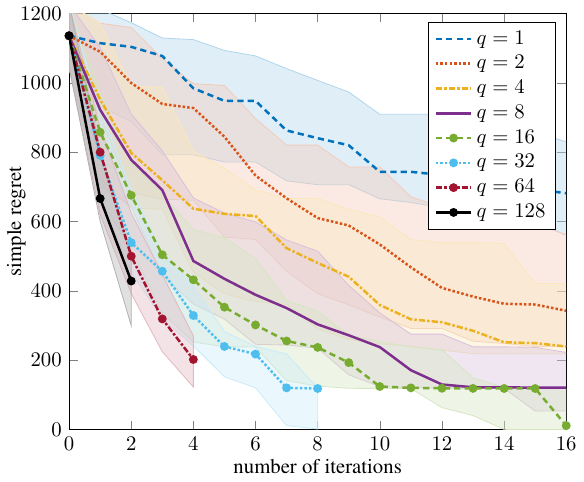}} \hfil
	\subfloat[8-D Schwefel]{\includegraphics[width=0.33\linewidth]{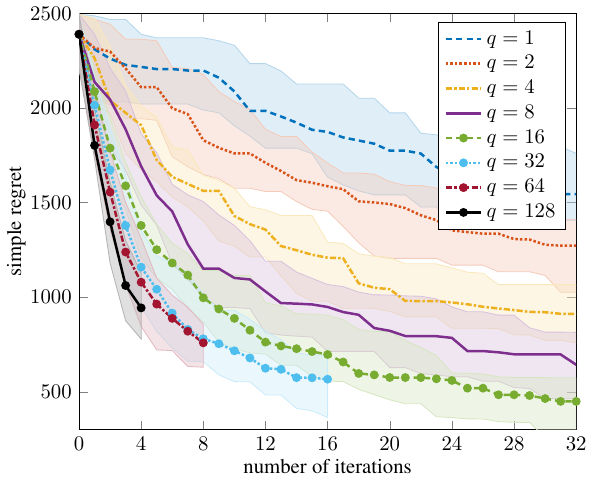}}   \\ 
	\subfloat[10-D Schwefel]{\includegraphics[width=0.33\linewidth]{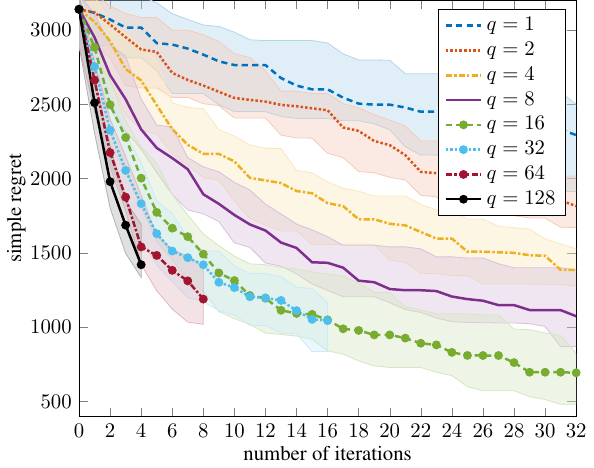}}\hfil
	\subfloat[20-D Schwefel]{\includegraphics[width=0.33\linewidth]{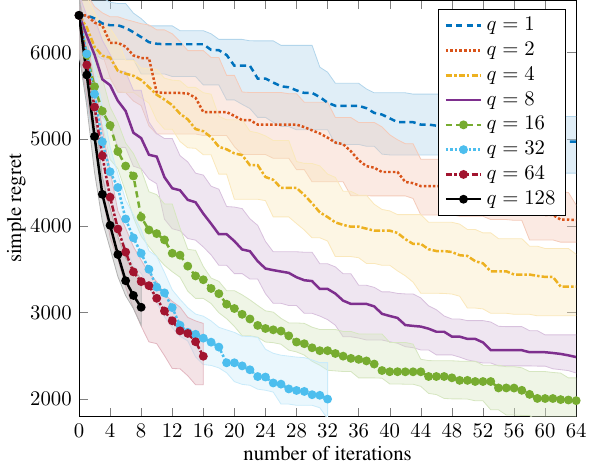}}\hfil
	\subfloat[30-D Schwefel]{\includegraphics[width=0.33\linewidth]{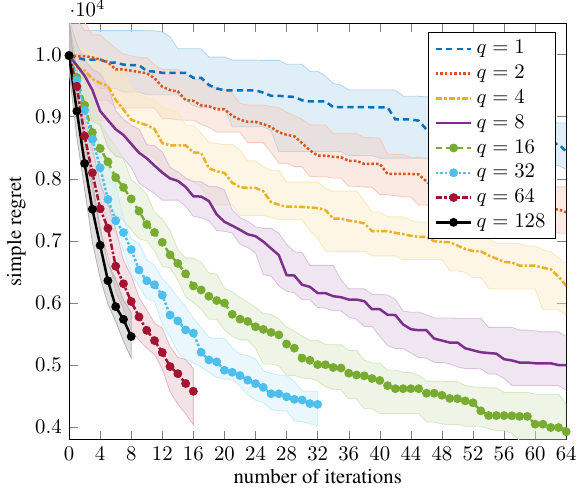}}\\
	\caption{Convergence histories of the proposed SSEI with different batch sizes on Michalewicz and Schwefel problems with respect to the number of iterations.}
	\label{fig_converge_iterations}
\end{figure}

In Fig.~\ref{fig_converge_iterations} we illustrate the convergence histories of the proposed SSEI approach with respect to  the number of iterations. This can show how well the proposed SSEI performs at the same wall-clock time when  all the batch samples are evaluated in parallel.  We can see that the performance of our proposed SSEI approach increases monotonously as the batch size grows.  When the number of query points increases, the proposed SSEI approach is able to explore more axis-aligned subspaces within one iteration, thus accelerate the convergence speed. From Fig.~\ref{fig_converge_iterations} we can see that, as the batch size increases from $1$ to $128$, the proposed batch approach converges faster and finds better optimization results at the end of iterations.  These experiment results can empirically prove the effectiveness of our proposed SSEI approach.

\subsection{Comparing SSEI with Batch Approaches}

In the following, we compare the proposed SSEI approach with well-established batch Bayesian optimization algorithms.  We implement our proposed subspace acquisition functions using BoTorch~\cite{Balandat_2020}.  We use a constant mean function and a squared exponential kernel for the GP model, and use the L-BFGS-B optimizer with $1024$ initial samples and  $20$ restarts to optimize the acquisition functions. The BoTorch implementation of our proposed SSEI is available at \url{https://github.com/zhandawei/SubSpace_Acquisition_Functions}.

The compared algorithms are the Random Search (RS) approach, the Multi-Point Expected Improvement ($q$EI) approach~\cite{Ginsbourger_2010}, the $q$logEI~\cite{Ament_2023}, the Fast Multi-Point Expected Improvement (F$q$EI) approach~\cite{Zhan_2023}, the Trust Region Bayesian Optimization (TuRBO)~\cite{Eriksson_2019},  the Multi-Scale Multi-Recommendations (MSMR) approach~\cite{Joy_2020}, the Multi-objective ACquisition Ensemble (MACE) approach~\cite{Lyu_2018}, the Pseudo Expected Improvement (PEI) approach~\cite{Zhan2017}, the Expected Improvement and Mutual Information (EIMI) approach~\cite{LiZheng_2016} and the Kriging Believer (KB)  approach~\cite{Ginsbourger_2010}.   The settings of the compared batch EI approaches are given in the following.  
\begin{enumerate}
	\item The Random Search (RS) approach randomly draws $q$ Sobol samples in each iteration. It is included in the comparison to assess the scale of improvement of other batch approaches.
	\item The $q$EI~\cite{Ginsbourger_2010} is a popular multi-point acquisition function in Bayesian optimization.  Its exact calculation method is available for smaller batch size~\cite{Chevalier_2013,Marmin_2015}, and Monte Carlo approximation is often used when the batch size is larger than $10$.  We use the BoTorch implementation~\cite{Balandat_2020} of the $q$EI~\footnote{\url{https://github.com/meta-pytorch/botorch}} in the experiments.
	\item The $q$logEI~\cite{Ament_2023} is a recently proposed logEI family of acquisition functions, which have approximately equal optima as the standard EI functions but are much easier to optimize numerically. The $q$logEI is implemented in BoTorch~\cite{Balandat_2020} and the open-source code of $q$logEI\footnote{\url{https://botorch.org/docs/tutorials/closed_loop_botorch_only}} is employed in the experiments.
	\item  The F$q$EI~\cite{Zhan_2023} is a recently proposed batch expected improvement acquisition function in order to tackle the high computational cost problem of the classical $q$EI  acquisition function. The F$q$EI approach uses cooperative approach to decompose the acquisition function into $q$ sub-problems with each sub-problem dealing with one acquisition point. In the experiment, the cooperative genetic algorithm is used to solve the  F$q$EI acquisition function. The population size for each sub-problem is set to $10d$ and the maximal generation is set to $100$. The open-source code of F$q$EI\footnote{\url{https://github.com/zhandawei/Fast_Multipoint_Expected_Improvement}} is used for the experiments.
	\item The TuRBO approach~\cite{Eriksson_2019} employs the trust region approach in Bayesian optimization and select a batch of query points through the Thompson Sampling (TS) method.  The open-source code of TuRBO\footnote{\url{https://github.com/uber-research/TuRBO}} is used in the comparison. The number of trust regions is set to five for $q=4$, ten for $q=16$ and twenty for $q=128$, respectively. 
	\item The MSMR approach~\cite{Joy_2020} uses multiple kernel length-scales to train multiple Gaussian process models and uses these different Gaussian process models to produce a set of candidate solutions. Then, the $k$-medoid clustering is employed to select a batch of  points from the candidates. The number of length-scales is set to $2q$ in the experiments.  We also use genetic algorithm with population size of $10d$ and $100$ generations to optimize these different EI functions for the MSMR approach. We implement the MSMR approach according to the work~\cite{Joy_2020}.
	\item The MACE approach~\cite{Lyu_2018} is a typical multi-objective approach for batch Bayesian optimization. It uses the most popular EI, PI and LCB functions as three objectives and optimizes the three acquisition functions by using a multi-objective evolutionary algorithm. A batch of solutions are then randomly selected from the obtained Pareto front approximation. We use the popular NSGA-II algorithm~\cite{Deb_2002} to solve the multi-objective problem in the experiment. The population size is set to $20d$ and the maximal number of generations of the NSGA-II is set to $100$ for the MACE approach. We implement the MACE approach according this work~\cite{Lyu_2018}.
	\item  The PEI approach~\cite{Zhan2017} uses the influence function to simulate the effect that new acquisition samples bring to the expected improvement function.  The PEI approach sequentially multiplies the influence function of previously selected sample to generate next acquisition samples. We also use genetic algorithm with population size of $10d$ and $100$ maximal generations to maximize the PEI function. The open-source code of PEI\footnote{\url{https://github.com/zhandawei/Bayesian_Optimization_Algorithms}} is used in the comparison.
	\item  The EIMI approach~\cite{LiZheng_2016}  uses the mutual information criterion to measure the similarity of selecting points and avoids to select too similar solutions by maximizing the product of the EI function and the MI function. We also use genetic algorithm with population size of $10d$ and $100$  generations to maximize the EIMI function in the experiment. We implement the EIMI approach according to the work~\cite{LiZheng_2016}.
	\item   The  KB approach~\cite{Ginsbourger_2010} is a popular batch approach. After obtaining the first acquisition point, the KB approach assigns a fake objective value to the acquisition point as the Kriging prediction. We also use genetic algorithm with population size of $10d$ and $100$ maximal generations to maximize the KB acquisition function in the experiments.    
\end{enumerate}

We set the batch size to $q=4, 16$ and $128$ to test the algorithms' performances under small, medium and large batch sizes, respectively.

\begin{figure}
	\centering
	\subfloat[2-D Michalewicz]{\includegraphics[width=0.33\linewidth]{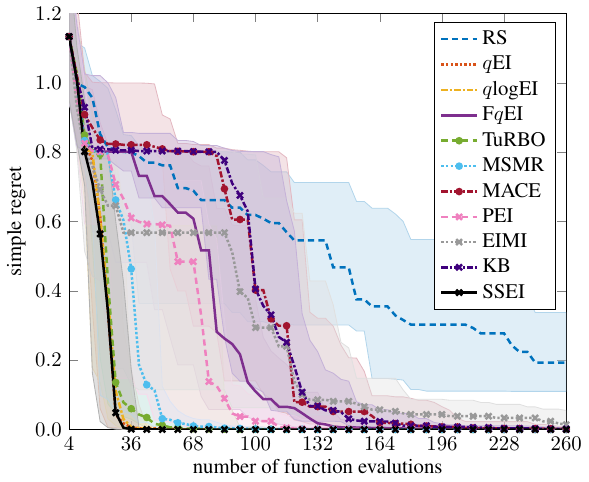}} \hfil
	\subfloat[4-D Michalewicz]{\includegraphics[width=0.33\linewidth]{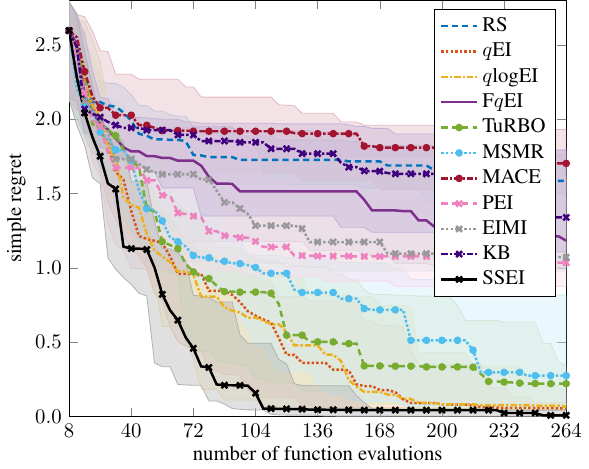}} \hfil
	\subfloat[8-D Michalewicz]{\includegraphics[width=0.33\linewidth]{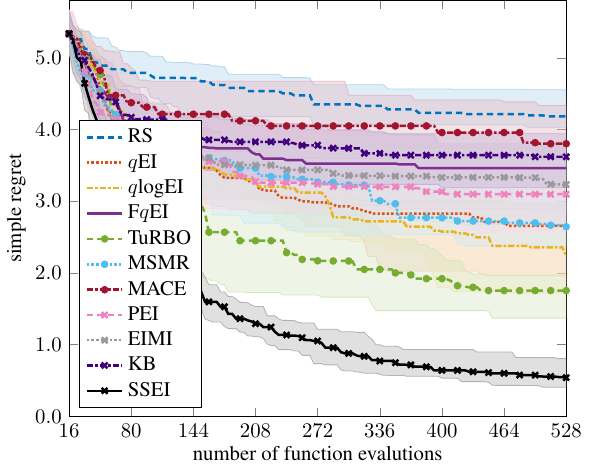}}   \\ 
	\subfloat[10-D Michalewicz]{\includegraphics[width=0.33\linewidth]{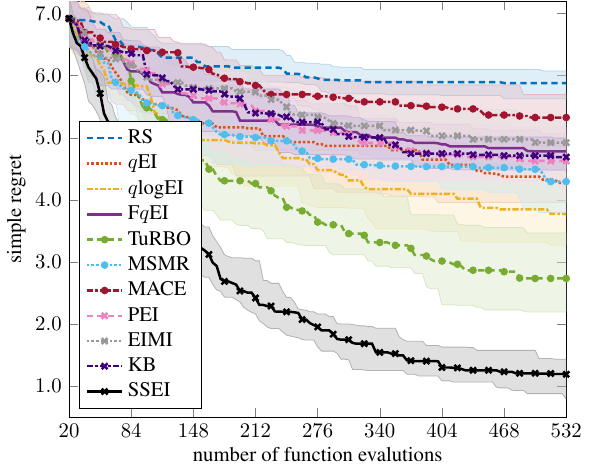}}\hfil
	\subfloat[20-D Michalewicz]{\includegraphics[width=0.33\linewidth]{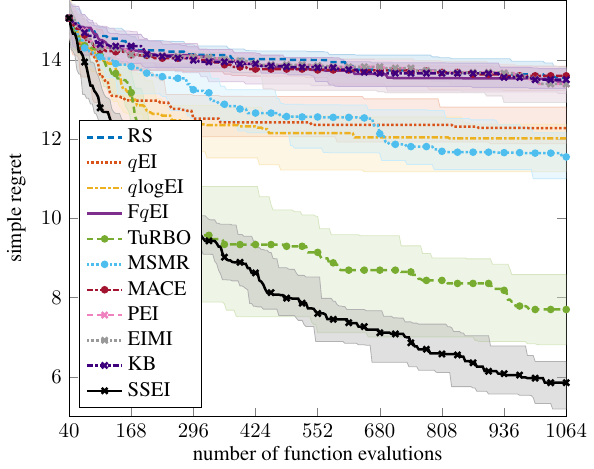}}\hfil
	\subfloat[30-D Michalewicz]{\includegraphics[width=0.33\linewidth]{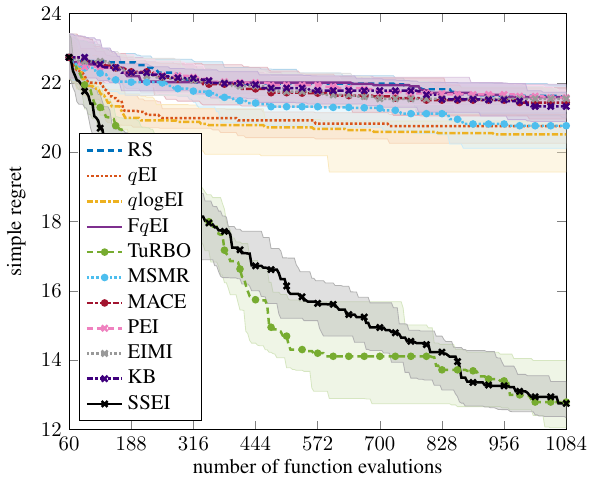}}\\
	\subfloat[2-D Schwefel]{\includegraphics[width=0.33\linewidth]{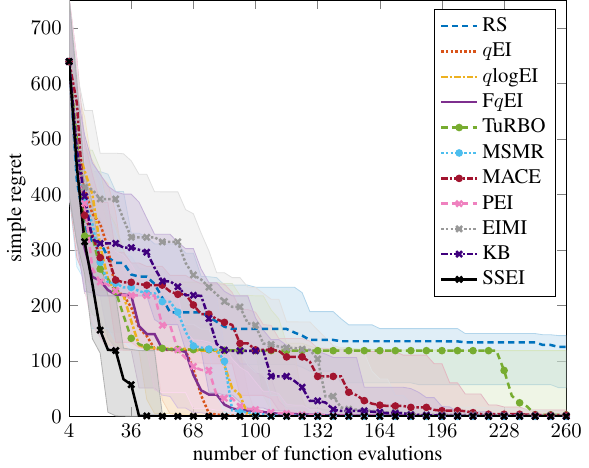}} \hfil
	\subfloat[4-D Schwefel]{\includegraphics[width=0.33\linewidth]{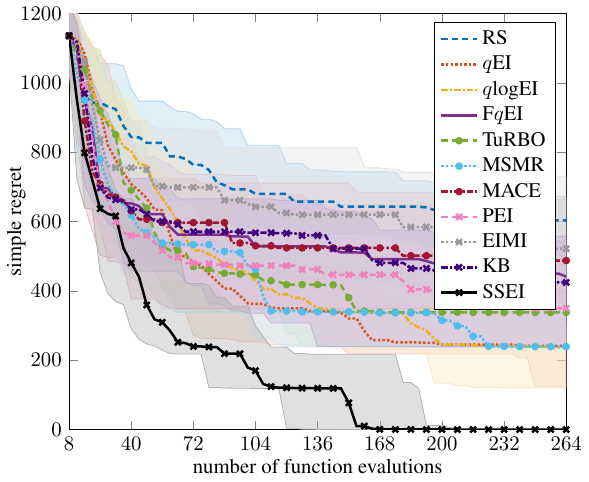}} \hfil
	\subfloat[8-D Schwefel]{\includegraphics[width=0.33\linewidth]{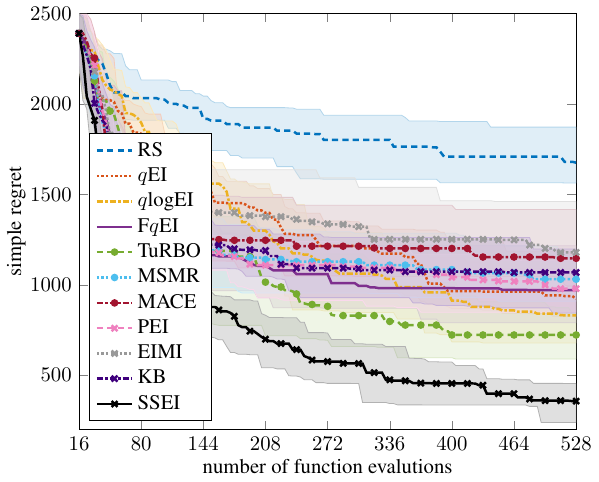}}   \\ 
	\subfloat[10-D Schwefel]{\includegraphics[width=0.33\linewidth]{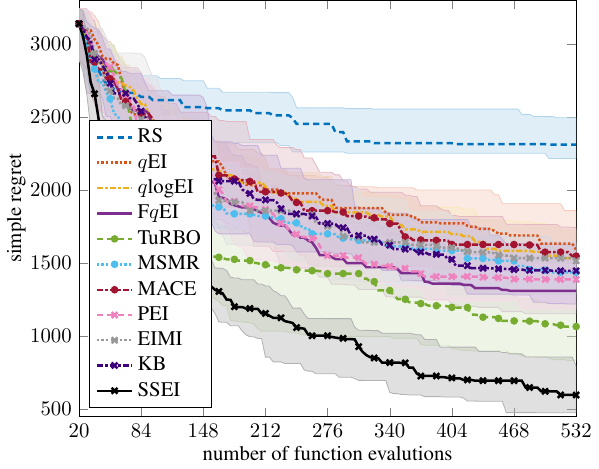}}\hfil
	\subfloat[20-D Schwefel]{\includegraphics[width=0.33\linewidth]{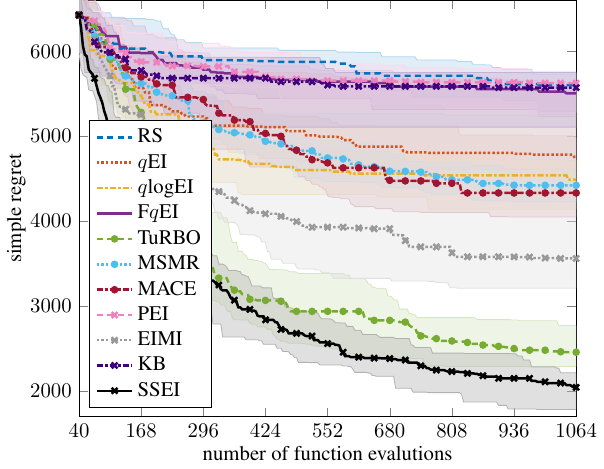}}\hfil
	\subfloat[30-D Schwefel]{\includegraphics[width=0.33\linewidth]{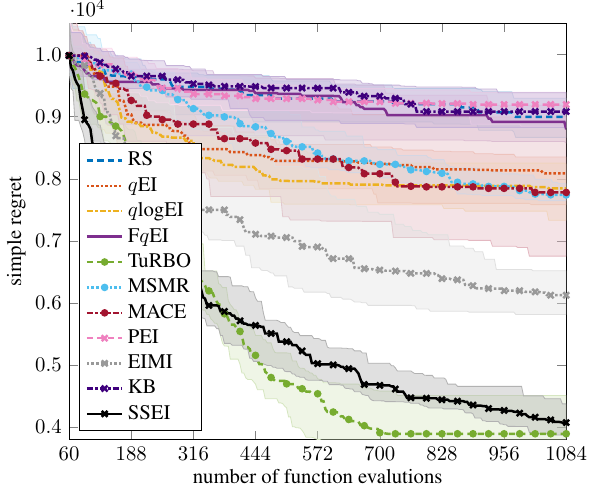}}\\
	\caption{Convergence histories of the compared batch approaches on Michalewicz and Schwefel problems with $q=4$.}
	\label{fig_batch_EI_q4}
\end{figure}

The convergence curves of the compared batch approaches on the test problems are plotted in Fig.~\ref{fig_batch_EI_q4} when a small batch size $q=4$ is used. We can first find that all the batch approaches outperform the random search approach on the shown problems. This means these batch approaches are more efficient than random guessing in selecting a small amount of query points. Compared with existing batch approaches, our proposed subspace acquisition approach performs very well on the shown problems across different dimensions. Specifically,  our proposed SSEI performs very competitively to the compared batch approaches on $2$-D and $4$-D problems, and outperforms the compared approaches and $8$-D and higher-dimensional problems.  The compared batch approaches select query points in the original or even higher-dimensional space, it becomes exponentially more difficult for them  to locate a  optimal solution when the dimension of the problem increases. In comparison, our proposed subspace acquisition function selects query points in lower-dimensional subspaces, which is often much less challenging to search than the original design space.  From the convergence curves, we can also find that the proposed SSEI and TuRBO generally performs better than the other batch approaches on high-dimensional problems. The SSEI introduces more local search by restricting the acquisition optimization in a part of dimensions while the TuRBO introduces more local search by utilizes local GP models. This indicates that introducing locality seems beneficial in high-dimensional spaces.

The simple regrets achieved by the compared batch approaches when $q=4$ is used are presented in Table~\ref{table_batch_q4} in Appendix~\ref{section_results}, where the mean values of $30$ runs are given. The results show that the proposed SSEI performs better than the compared batch approaches on most of the test problems across different dimensions. The $q$EI~\cite{Ginsbourger_2010}, $q$logEI~\cite{Ament_2023},  and F$q$EI~\cite{Zhan_2023} all measure the total expected improvement when a batch of points are considered. However, the dimensions of them are $d \times q$, which is $q$ times larger than the dimension of the original problem. Although we can use decomposition method to ease the curse of dimensionality~\cite{Zhan_2023}, optimizing these high-dimensional acquisition functions is still very challenging. The experiment results show that our proposed SSEI performs significantly better on fifty-two, forty-seven and forty-nine problems when compared with  the $q$EI, $q$logEI, and F$q$EI respectively. The TuRBO~\cite{Eriksson_2019} employs the trust region approach to build local Gaussian process models and uses the Thompson sampling approach to select query points. The experiment results show our proposed SSEI is able to achieve significantly better results on thirty-five problems. The MSMR approach~\cite{Joy_2020} trains multiple Gaussian process models and utilizes these different models to select different acquisition samples. The experiment results show that the proposed SSEI finds significantly better results than the MSMR on fifty-five test problems. This indicates that selecting from multiple subspaces is more efficient than selecting from multiple Gaussian process models.  The MACE approach~\cite{Lyu_2018} is a typical multi-objective approach. The results show that the proposed SSEI finds better results on forty-three problems than the MACE approach when $q=4$ is used. It seems that selecting acquisition samples from multiple subspaces is more promising than selecting from the Pareto front approximation.  Finally, the PEI~\cite{Zhan2017}, EIMI~\cite{LiZheng_2016} and KB~\cite{Ginsbourger_2010} are three penalty approaches. They select acquisition samples sequentially within one iteration and then evaluate them in parallel. The results show that our proposed SSEI performs significantly better than the PEI, EIMI and KB on fifty-one, fifty and fifty problems.  In summary, the proposed SSEI approach is able to find significantly better results on most of test problems when compared with different kinds of batch EI approaches as a small batch size $q=4$ in employed. The experiment results can empirically  prove the effectiveness of the proposed ESSI approach in small batch size setting.

\begin{figure}
	\centering
	\subfloat[2-D Michalewicz]{\includegraphics[width=0.33\linewidth]{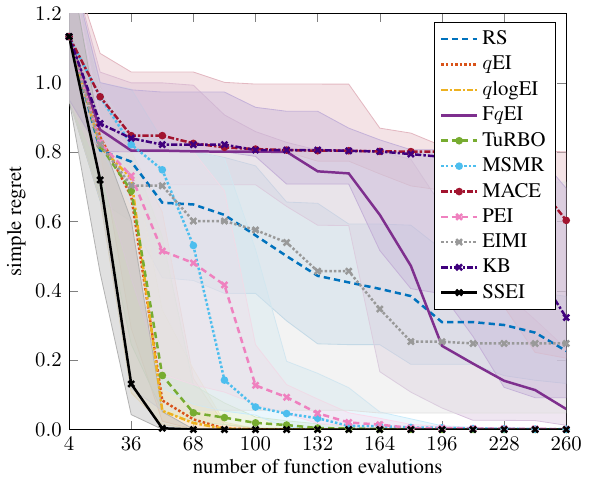}} \hfil
	\subfloat[4-D Michalewicz]{\includegraphics[width=0.33\linewidth]{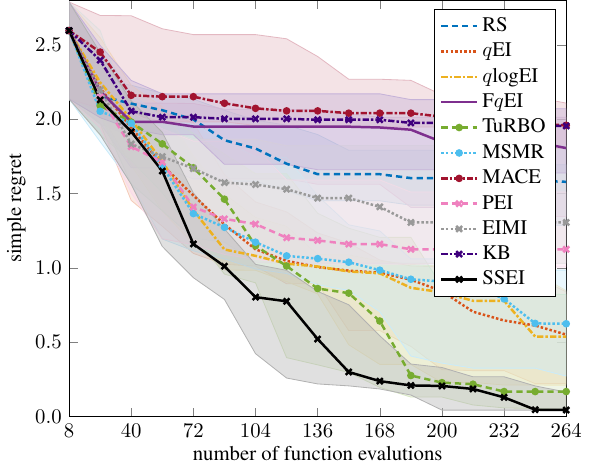}} \hfil
	\subfloat[8-D Michalewicz]{\includegraphics[width=0.33\linewidth]{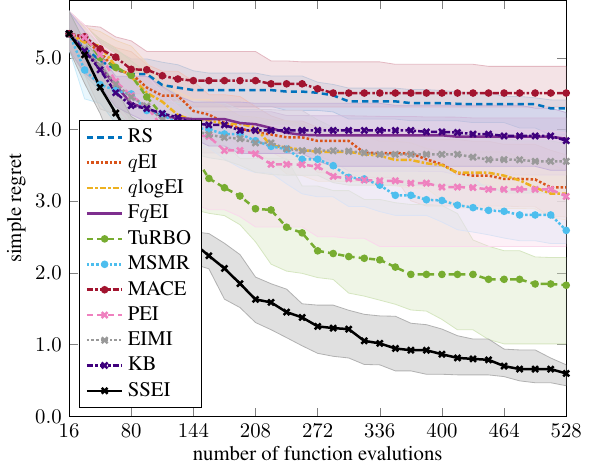}}   \\ 
	\subfloat[10-D Michalewicz]{\includegraphics[width=0.33\linewidth]{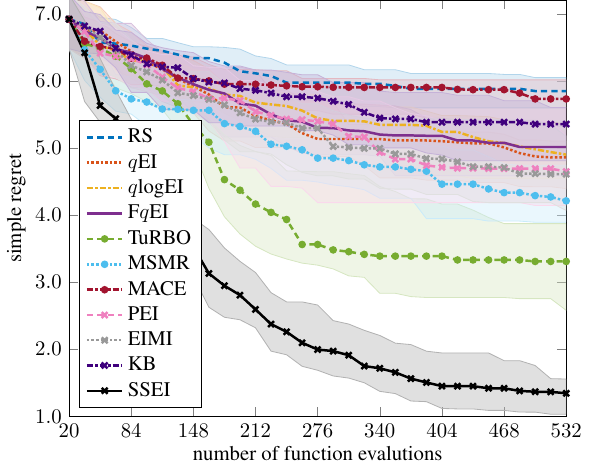}}\hfil
	\subfloat[20-D Michalewicz]{\includegraphics[width=0.33\linewidth]{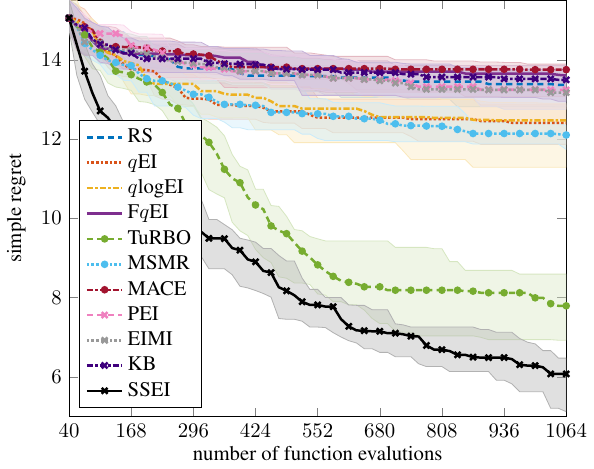}}\hfil
	\subfloat[30-D Michalewicz]{\includegraphics[width=0.33\linewidth]{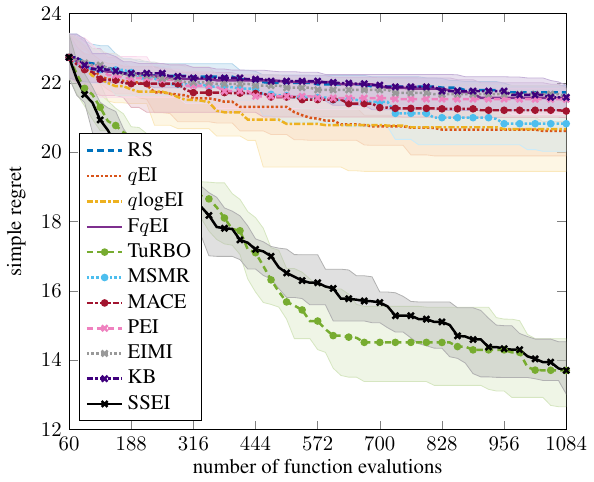}}\\
	\subfloat[2-D Schwefel]{\includegraphics[width=0.33\linewidth]{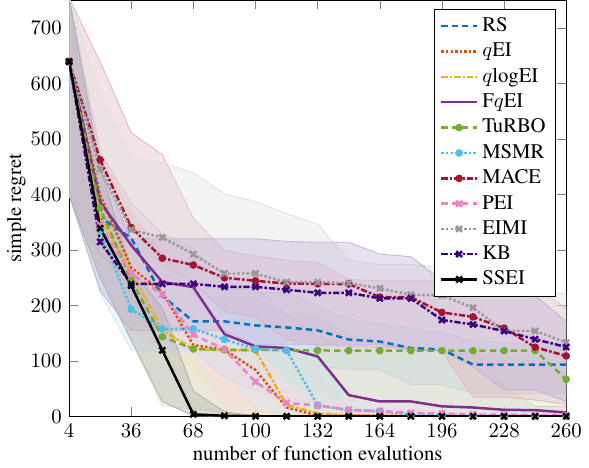}} \hfil
	\subfloat[4-D Schwefel]{\includegraphics[width=0.33\linewidth]{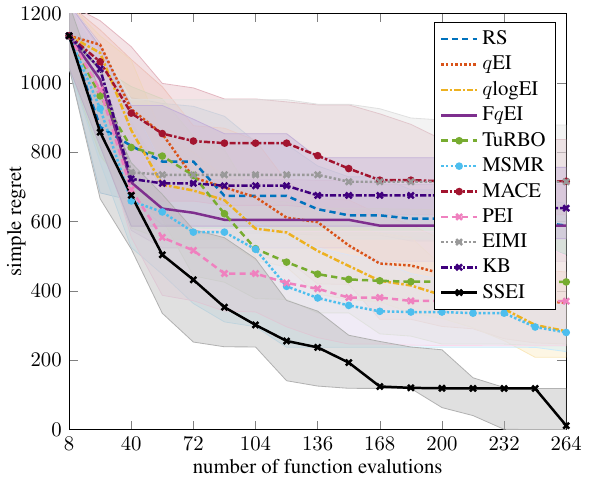}} \hfil
	\subfloat[8-D Schwefel]{\includegraphics[width=0.33\linewidth]{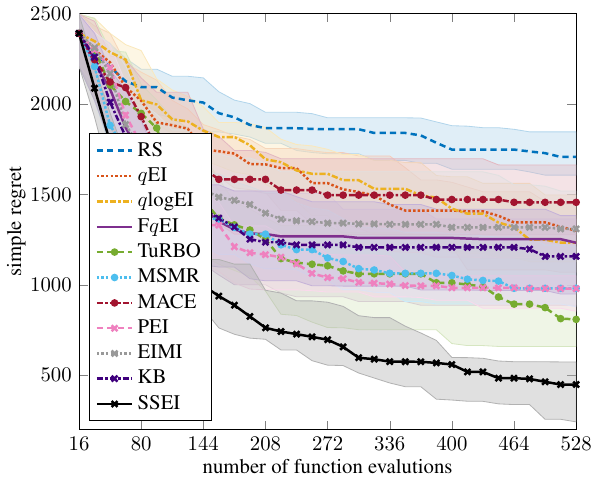}}   \\ 
	\subfloat[10-D Schwefel]{\includegraphics[width=0.33\linewidth]{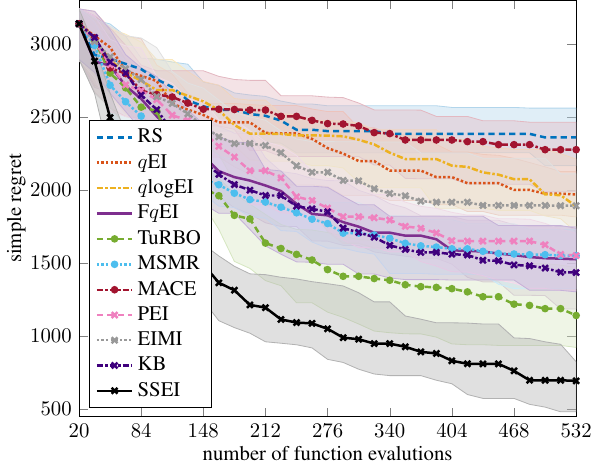}}\hfil
	\subfloat[20-D Schwefel]{\includegraphics[width=0.33\linewidth]{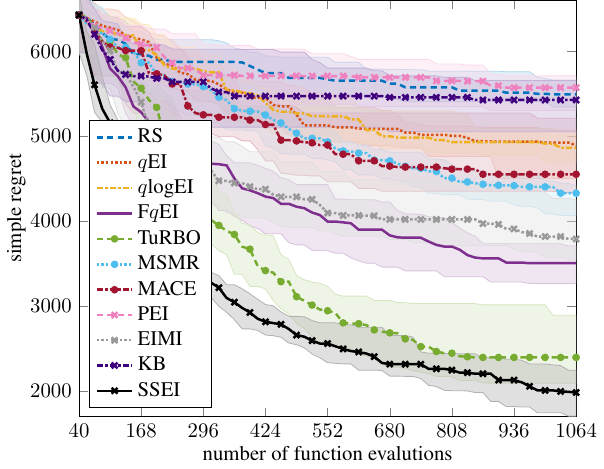}}\hfil
	\subfloat[30-D Schwefel]{\includegraphics[width=0.33\linewidth]{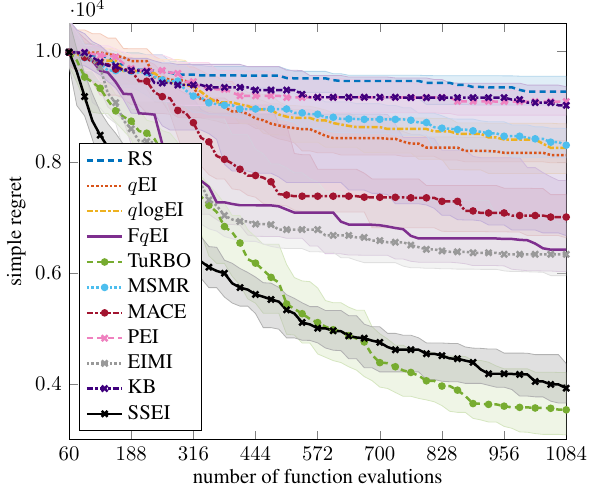}}\\
	\caption{Convergence histories of the compared batch approaches on Michalewicz and Schwefel problems with $q=16$.}
	\label{fig_batch_EI_q16}
\end{figure}

\begin{figure}
	\centering
	\subfloat[2-D Michalewicz]{\includegraphics[width=0.33\linewidth]{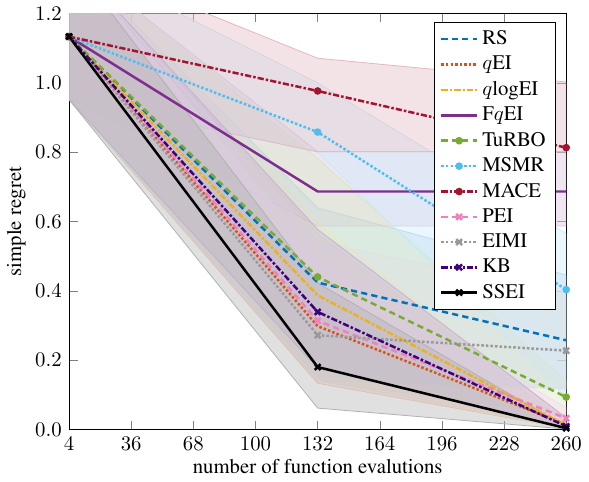}} \hfil
	\subfloat[4-D Michalewicz]{\includegraphics[width=0.33\linewidth]{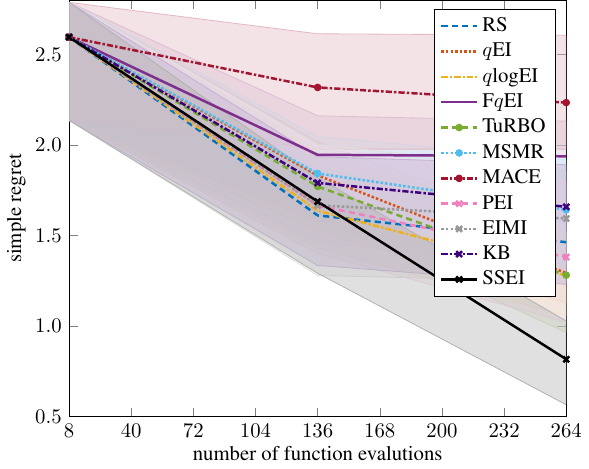}} \hfil
	\subfloat[8-D Michalewicz]{\includegraphics[width=0.33\linewidth]{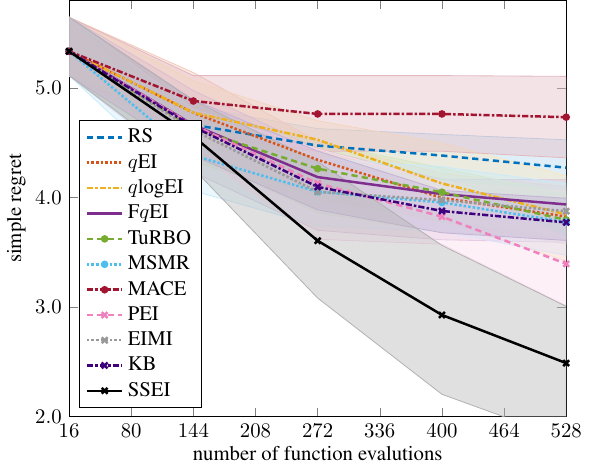}}   \\ 
	\subfloat[10-D Michalewicz]{\includegraphics[width=0.33\linewidth]{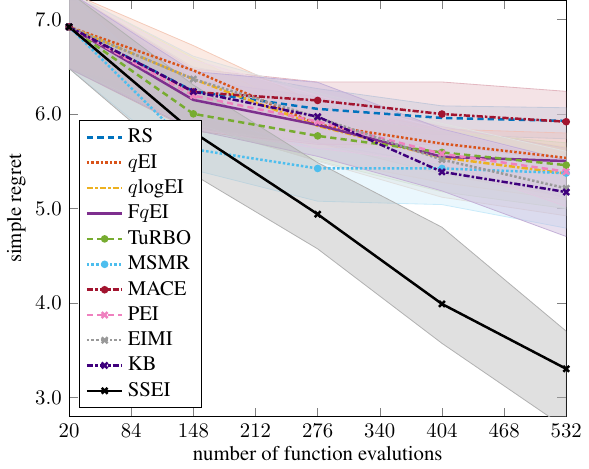}}\hfil
	\subfloat[20-D Michalewicz]{\includegraphics[width=0.33\linewidth]{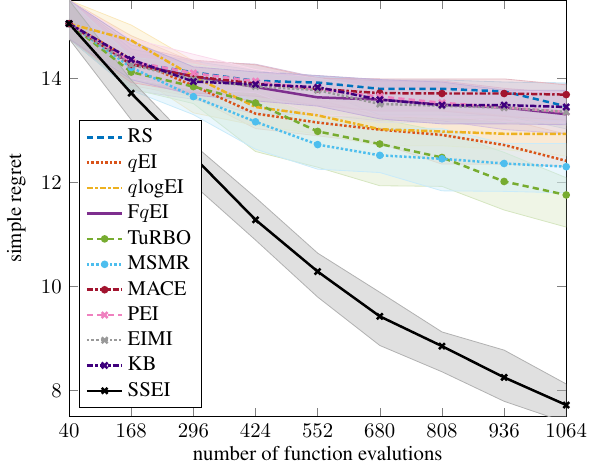}}\hfil
	\subfloat[30-D Michalewicz]{\includegraphics[width=0.33\linewidth]{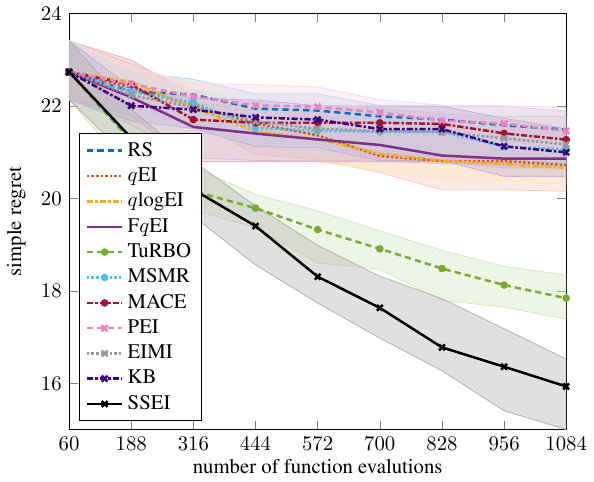}}\\
	\subfloat[2-D Schwefel]{\includegraphics[width=0.33\linewidth]{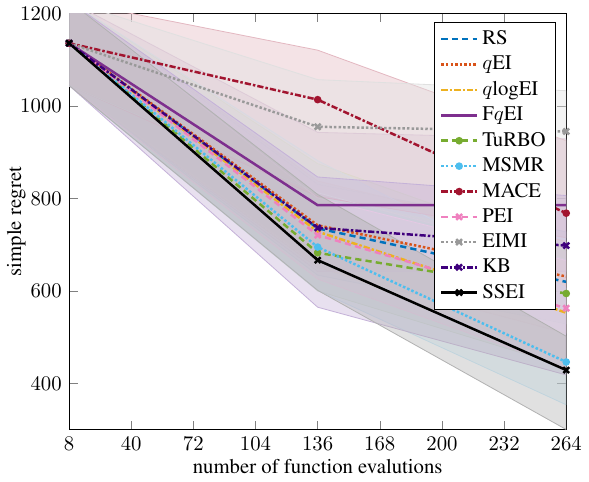}} \hfil
	\subfloat[4-D Schwefel]{\includegraphics[width=0.33\linewidth]{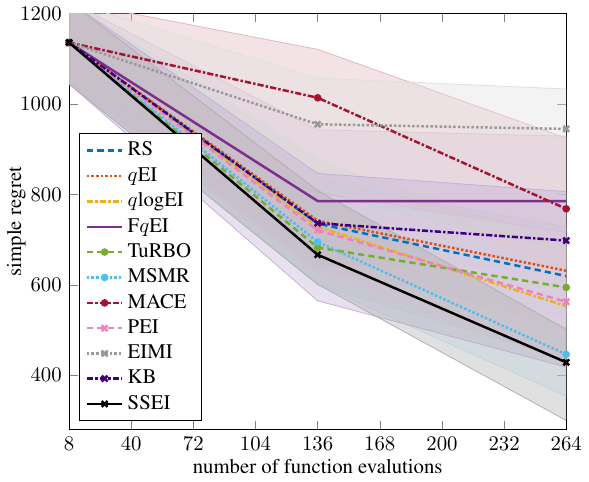}} \hfil
	\subfloat[8-D Schwefel]{\includegraphics[width=0.33\linewidth]{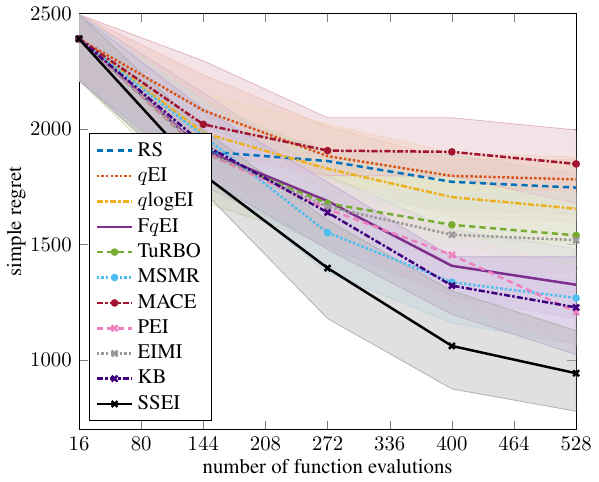}}   \\ 
	\subfloat[10-D Schwefel]{\includegraphics[width=0.33\linewidth]{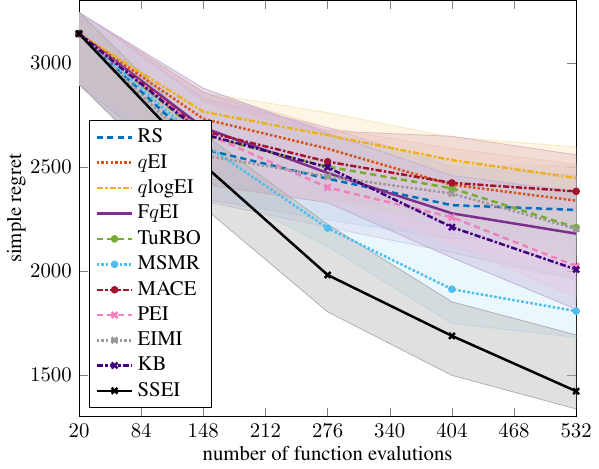}}\hfil
	\subfloat[20-D Schwefel]{\includegraphics[width=0.33\linewidth]{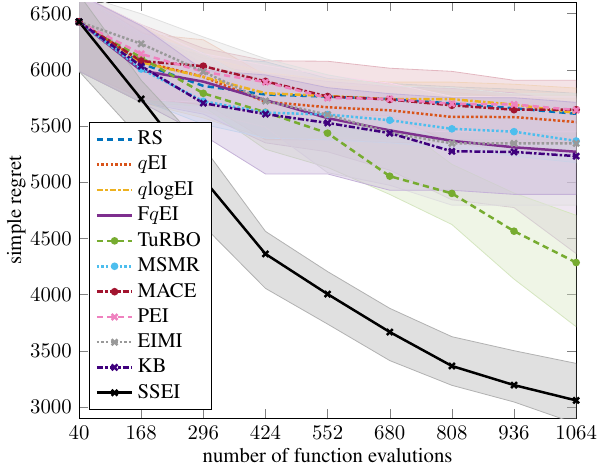}}\hfil
	\subfloat[30-D Schwefel]{\includegraphics[width=0.33\linewidth]{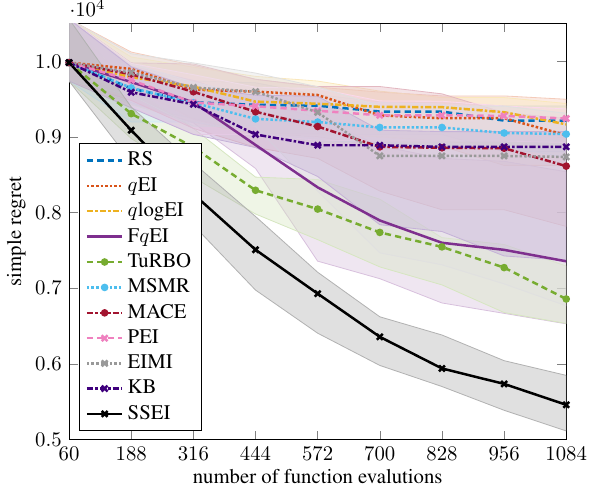}}\\
	\caption{Convergence histories of the compared batch approaches on Michalewicz and Schwefel problems with $q=128$.}
	\label{fig_batch_EI_q128}
\end{figure}

Then, we compare these batch EIs using a medium batch size $q=16$.  The convergence curves of the compared approaches on the  selected test problems are shown in Fig.~\ref{fig_batch_EI_q16} and the detailed experiment results of the compared batch approaches are given in Table~\ref{table_batch_q16} in Appendix~\ref{section_results}.    From the convergence curves, we can find the proposed SSEI approach is able to achieve faster convergence speed than most of the compared batch approaches. At the end of iterations, the proposed SSEI approach is also able to locate better solutions than the compared batch approaches. Also, we can find the advantage of our proposed SSEI grows as the dimension of the problem increases.  The final optimization results obtained by our proposed SSEI is either the best or among the best on all the test problems.  We can find from the table that our proposed SSEI is able to find significantly better results on most of the test problems than the ten compared batch approaches. Among the sixty test problems, the proposed SSEI approach is able to achieve significantly smaller regrets on sixty, fifty-two, forty-eight, forty-nine, thirty-five, fifty-one, forty-six, forty-seven, fifty-two and forty-seven of them when compared with the RS, $q$EI, $q$logEI, F$q$EI, TuRBO, MSMR, MACE, PEI, EIMI, and KB respectively.  The results indicate that our new idea of selecting a batch of query points in different subspaces is effective under a medium batch size.  The proposed SSEI approach shows very competitive performances on the test problems when the batch size $q=16$ is used.

Finally, we compare our proposed SSEI with the batch approaches under large batch size $q=128$ setting. The convergence curves are shown in Fig.~\ref{fig_batch_EI_q128} and the experiment results are given in Table~\ref{table_batch_q128} in Appendix~\ref{section_results}. From the convergence curves, we can see the proposed SSEI approach converges faster than most of the compared approaches when $q=128$ is used.  At the end of iterations, the proposed SSEI is able to locate better solutions than the compared batch approaches on most of the shown problems. On $2$-D, $4$-D and $8$-D problems, the proposed SSEI performs very competitively compared with the ten batch approaches. The advantage of the proposed SSEI becomes larger on $10$-D, $20$-D and $30$-D problems.   From Table~\ref{table_batch_q128} we can find the proposed SSEI is still able to performs very well on the test problems when a large batch size is employed. It outperforms the compared batch approaches on a majority of the test problems in term of the final optimization results.  Specifically, It performs better on sixty, forty-five, thirty-three, forty-two, forty-four, forty-one, fifty, thirty-six, forth-eight and forty-five problems compared with the RS, $q$EI, $q$logEI, F$q$EI, TuRBO, MSMR, MACE, PEI, EIMI and KB respectively. The results obtained by the proposed SSEI are either the best or among the best among all eleven batch approaches. The experiment results indicate that the idea of selecting query points from multiple subspaces is promising when a large batch size is required. 

\begin{figure}
	\centering
	\subfloat[2D Michalewicz]{\includegraphics[width=0.33\linewidth]{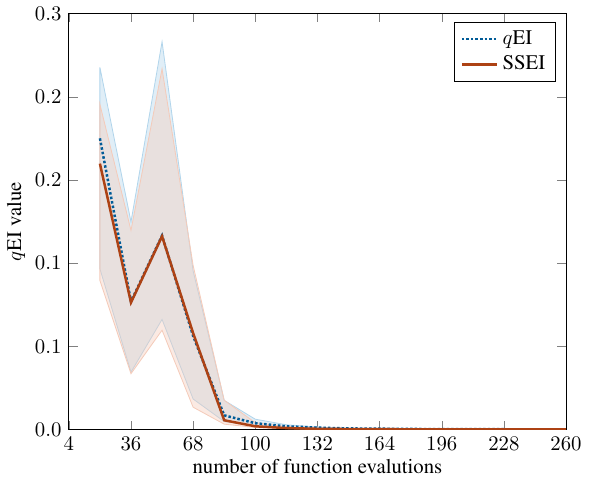}}\hfil
	\subfloat[4D Michalewicz]{\includegraphics[width=0.33\linewidth]{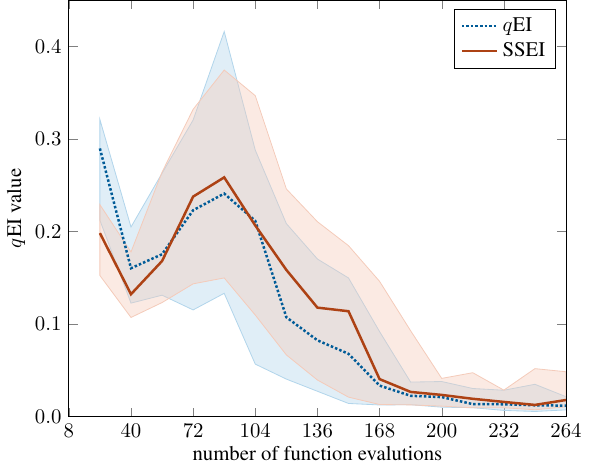}}\hfil
	\subfloat[8D Michalewicz]{\includegraphics[width=0.33\linewidth]{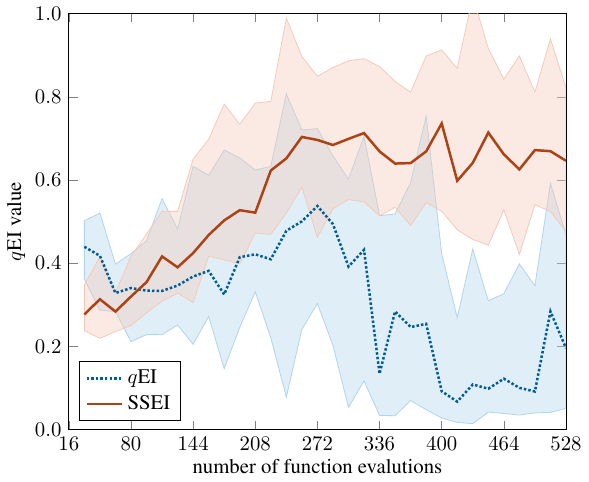}}\\ 
	\subfloat[10D Michalewicz]{\includegraphics[width=0.33\linewidth]{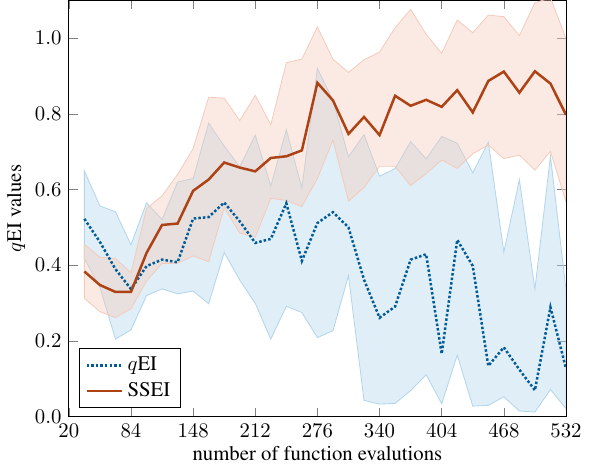}}\hfil
	\subfloat[20D Michalewicz]{\includegraphics[width=0.33\linewidth]{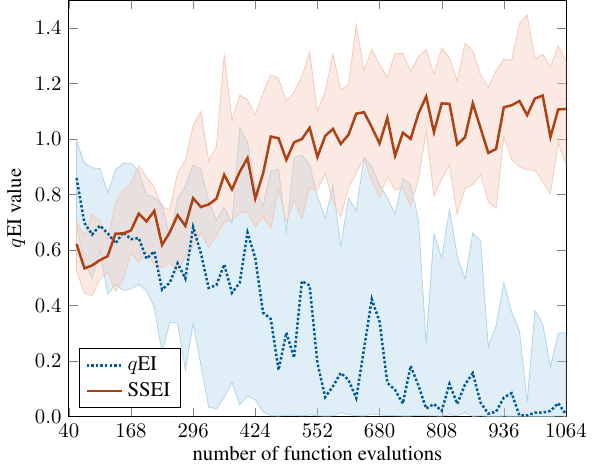}}\hfil
	\subfloat[30D Michalewicz]{\includegraphics[width=0.33\linewidth]{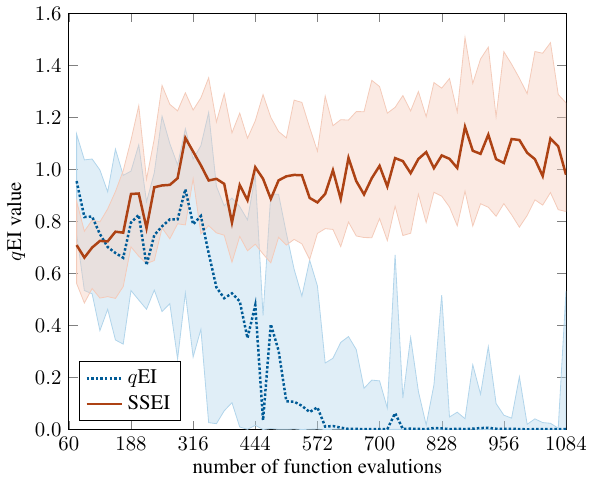}}\\ 
	\subfloat[2D Schwefel]{\includegraphics[width=0.33\linewidth]{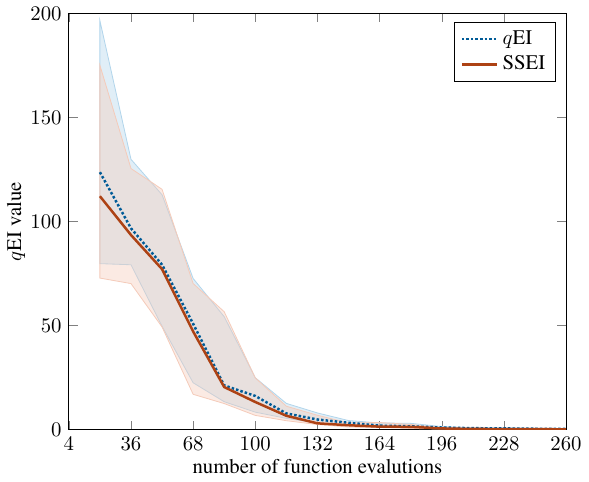}}\hfil
	\subfloat[4D Schwefel]{\includegraphics[width=0.33\linewidth]{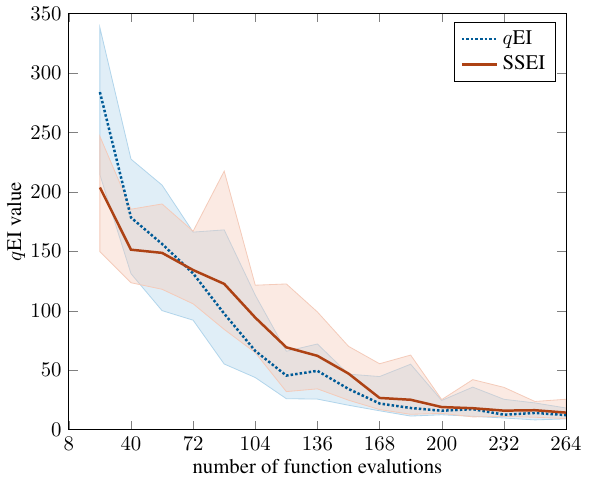}}\hfil
	\subfloat[8D Schwefel]{\includegraphics[width=0.33\linewidth]{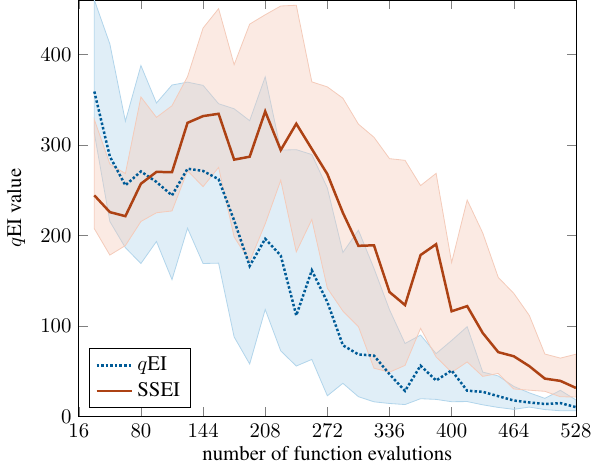}}\\ 
	\subfloat[10D Schwefel]{\includegraphics[width=0.33\linewidth]{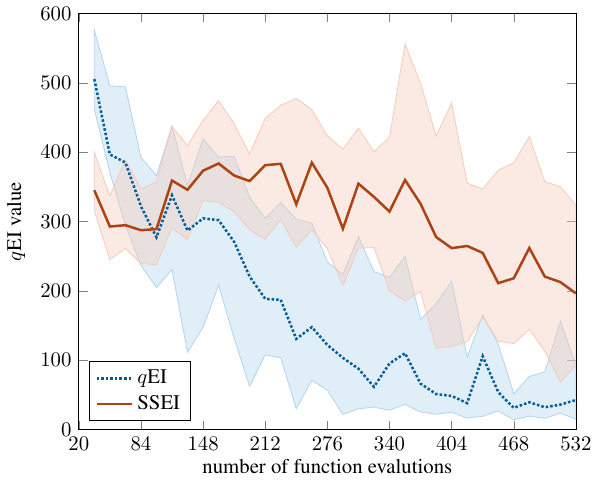}}\hfil
	\subfloat[20D Schwefel]{\includegraphics[width=0.33\linewidth]{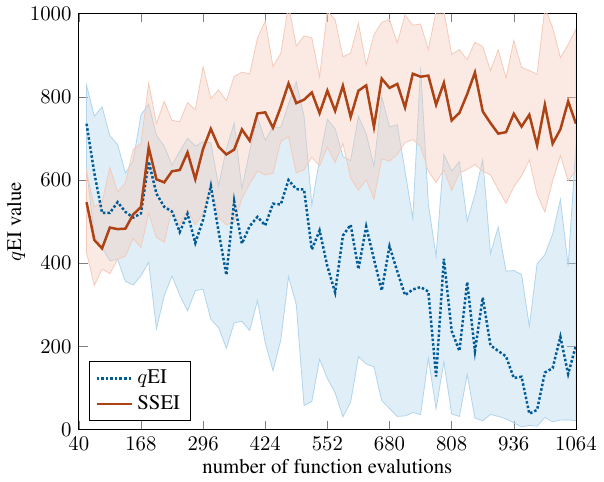}}\hfil
	\subfloat[30D Schwefel]{\includegraphics[width=0.33\linewidth]{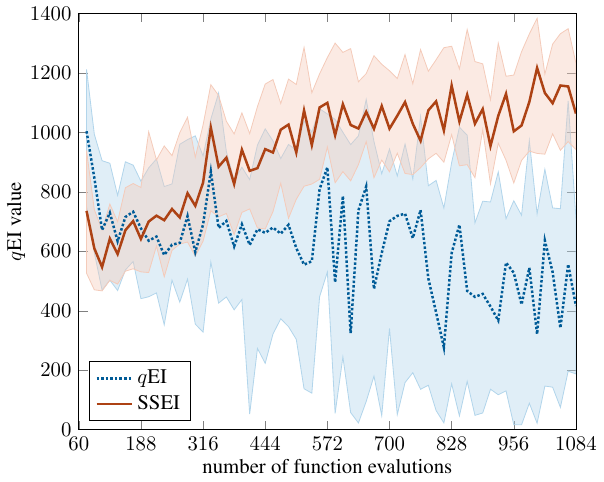}}\\ 
	\caption{$q$EI values obtained by the SSEI acquisition function and the $q$EI  acquisition function along the iterations.}
	\label{fig_qEI_record}
\end{figure}

To further investigate why the proposed subspace acquisition function performs better than the batch acquisition function, we collect the $q$EI values that the SSEI and the $q$EI obtained along the iterations. The results are shown in Fig.~\ref{fig_qEI_record}, where the median, the first quartile and the third quartile of $30$ runs are shown. From the figures, we can find very similar trends in the single-point acquisition function. On 2-D and 4-D problems, our subspace acquisition function and the batch acquisition function finds very similar $q$EI values along the iterations.  On 8-D and higher-dimensional problems, the batch acquisition function finds higher $q$EI values at the beginning of iterations, but lower $q$EI values at the middle and end of iterations compared with our subspace acquisition function. This might be the reason why our subspace acquisition function performs better than the batch acquisition function on high-dimensional problems.

In summary, the experiment results show that the proposed SSEI approach is able to outperform the ten state-of-the-art batch approaches on most of the test problems in small, medium and large batch sizes. These experiment results empirically  prove the effectiveness of our proposed SSEI approach and demonstrate its good scalability with both the problem dimension and the batch size.

\subsection{Comparison with Entropy-based Acquisition Functions}

Entropy-based Acquisition functions have been actively studied in Bayesian optimization~\cite{Hennig_2012,Hernandez_2014,WangZi_2017,Moss_2021}. We compare our proposed subspace acquisition function with two entropy-based acquisition functions: the MES (Max-value Entropy Search)~\cite{WangZi_2017} and GIBBON (General-purpose Information-Based Bayesian OptimisatioN)~\cite{Moss_2021} in this subsection. We also use the BoTorch implementation of the MES~\footnote{\url{https://botorch.org/docs/tutorials/max_value_entropy/}} and GIBBON~\footnote{\url{https://botorch.org/docs/tutorials/GIBBON_for_efficient_batch_entropy_search/}} in the comparison. We use the same Gaussian process model and acquisition optimizer for the three compared methods. The only difference among them lies in the acquisition functions. 

\begin{figure}
	\centering
	\subfloat[2-D Michalewicz]{\includegraphics[width=0.33\linewidth]{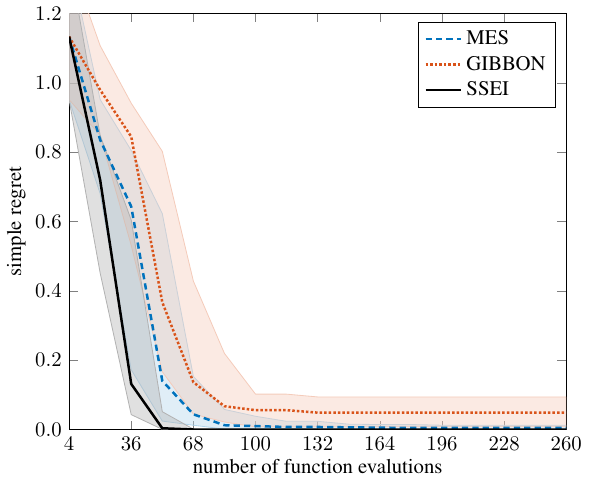}} \hfil
	\subfloat[4-D Michalewicz]{\includegraphics[width=0.33\linewidth]{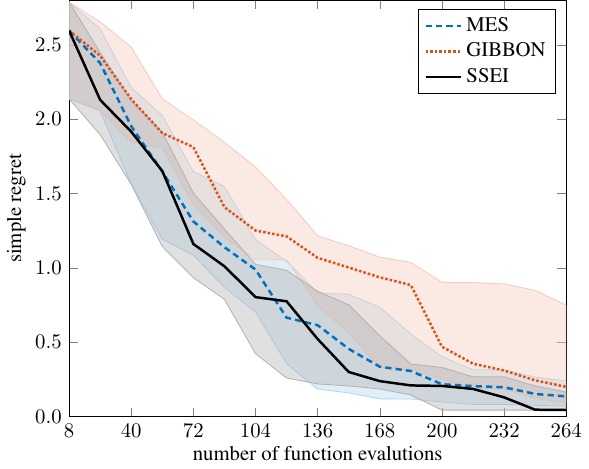}} \hfil
	\subfloat[8-D Michalewicz]{\includegraphics[width=0.33\linewidth]{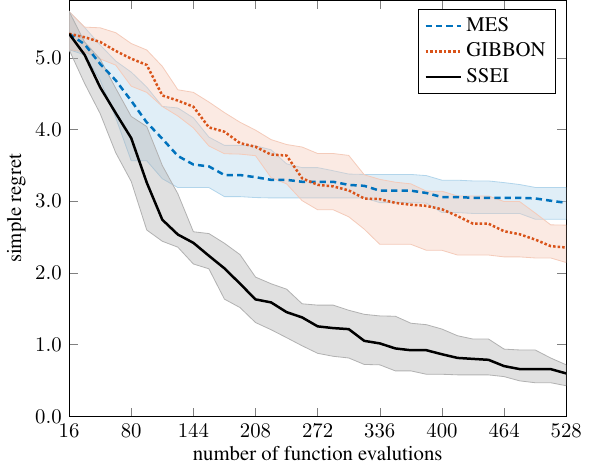}}   \\ 
	\subfloat[10-D Michalewicz]{\includegraphics[width=0.33\linewidth]{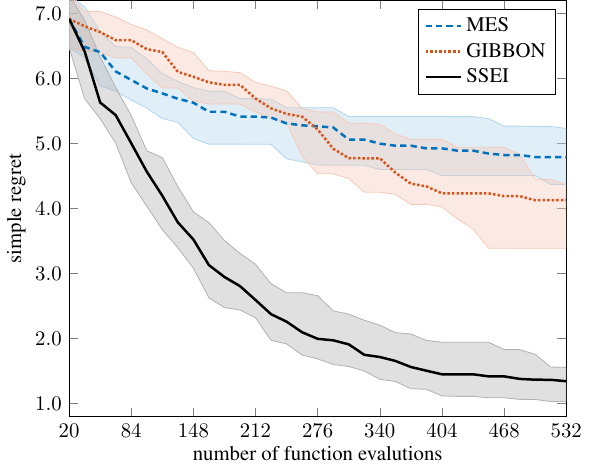}}\hfil
	\subfloat[20-D Michalewicz]{\includegraphics[width=0.33\linewidth]{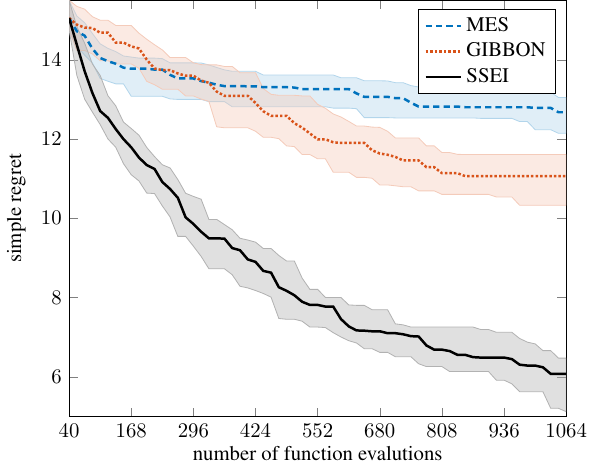}}\hfil
	\subfloat[30-D Michalewicz]{\includegraphics[width=0.33\linewidth]{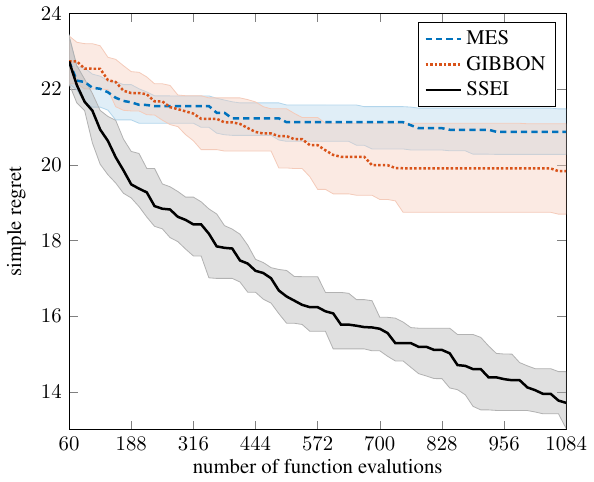}}\\
	\subfloat[2-D Schwefel]{\includegraphics[width=0.33\linewidth]{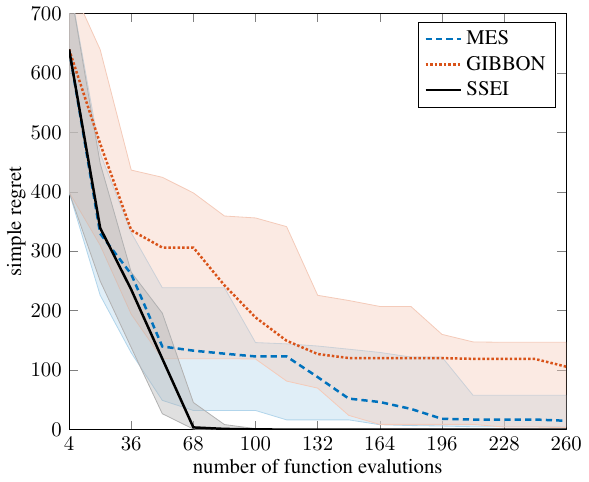}} \hfil
	\subfloat[4-D Schwefel]{\includegraphics[width=0.33\linewidth]{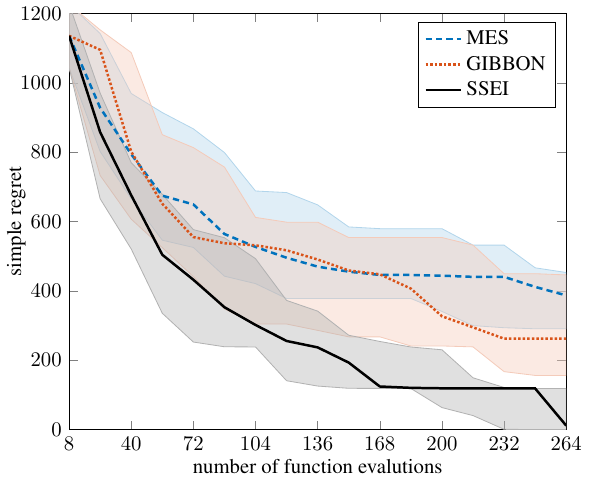}} \hfil
	\subfloat[8-D Schwefel]{\includegraphics[width=0.33\linewidth]{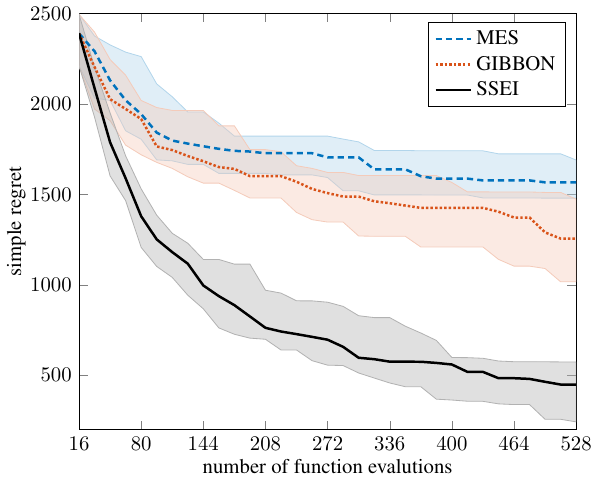}}   \\ 
	\subfloat[10-D Schwefel]{\includegraphics[width=0.33\linewidth]{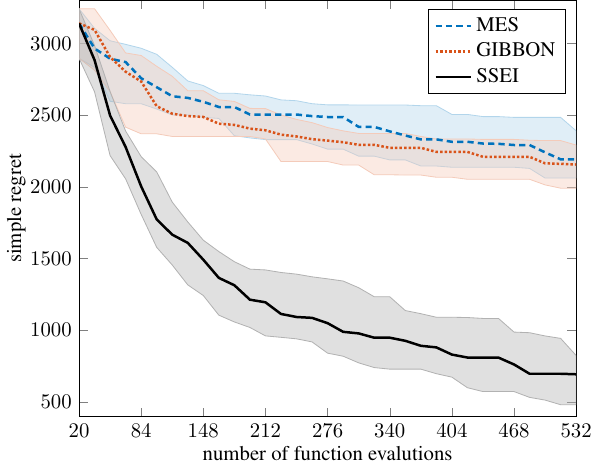}}\hfil
	\subfloat[20-D Schwefel]{\includegraphics[width=0.33\linewidth]{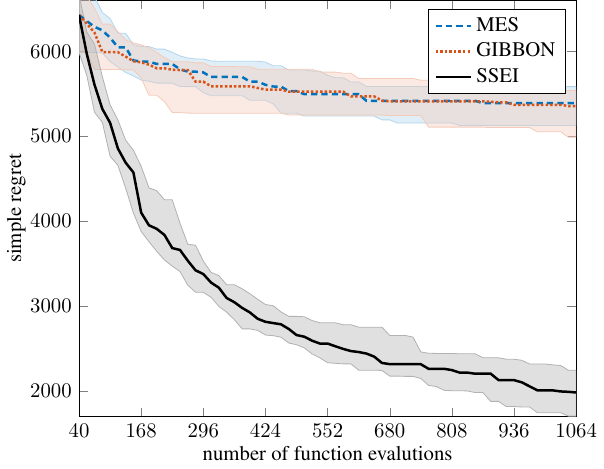}}\hfil
	\subfloat[30-D Schwefel]{\includegraphics[width=0.33\linewidth]{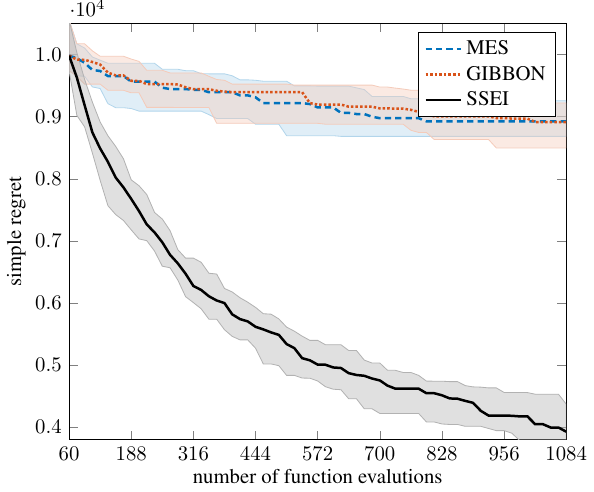}}\\
	\caption{Convergence histories of the compared MES, GIBBON and SSEI on Michalewicz and Schwefel problems with $q=16$.}
	\label{fig_entropy_q16}
\end{figure}

The convergence curves of the three compared algorithms on the on Michalewicz and Schwefel problems with $q=16$ are plotted in Fig.~\ref{fig_entropy_q16}, where  the median, the first quartile, and the third quartile of 30 runs are shown. We can see from the figures that all the three acquisition functions can find solutions close to the global optimum on $2$-D and $4$-D problems. Compared with the two entropy-based acquisition functions, the proposed SSEI converges slightly faster. Compared with MES, the GIBBON converges  slightly slower and finds slightly worse results at the end of iterations on the $2$-D and $4$-D problems. As the dimension of the problem increases, the advantage of the proposed SSEI over the two entropy-based acquisition functions becomes clearer. On $8$-D and higher-dimensional problems, the proposed SSEI converges significantly faster than the MES and GIBBON approaches, and finds significantly better optimization results at the end of iterations. Compared with MES, the GIBBON locates better optimization results on $8$-D and high-dimensional problems.  The experiment results show the advantage of our proposed SSEI acquisition function over the two entropy-based acquisition functions, especially on high-dimensional problems.

\subsection{Applying the Subspace Approach to Other Acquisition Functions}
\label{section_SSPI}
\begin{figure}
	\centering
	\subfloat[2-D Michalewicz]{\includegraphics[width=0.33\linewidth]{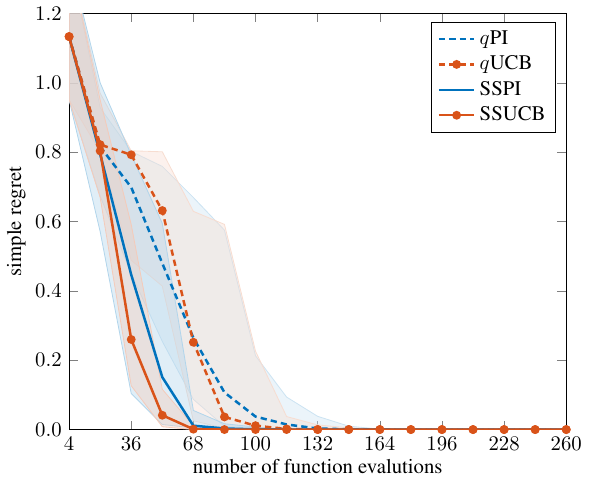}} \hfil
	\subfloat[4-D Michalewicz]{\includegraphics[width=0.33\linewidth]{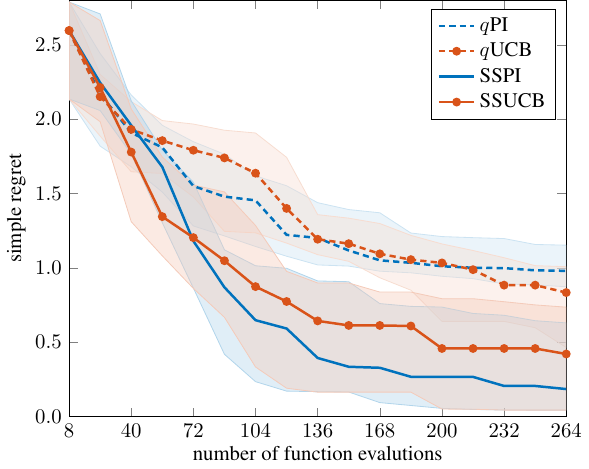}} \hfil
	\subfloat[8-D Michalewicz]{\includegraphics[width=0.33\linewidth]{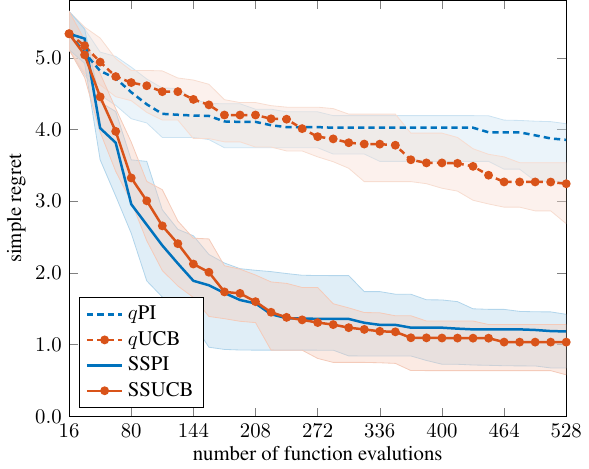}}   \\ 
	\subfloat[10-D Michalewicz]{\includegraphics[width=0.33\linewidth]{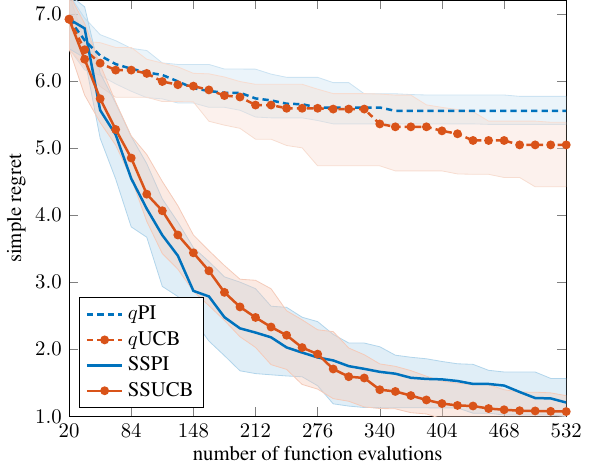}}\hfil
	\subfloat[20-D Michalewicz]{\includegraphics[width=0.33\linewidth]{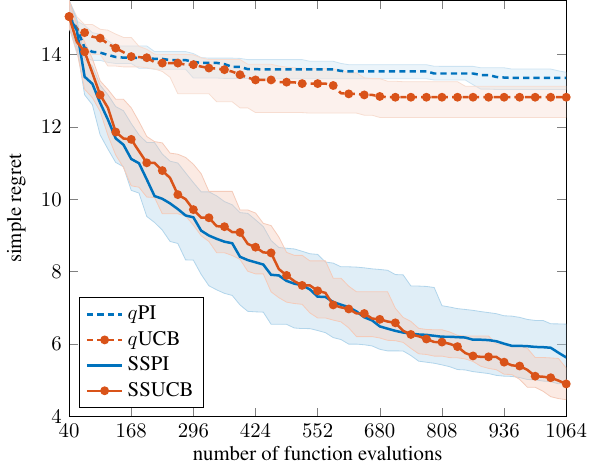}}\hfil
	\subfloat[30-D Michalewicz]{\includegraphics[width=0.33\linewidth]{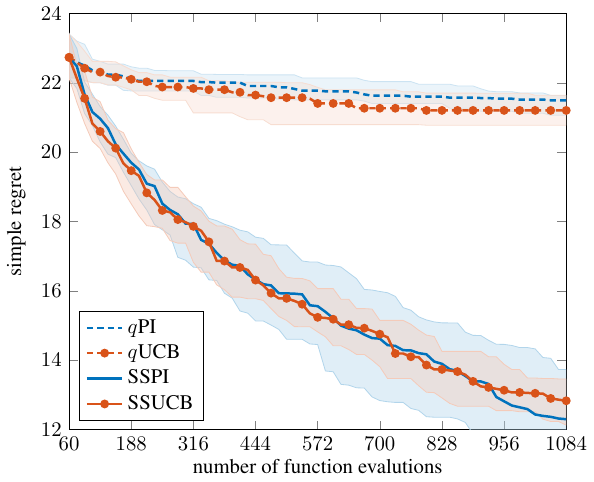}}\\
	\subfloat[2-D Schwefel]{\includegraphics[width=0.33\linewidth]{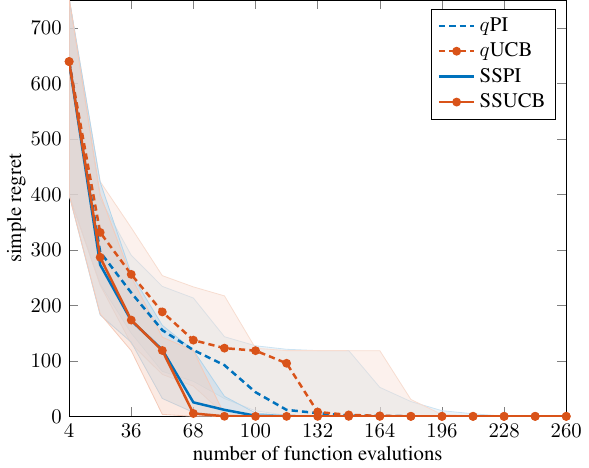}} \hfil
	\subfloat[4-D Schwefel]{\includegraphics[width=0.33\linewidth]{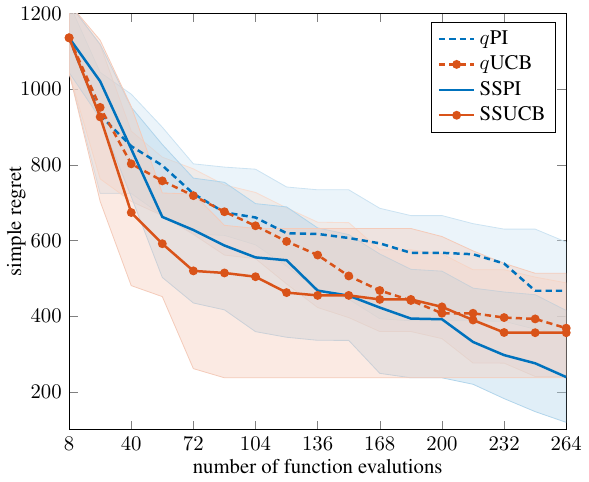}} \hfil
	\subfloat[8-D Schwefel]{\includegraphics[width=0.33\linewidth]{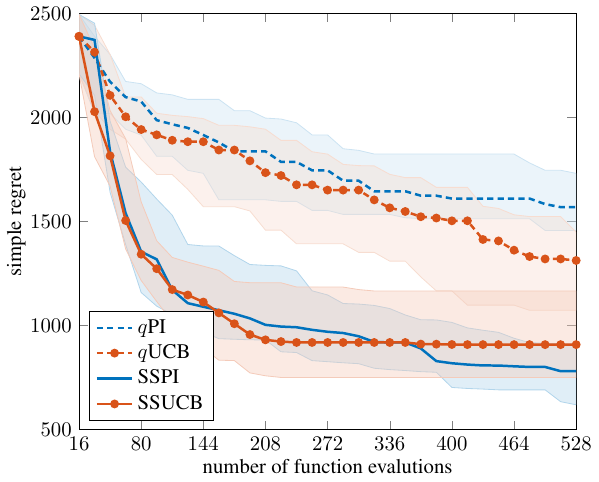}}   \\ 
	\subfloat[10-D Schwefel]{\includegraphics[width=0.33\linewidth]{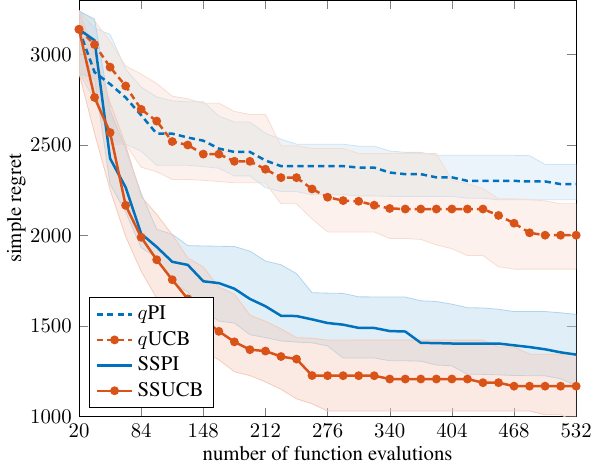}}\hfil
	\subfloat[20-D Schwefel]{\includegraphics[width=0.33\linewidth]{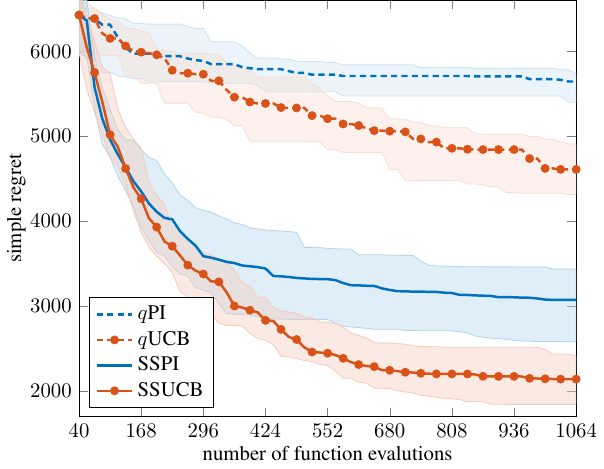}}\hfil
	\subfloat[30-D Schwefel]{\includegraphics[width=0.33\linewidth]{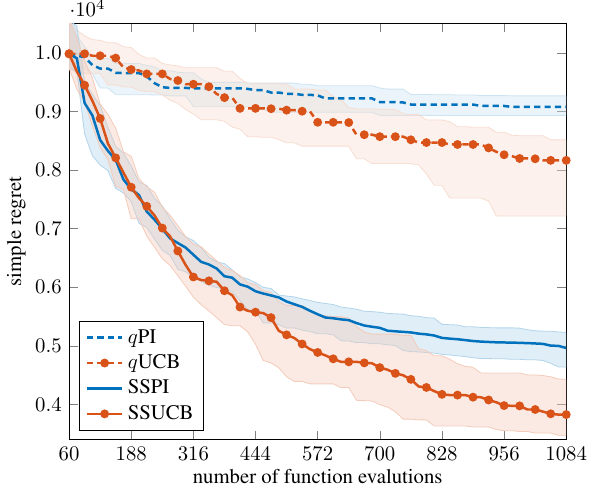}}\\
	\caption{ Convergence histories of $q$PI, $q$UCB, SSPI and SSUCB with $q=16$.}
	\label{fig_other_acquisition_q16}
\end{figure}

Besides the EI function, our proposed subspace acquisition approach can be easily applied to other acquisition functions . In this subsection, we apply the subspace acquisition approach to the PI (Probability of Improvement) and UCB (Upper Confidence Bound) acquisition functions. The subspace versions of the PI and UCB functions are called SSPI (SubSpace Probability of Improvement) and SSUCB (SubSpace Upper Confidence Bound) in this work. We compare the SSPI and SSUCB with their multi-point counterparts, which are the $q$PI and $q$UCB. All the compared acquisition functions are implemented in BoTorch. The Gaussian process model and acquisition optimizer are the same for the compared acquisition functions. The batch size is set to $q=16$ in the comparison. The BoTorch implementations of our SSPI and SSUCB are also available at \url{https://github.com/zhandawei/SubSpace_Acquisition_Functions}.

The convergence curves of the four compared acquisition functions are given in Fig.~\ref{fig_other_acquisition_q16}, where the median, the first quartile, and the third quartile of 30 runs are plotted. From the figures, we can see that the subspace versions of PI and UCB outperforms their multi-point counterparts on both the Michalewicz and Schwefel problems across different dimensions.  On $2$-D and $4$-D problems, the differences between the subspace acquisition functions and the multi-point acquisition functions are not that significant. The subspace acquisition functions converge slightly faster than the multi-point counterparts and they find similar optimization results at the end of iterations. On $8$-D and higher-dimensional problems, the SSPI and SSUCB converges significantly faster and finds much better results in the end compared with the $q$PI and $q$UCB acquisition functions.  The $q$PI and $q$UCB search in the $(d \times q)$-dimensional space to locate a batch of query points, which is very challenging when $d$ and $q$ is large. In comparison, the proposed SSPI and SSUCB search in $q$ different subspaces to locate $q$ query points, which is often much less challenging than searching the $(d \times q)$-dimensional space. The experiment results show that searching in axis-aligned subspaces can improve the optimization efficiency for the PI and UCB acquisition functions than searching in  $(d \times q)$-dimensional space. This also indicates that the proposed subspace acquisition approach can be well applied to the PI and UCB acquisition functions.

\section{Conclusions}
\label{section_conclusion}
Extending Bayesian optimization to parallel computing is an interesting and meaningful task. In this work, we propose a novel idea to develop batch Bayesian optimization algorithms. Different from existing methods, the proposed method selects multiple query points in multiple different subspaces. The proposed approach has no additional parameter,  is easy to understand and simple to implement. The proposed approach has shown very competitive performances in optimization efficiency when compared with the standard Bayesian optimization and ten well-established batch  approaches through numerical experiments. We also show that the proposed approach can be well applied to different acquisition functions.

However, there are also limitations of our work. Here, we address one limitation for future improvement. In this work, we employed the random approach to select a batch of subspaces  for simplicity, but did not consider the diversity of the selected subspaces.  Using more complex heuristic approach to select the subspaces, for example using the Hamming distance to  measure the similarity of different subspaces, might be a way to improve the diversity of the selected subspaces.  However, this may increase the computational complexity for selecting the subspaces. Therefore, designing a better subspace selection strategy that takes good trade-off between optimization efficiency and computational cost is of great interest.




\newpage
\appendix
\section{Test Problems and Experiment Results}
\subsection{Test Problems}
\label{section_test_problems}
\begin{enumerate}
	\item The first test problem is the GPMean function. This function is sampled from a Gaussian process, therefore it can eliminate model misspecification and allow a direct comparison of different acquisition functions. Following~\cite{Hennig_2012,Hernandez_2014}, we randomly select 1000  samples in the domain $\mathcal{X} = [0,1]^d$, and draw their function values jointly from a Gaussian process with zero mean and a squared exponential kernel with length scale $l=0.1$.  The resulting posterior mean is then used as the test function. A two-dimensional GP mean function is illustrated  in Fig.~\ref{fig_GPMean}.  Since the Gaussian process mean functions are randomly generated, their global minima are unknown. To calculate the simple regret of the compared algorithms on these functions, we employ a real-coded genetic algorithm to find their approximate global minima using $10000d$ function evaluations and $100$ repetitions.
	\begin{figure}[h]
		\centering
		\includegraphics[width=0.36\linewidth]{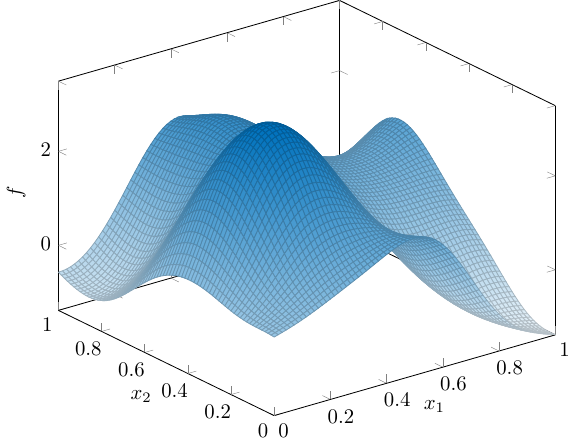}
		\caption{Landscape plot of the GPMean function for $d=2$.}
		\label{fig_GPMean}
	\end{figure}
	
	\item The second test problem is the SumSquares function, whose expression is 
	\begin{equation*}
		f(\bm{x}) = \sum_{i=1}^{d} i x_i^2.
	\end{equation*}
	The search domain of this problem is $\mathcal{X} = [-5.12,5.12]^d$. The landscape of this function for $d=2$ is illustrated in Fig.~\ref{fig_SumSquares}. As can be seen, the SumSquares function is continuous, convex and unimodal.  
	\begin{figure}[h]
		\centering
		\includegraphics[width=0.36\linewidth]{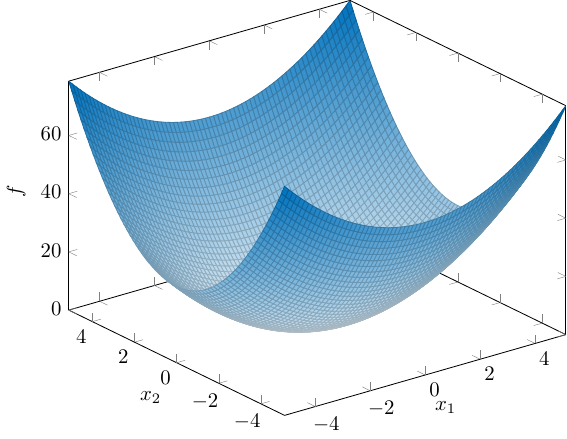}
		\caption{Landscape plot of the SumSquares function for $d=2$.}
		\label{fig_SumSquares}
	\end{figure}
	\item The third selected test problem is the Rosenbrock function, whose formula is 
	\begin{equation*}
		f(\bm{x}) = \sum_{i=1}^{d-1} \left[ 100\left(x_{i+1} - x_i^2\right)^2 + \left(x_i-1\right)^2\right].
	\end{equation*}
	We optimize this function in the domain of $\mathcal{X} = [-5,10]^d$. The landscape of the Rosenbrock function for $d=2$ is illustrated in Fig.~\ref{fig_Rosenbrock}. Although this function is unimodal, its global optimum lies in a narrow valley, which makes the global minimum difficult to find.  
	\begin{figure}[h]
		\centering
		\includegraphics[width=0.36\linewidth]{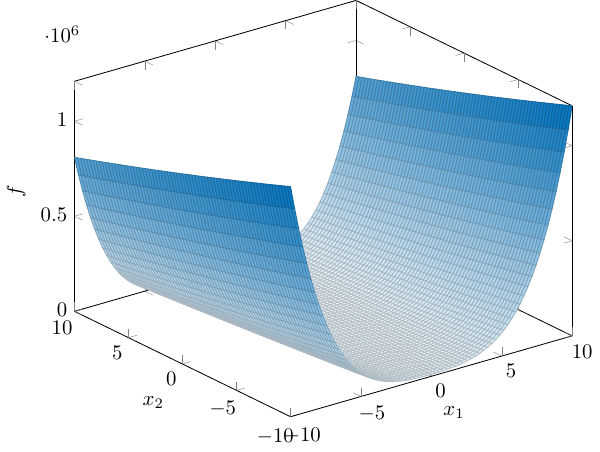}
		\caption{Landscape plot of the Rosenbrock function for $d=2$.}
		\label{fig_Rosenbrock}
	\end{figure}
	\item The fourth test problem is the DixonPrice function. Its expression is
	\begin{equation*}
		f(\bm{x}) = (x_1 - 1)^2 + \sum_{i=2}^{d} i (2x_i^2 - x_{i-1})^2.
	\end{equation*}
	The search domain is $\mathcal{X}=[-10,10]^d$. The landscape plot of this function for $d=2$ is given in Fig.~\ref{fig_DixonPrice}. The global minimum of this function also lies in a narrow valley. 
	\begin{figure}[h]
		\centering
		\includegraphics[width=0.36\linewidth]{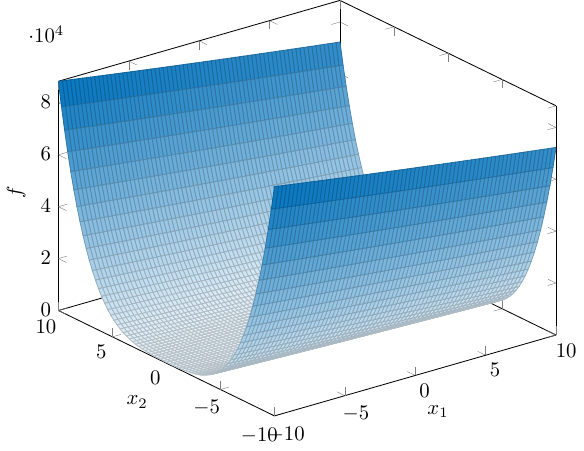}
		\caption{Landscape plot of the DixonPrice function for $d=2$.}
		\label{fig_DixonPrice}
	\end{figure}
	\item The fifth test problem is the Ackley function, whose expression is 
	\begin{equation*}
		f(\bm{x}) = -20 \exp \left(-0.2\sqrt{\frac{1}{d}\sum_{i=1}^{d}x_i^2}\right) - \exp \left(\frac{1}{d} \sum_{i=1}^{d} \cos (2\pi x_i)\right) + 20 + \exp(1).
	\end{equation*}
	We minimize the Ackley function in the domain $\mathcal{X}=[-32.768,32.768]^d$.  The landscape of this function for $d=2$ is illustrated in Fig.~\ref{fig_Ackley}. As we can see, the Ackley function has many local optima, which makes its global minimum difficult to find. 
	\begin{figure}[h]
		\centering
		\includegraphics[width=0.36\linewidth]{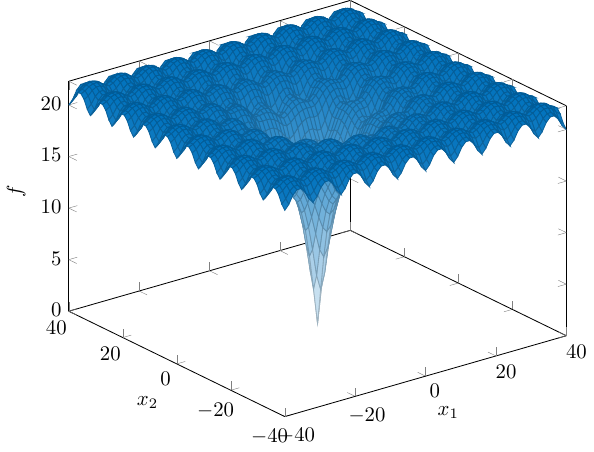}
		\caption{Landscape plot of the Ackley function for $d=2$.}
		\label{fig_Ackley}
	\end{figure}
	\item The sixth test problem is the Rastrigin function, whose formula is 
	\begin{equation*}
		f(\bm{x}) = 10d + \sum_{i=1}^{d} \left[x_i^2 - 10 \cos (2\pi x_i)\right].
	\end{equation*}
	The search domain of this function is $\mathcal{X} = [-5.12,5.12]^d$. The landscape of this function for $d=2$ is given in Fig.~\ref{fig_Rastrigin}. We can see that the Rastrigin function is highly multimodal. It has many local minima that can easily trap the search algorithm.
	\begin{figure}[h]
		\centering
		\includegraphics[width=0.36\linewidth]{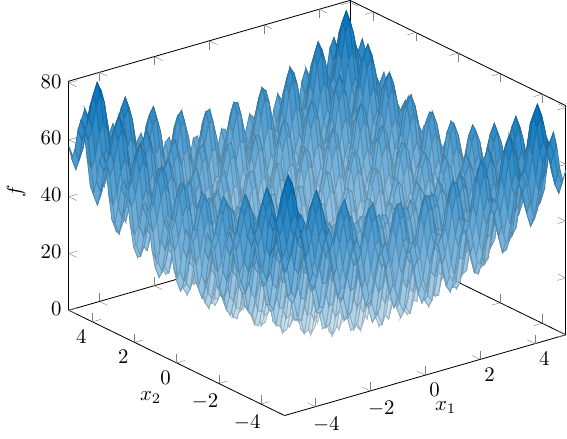}
		\caption{Landscape plot of the Rastrigin function for $d=2$.}
		\label{fig_Rastrigin}
	\end{figure}
	\item The seventh test problem is the Griewank function. The formula of this function is
	\begin{equation*}
		f(\bm{x}) = \sum_{i=1}^{d}\frac{x_i^2}{4000} - \prod_{i=1}^{d} \cos \left(\frac{x_i}{\sqrt{i}}\right) + 1.
	\end{equation*}
	The search domain of this problem is $\mathcal{X}=[-600,600]^d$. The landscapes of the Griewank function in the region of $[-600,600]^2$ and $[-10,10]^2$ are plotted in Fig.~\ref{fig_Griewank}(a) and Fig.~\ref{fig_Griewank}(b), respectively. As can be seen, the Griewank function has a global trend in the whole search space but has highly multimodal surface in the local spaces.
	\begin{figure}[h]
		\centering
		\subfloat[{$\mathcal{X} = [-600,600]^2$}]{\includegraphics[width=0.36\linewidth]{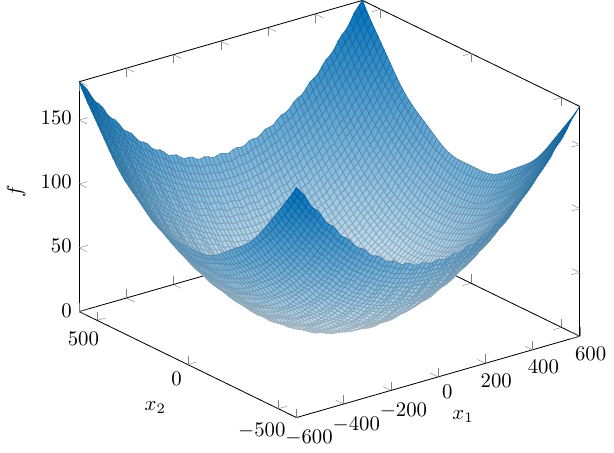}} \hfil
		\subfloat[{$\mathcal{X} = [-10,10]^2$}]{\includegraphics[width=0.36\linewidth]{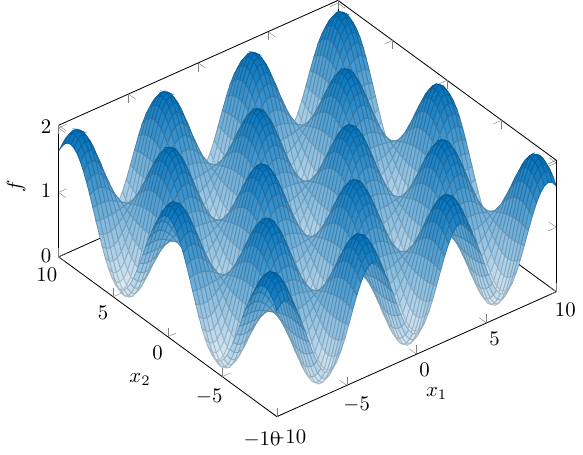}} \hfil
		\caption{Landscape plots of the Griewank function in $d=2$.}
		\label{fig_Griewank}
	\end{figure}
	\item The eighth test problem we select is the Levy function. The formula of this function is 
	\begin{equation*}
		f(\bm{x}) = \sin^2(\pi w_1) + \sum_{i=1}^{d-1} (w_i-1)^2 \left[1+10\sin^2 (\pi w_i +1)\right] + (w_d-1)^2 \left[1+\sin^2 (2\pi w_d)\right],
	\end{equation*}
	where $w_i  = 1 + \frac{x_i-1}{4}$ for $i=1,2,\cdots,d$. The search space of this function is $\mathcal{X} = [-10,10]^d$. The landscape of this function for $d=2$ is shown in Fig.~\ref{fig_Levy}. The Levy function also has many local minima.
	\begin{figure}[h]
		\centering
		\includegraphics[width=0.36\linewidth]{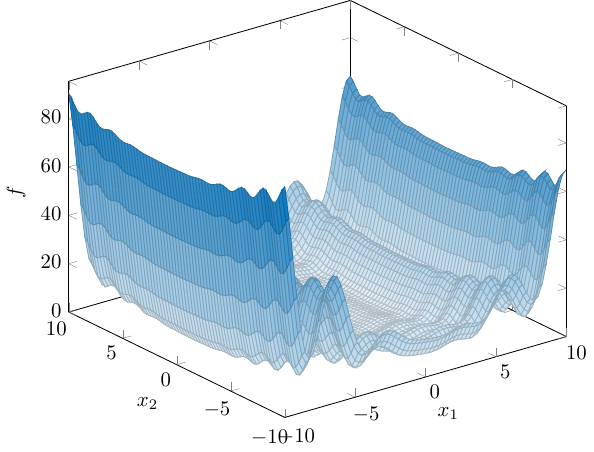}
		\caption{Landscape plot of the Levy function for $d=2$.}
		\label{fig_Levy}
	\end{figure}
	\item The ninth test problem we select is the Michalewicz function. Its formula is 
	\begin{equation*}
		f(\bm{x}) = - \sum_{i=1}^{d} \sin(x_i) \sin^{20}\left(\frac{i x_i^2}{\pi}\right).
	\end{equation*}
	The search domain of this function is $\mathcal{X} = [0,\pi]^d$. The landscape of this function for $d=2$ is plotted in Fig.~\ref{fig_Michalewicz}. The Michalewicz function has $d!$ local minima. This function has many valleys and ridges, which makes it difficult to optimize. 
	\begin{figure}[h]
		\centering
		\includegraphics[width=0.36\linewidth]{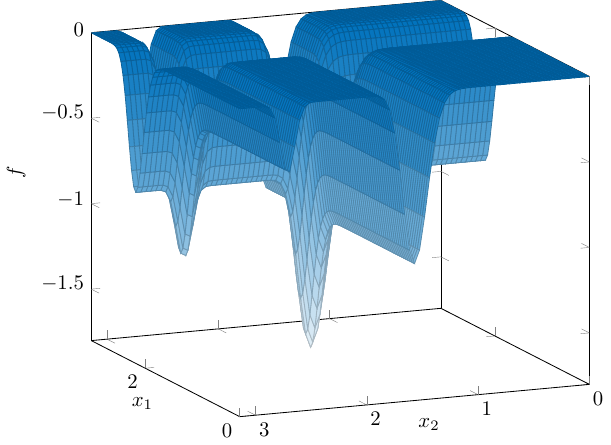}
		\caption{Landscape plot of the Michalewicz function for $d=2$.}
		\label{fig_Michalewicz}
	\end{figure}
	\item The last test problem we select is the Schwefel function. Its formula is 
	\begin{equation*}
		f(\bm{x}) = 418.9829d - \sum_{i=1}^{d} x_i \sin (\sqrt{\abs{x_i}}).
	\end{equation*}
	The search domain of this problem is $\mathcal{X} = [-500,500]^d$. The landscape of this function for $d=2$ is plotted in Fig.~\ref{fig_Schwefel}. As can be seen, the Schwefel function is very complex and has many local minima.
	\begin{figure}[h]
		\centering
		\includegraphics[width=0.36\linewidth]{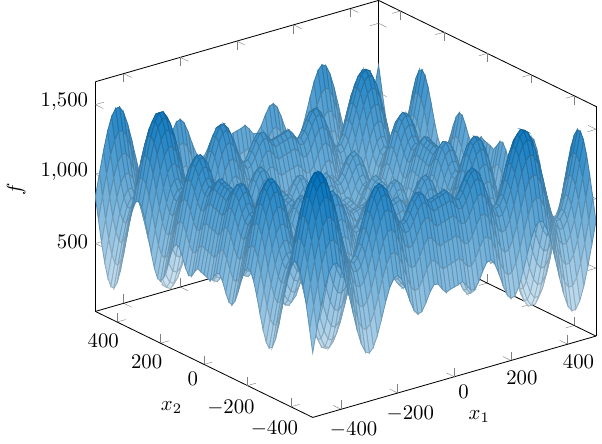}
		\caption{Landscape plot of the Schwefel function for $d=2$.}
		\label{fig_Schwefel}
	\end{figure}
\end{enumerate}

\newpage

\subsection{Experiment Results}
\label{section_results}

\begin{table}[h]
	\caption{Simple regrets obtained by the  EI and SSEI when only one query point is selected in each iteration}
	\label{table_EI_SSEI}
	\centering
	\renewcommand{\arraystretch}{1.1}
	\resizebox{0.45\textwidth}{!}{
		\begin{tabular}{c c c c } 
			\toprule
			$d$  &  $f$   & EI & SSEI  \\
			\midrule
			2 & GPMean & 1.98E-04 (2.41E-06) & \textbf{1.98E-04 (1.54E-08)} $+$ \\
			2 & SumSquares & 2.61E-07 (3.59E-07) & 1.27E-07 (2.46E-07) $\approx$ \\
			2 & Rosenbrock & 8.84E-02 (1.01E-01) & 1.51E-01 (2.30E-01) $\approx$ \\
			2 & DixonPrice & 2.99E-02 (7.75E-02) & 1.28E-02 (3.79E-02) $\approx$ \\
			2 & Ackley & 1.98E+00 (2.05E+00) & \textbf{8.90E-01 (9.32E-01)} $+$ \\
			2 & Rastrigin & 3.13E-01 (4.60E-01) & \textbf{ 1.53E-05 (2.46E-05) }$+$ \\
			2 & Griewank & 4.61E-01 (2.51E-01) & \textbf{2.80E-01 (1.27E-01) }$+$ \\
			2 & Levy & 3.26E-05 (9.88E-05) & \textbf{1.14E-05 (3.28E-05)} $+$ \\
			2 & Michalewicz & 2.07E-05 (4.74E-05) & 1.05E-05 (2.41E-05) $\approx$ \\
			2 & Schwefel & 3.65E-03 (7.95E-03) & \textbf{6.79E-04 (1.13E-03)} $+$ \\
			4 & GPMean & 3.33E-03 (6.24E-03) & \textbf{2.36E-04 (2.48E-03) }$+$ \\
			4 & SumSquares & 2.52E-02 (2.64E-02) & \textbf{1.03E-05 (1.23E-05)} $+$ \\
			4 & Rosenbrock & 9.46E+00 (9.96E+00) & \textbf{5.08E+00 (3.82E+00)} $+$ \\
			4 & DixonPrice & 2.34E+00 (2.53E+00) & \textbf{1.20E+00 (1.75E+00)} $+$ \\
			4 & Ackley & 4.07E+00 (3.68E+00) & \textbf{2.81E+00 (5.16E+00) }$+$ \\
			4 & Rastrigin & 3.29E+00 (2.50E+00) & \textbf{1.83E+00 (1.42E+00) }$+$ \\
			4 & Griewank & 9.59E-01 (2.57E-01) & \textbf{4.90E-01 (2.26E-01) }$+$ \\
			4 & Levy & 1.91E-01 (8.66E-02) & \textbf{2.25E-02 (3.97E-02) }$+$ \\
			4 & Michalewicz & 8.47E-02 (1.57E-01) & \textbf{5.07E-02 (1.46E-01)} $+$ \\
			4 & Schwefel & 2.14E+02 (1.18E+02) & \textbf{3.97E+01 (9.50E+01) }$+$ \\
			8 & GPMean & 1.53E-01 (2.60E-01) & 4.15E-01 (5.68E-01) $\approx$ \\
			8 & SumSquares & 2.37E-01 (2.44E-01) & \textbf{1.76E-04 (2.47E-04)} $+$ \\
			8 & Rosenbrock & 1.67E+02 (1.02E+02) & \textbf{6.48E+01 (6.66E+01)} $+$ \\
			8 & DixonPrice & 1.26E+02 (5.82E+01) & \textbf{3.89E+01 (3.37E+01) }$+$ \\
			8 & Ackley & 4.45E+00 (2.80E+00) & 3.50E+00 (4.28E+00) $\approx$ \\
			8 & Rastrigin & 2.22E+01 (7.67E+00) &\textbf{ 6.88E+00 (4.11E+00)} $+$ \\
			8 & Griewank & 2.20E+00 (7.14E-01) & \textbf{5.62E-01 (1.19E-01)} $+$ \\
			8 & Levy & 1.25E+00 (1.35E+00) & \textbf{2.77E-01 (3.97E-01) }$+$ \\
			8 & Michalewicz & 1.49E+00 (4.97E-01) & \textbf{4.90E-01 (3.83E-01)} $+$ \\
			8 & Schwefel & 7.53E+02 (2.46E+02) & \textbf{3.45E+02 (1.81E+02) }$+$ \\
			10 & GPMean & 4.60E-01 (2.98E-01) & 5.43E-01 (2.84E-01) $\approx$ \\
			10 & SumSquares & 4.62E-01 (3.78E-01) & \textbf{8.54E-04 (5.90E-04)} $+$ \\
			10 & Rosenbrock & 5.46E+02 (2.83E+02) & \textbf{1.46E+02 (1.06E+02)} $+$ \\
			10 & DixonPrice & 3.33E+02 (1.58E+02) & \textbf{7.52E+01 (4.95E+01) }$+$ \\
			10 & Ackley & 4.74E+00 (2.92E+00) & \textbf{2.79E+00 (3.62E+00)} $+$ \\
			10 & Rastrigin & 3.49E+01 (7.60E+00) & \textbf{1.63E+01 (6.13E+00) }$+$ \\
			10 & Griewank & 2.81E+00 (1.33E+00) & \textbf{7.26E-01 (1.68E-01) }$+$ \\
			10 & Levy & 1.83E+00 (1.57E+00) & \textbf{9.15E-01 (1.43E+00)} $+$ \\
			10 & Michalewicz & 3.24E+00 (8.04E-01) &\textbf{ 1.08E+00 (4.05E-01)} $+$ \\
			10 & Schwefel & 1.33E+03 (2.64E+02) &\textbf{ 6.41E+02 (2.51E+02) }$+$ \\
			20 & GPMean & 3.07E+00 (6.61E-03) & \textbf{2.29E+00 (8.32E-01) }$+$ \\
			20 & SumSquares & 2.09E+01 (2.30E+01) & \textbf{1.66E-02 (1.36E-02)} $+$ \\
			20 & Rosenbrock & 3.93E+03 (1.14E+03) & \textbf{8.89E+02 (3.94E+02)} $+$ \\
			20 & DixonPrice & 2.89E+03 (9.22E+02) & \textbf{3.14E+02 (1.91E+02) }$+$ \\
			20 & Ackley & 1.21E+01 (3.04E+00) & \textbf{2.46E+00 (2.47E-01) }$+$ \\
			20 & Rastrigin & 1.15E+02 (1.61E+01) & \textbf{5.67E+01 (1.10E+01)} $+$ \\
			20 & Griewank & 3.50E+01 (1.24E+01) &\textbf{ 9.37E-01 (8.69E-02)} $+$ \\
			20 & Levy & 3.60E+00 (1.29E+00) & \textbf{1.12E+00 (1.38E+00)} $+$ \\
			20 & Michalewicz & 1.21E+01 (7.80E-01) & \textbf{5.60E+00 (6.77E-01)} $+$ \\
			20 & Schwefel & 4.37E+03 (4.06E+02) & \textbf{2.01E+03 (3.46E+02) }$+$ \\
			30 & GPMean & 2.41E+00 (3.11E-04) & \textbf{2.36E+00 (2.04E-01)} $+$ \\
			30 & SumSquares & 1.91E+02 (8.33E+01) & \textbf{1.81E-01 (1.60E-01)} $+$ \\
			30 & Rosenbrock & 1.48E+04 (4.51E+03) & \textbf{2.11E+03 (8.08E+02)} $+$ \\
			30 & DixonPrice & 9.37E+03 (1.99E+03) & \textbf{1.13E+03 (6.36E+02) }$+$ \\
			30 & Ackley & 1.46E+01 (4.53E+00) & \textbf{3.29E+00 (1.85E-01) }$+$ \\
			30 & Rastrigin & 2.20E+02 (1.64E+01) & \textbf{1.43E+02 (1.79E+01)} $+$ \\
			30 & Griewank & 1.00E+02 (3.02E+01) & \textbf{1.04E+00 (2.78E-02)} $+$ \\
			30 & Levy & 6.57E+00 (1.68E+00) & \textbf{2.02E+00 (1.94E+00)} $+$ \\
			30 & Michalewicz & 2.05E+01 (8.12E-01) & \textbf{1.30E+01 (1.21E+00)} $+$ \\
			30 & Schwefel & 8.03E+03 (4.62E+02) & \textbf{4.21E+03 (5.46E+02) }$+$ \\
			\midrule
			\multicolumn{2}{c}{$+$/$\approx$/$-$}  &   \multicolumn{2}{c}{53/7/0}  \\
			\bottomrule
		\end{tabular}
	}
\end{table}


\begin{table*}
	\caption{Simple regrets obtained by the SSEI when different batch sizes are used}
	\label{table_SSEI}
	\centering
	\renewcommand{\arraystretch}{1.1}
	\resizebox{0.8\textwidth}{!}{
		\begin{tabular}{c c c c c c c c c c } 
			\toprule
			$d$  &   $f$   & $q=1$ & $q=2$ & $q=4$ & $q=8$ & $q=16$  & $q=32$  & $q=64$  & $q=128$   \\
			\midrule
			2 & GPMean & 1.98E-04 & \textbf{1.98E-04} & 1.98E-04 & 1.98E-04 & 1.98E-04 & 1.98E-04 & 1.98E-04 & 2.72E-04 \\
			2 & SumSquares & 1.27E-07 & \textbf{5.39E-08} & 1.59E-07 & 1.15E-07 & 7.43E-08 & 1.32E-07 & 1.42E-06 & 8.63E-06 \\
			2 & Rosenbrock & 1.51E-01 & 9.99E-02 & 7.81E-02 & 3.66E-02 & 4.59E-02 & \textbf{3.30E-02} & 7.40E-02 & 2.49E-01 \\
			2 & DixonPrice & 1.28E-02 & 1.10E-02 & \textbf{2.27E-03} & 2.85E-03 & 6.74E-03 & 5.48E-03 & 1.77E-02 & 7.82E-02 \\
			2 & Ackley & \textbf{8.90E-0}1 & 1.58E+00 & 1.19E+00 & 1.31E+00 & 1.39E+00 & 1.81E+00 & 3.51E+00 & 4.84E+00 \\
			2 & Rastrigin & 1.53E-05 & 2.11E-05 & 1.83E-05 & \textbf{8.27E-06} & 2.63E-05 & 2.65E-03 & 1.82E-01 & 1.09E+00 \\
			2 & Griewank & 2.80E-01 & 2.53E-01 & \textbf{2.03E-01} & 2.95E-01 & 2.83E-01 & 3.90E-01 & 4.15E-01 & 6.88E-01 \\
			2 & Levy & 1.14E-05 & 4.77E-05 & \textbf{7.56E-06} & 1.80E-05 & 1.47E-05 & 1.60E-04 & 4.25E-03 & 3.98E-02 \\
			2 & Michalewicz & 1.05E-05 & 8.53E-06 & 4.04E-06 & 1.12E-06 & \textbf{8.32E-07} & 1.11E-06 & 8.96E-05 & 1.18E-02 \\
			2 & Schwefel & 6.79E-04 & 5.98E-04 & 3.99E-04 & 7.30E-05 & \textbf{5.58E-05} & 6.21E-04 & 1.33E-01 & 1.54E+01 \\
			4 & GPMean & \textbf{2.36E-04} & 3.14E-02 & 6.16E-02 & 1.01E-01 & 2.14E-01 & 2.67E-01 & 4.16E-01 & 5.60E-01 \\
			4 & SumSquares & 1.03E-05 & 9.70E-06 & 1.08E-05 & \textbf{9.06E-06} & 2.10E-05 & 5.07E-05 & 2.61E-04 & 4.90E-03 \\
			4 & Rosenbrock & 5.08E+00 & 5.52E+00 & \textbf{3.95E+00} & 4.21E+00 & 4.11E+00 & 3.97E+00 & 1.33E+01 & 2.83E+02 \\
			4 & DixonPrice & 1.20E+00 & 2.14E+00 & 4.33E+00 & 4.64E+00 & 9.30E-01 & \textbf{6.59E-01} & 2.67E+00 & 5.50E+01 \\
			4 & Ackley & \textbf{2.81E+00} & 3.51E+00 & 6.17E+00 & 8.96E+00 & 9.41E+00 & 8.38E+00 & 9.04E+00 & 1.06E+01 \\
			4 & Rastrigin & \textbf{1.83E+00} & 2.42E+00 & 2.16E+00 & 2.76E+00 & 3.50E+00 & 3.46E+00 & 5.60E+00 & 1.04E+01 \\
			4 & Griewank & 4.90E-01 & \textbf{4.41E-01} & 4.70E-01 & 5.41E-01 & 6.58E-01 & 8.33E-01 & 7.68E-01 & 1.36E+00 \\
			4 & Levy & 2.25E-02 & 1.97E-02 & \textbf{8.72E-03} & 4.82E-02 & 5.20E-02 & 1.10E-01 & 2.73E-01 & 6.55E-01 \\
			4 & Michalewicz & 5.07E-02 & \textbf{2.40E-02} & 2.59E-02 & 7.97E-02 & 9.54E-02 & 1.25E-01 & 2.87E-01 & 8.03E-01 \\
			4 & Schwefel & 3.97E+01 & 2.33E+01 & 1.59E+01 & \textbf{8.10E+00 }& 7.18E+01 & 1.11E+02 & 1.88E+02 & 4.00E+02 \\
			8 & GPMean & \textbf{4.15E-01} & 4.36E-01 & 5.11E-01 & 5.27E-01 & 6.12E-01 & 6.99E-01 & 6.97E-01 & 1.14E+00 \\
			8 & SumSquares & \textbf{1.76E-04} & 1.84E-04 & 1.82E-04 & 2.35E-04 & 3.47E-04 & 4.39E-04 & 2.06E-03 & 1.26E-01 \\
			8 & Rosenbrock & \textbf{6.48E+01} & 7.87E+01 & 7.95E+01 & 6.83E+01 & 8.95E+01 & 1.80E+02 & 2.79E+02 & 1.80E+03 \\
			8 & DixonPrice & 3.89E+01 & 3.83E+01 & \textbf{2.65E+01} & 5.40E+01 & 4.64E+01 & 5.25E+01 & 1.29E+02 & 7.07E+02 \\
			8 & Ackley & \textbf{3.50E+00} & 3.94E+00 & 3.37E+00 & 4.95E+00 & 6.87E+00 & 1.09E+01 & 1.31E+01 & 1.58E+01 \\
			8 & Rastrigin & \textbf{6.88E+00} & 8.64E+00 & 1.35E+01 & 1.18E+01 & 1.40E+01 & 1.56E+01 & 2.46E+01 & 3.12E+01 \\
			8 & Griewank & 5.62E-01 & 5.61E-01 & \textbf{5.48E-01} & 5.66E-01 & 6.16E-01 & 6.16E-01 & 7.87E-01 & 1.08E+00 \\
			8 & Levy & 2.77E-01 & \textbf{1.68E-01 }& 3.37E-01 & 4.88E-01 & 7.28E-01 & 1.30E+00 & 2.43E+00 & 5.14E+00 \\
			8 & Michalewicz & \textbf{4.90E-01} & 5.53E-01 & 6.10E-01 & 6.10E-01 & 6.02E-01 & 7.11E-01 & 1.24E+00 & 2.41E+00 \\
			8 & Schwefel & \textbf{3.45E+02} & 3.47E+02 & 3.61E+02 & 3.61E+02 & 4.33E+02 & 5.51E+02 & 7.48E+02 & 9.55E+02 \\
			10 & GPMean & \textbf{5.43E-01} & 5.73E-01 & 6.24E-01 & 8.26E-01 & 7.00E-01 & 7.56E-01 & 6.73E-01 & 9.36E-01 \\
			10 & SumSquares & \textbf{8.54E-04} & 1.19E-03 & 1.25E-03 & 1.17E-03 & 1.84E-03 & 2.62E-03 & 2.29E-02 & 4.32E-01 \\
			10 & Rosenbrock & \textbf{1.46E+02} & 2.36E+02 & 2.12E+02 & 2.77E+02 & 4.14E+02 & 4.08E+02 & 8.03E+02 & 3.12E+03 \\
			10 & DixonPrice & \textbf{7.52E+01 }& 8.54E+01 & 8.01E+01 & 1.18E+02 & 1.23E+02 & 1.65E+02 & 2.72E+02 & 1.12E+03 \\
			10 & Ackley & 2.79E+00 & 3.24E+00 & \textbf{2.53E+00} & 4.31E+00 & 8.90E+00 & 1.16E+01 & 1.47E+01 & 1.65E+01 \\
			10 & Rastrigin & \textbf{1.63E+01} & 1.78E+01 & 1.87E+01 & 1.96E+01 & 2.26E+01 & 2.79E+01 & 3.41E+01 & 5.04E+01 \\
			10 & Griewank & 7.26E-01 & 7.37E-01 & 7.35E-01 & \textbf{7.19E-01} & 7.63E-01 & 8.37E-01 & 9.32E-01 & 1.28E+00 \\
			10 & Levy & 9.15E-01 & 6.61E-01 & \textbf{6.02E-01} & 1.37E+00 & 1.52E+00 & 2.51E+00 & 4.26E+00 & 7.02E+00 \\
			10 & Michalewicz & \textbf{1.08E+00} & 1.17E+00 & 1.17E+00 & 1.27E+00 & 1.31E+00 & 1.45E+00 & 2.18E+00 & 3.33E+00 \\
			10 & Schwefel & 6.41E+02 & \textbf{6.01E+02 }& 6.41E+02 & 7.44E+02 & 7.01E+02 & 9.95E+02 & 1.18E+03 & 1.49E+03 \\
			20 & GPMean & \textbf{2.29E+00} & 2.56E+00 & 2.60E+00 & 2.56E+00 & 2.49E+00 & 2.46E+00 & 2.82E+00 & 2.81E+00 \\
			20 & SumSquares & 1.66E-02 & 1.38E-02 & 1.39E-02 & \textbf{1.18E-02} & 1.86E-02 & 2.83E-02 & 1.11E-01 & 8.52E-01 \\
			20 & Rosenbrock & 8.89E+02 &\textbf{ 8.31E+02} & 8.32E+02 & 9.59E+02 & 1.04E+03 & 9.57E+02 & 2.22E+03 & 3.37E+03 \\
			20 & DixonPrice & 3.14E+02 & \textbf{3.13E+02} & 4.15E+02 & 4.04E+02 & 4.87E+02 & 6.32E+02 & 1.07E+03 & 2.21E+03 \\
			20 & Ackley & 2.46E+00 & \textbf{2.39E+00} & 2.45E+00 & 2.63E+00 & 2.72E+00 & 3.71E+00 & 8.09E+00 & 1.31E+01 \\
			20 & Rastrigin & 5.67E+01 & 6.04E+01 & \textbf{5.51E+01} & 6.14E+01 & 6.50E+01 & 7.07E+01 & 7.46E+01 & 9.39E+01 \\
			20 & Griewank & \textbf{9.37E-01} & 9.59E-01 & 9.39E-01 & 9.47E-01 & 9.62E-01 & 9.55E-01 & 1.05E+00 & 1.50E+00 \\
			20 & Levy & \textbf{1.12E+00} & 1.55E+00 & 2.07E+00 & 1.92E+00 & 3.56E+00 & 4.29E+00 & 7.47E+00 & 1.17E+01 \\
			20 & Michalewicz & 5.60E+00 & \textbf{5.47E+00 }& 5.79E+00 & 5.57E+00 & 5.90E+00 & 5.99E+00 & 6.59E+00 & 7.74E+00 \\
			20 & Schwefel & 2.01E+03 & \textbf{1.91E+03} & 1.98E+03 & 1.96E+03 & 1.95E+03 & 2.07E+03 & 2.49E+03 & 3.06E+03 \\
			30 & GPMean & \textbf{2.36E+00} & 2.41E+00 & 2.41E+00 & 2.41E+00 & 2.40E+00 & 2.41E+00 & 2.41E+00 & 2.41E+00 \\
			30 & SumSquares & 1.81E-01 & \textbf{1.21E-01} & 1.36E-01 & 1.56E-01 & 2.23E-01 & 3.82E-01 & 1.40E+00 & 9.00E+00 \\
			30 & Rosenbrock &\textbf{ 2.11E+03} & 2.23E+03 & 2.25E+03 & 2.39E+03 & 3.27E+03 & 3.96E+03 & 5.96E+03 & 1.08E+04 \\
			30 & DixonPrice & \textbf{1.13E+03} & 1.15E+03 & 1.29E+03 & 1.65E+03 & 1.88E+03 & 2.23E+03 & 3.80E+03 & 6.36E+03 \\
			30 & Ackley & \textbf{3.29E+00} & 3.31E+00 & 3.41E+00 & 3.47E+00 & 3.78E+00 & 4.73E+00 & 7.77E+00 & 1.22E+01 \\
			30 & Rastrigin & \textbf{1.43E+02} & 1.51E+02 & 1.53E+02 & 1.44E+02 & 1.60E+02 & 1.62E+02 & 1.65E+02 & 1.91E+02 \\
			30 & Griewank & 1.04E+00 & \textbf{1.03E+00} & 1.06E+00 & 1.10E+00 & 1.08E+00 & 1.14E+00 & 1.58E+00 & 3.87E+00 \\
			30 & Levy & \textbf{2.02E+00} & 2.90E+00 & 2.38E+00 & 5.12E+00 & 4.63E+00 & 8.68E+00 & 1.38E+01 & 2.39E+01 \\
			30 & Michalewicz & 1.30E+01 & 1.33E+01 & \textbf{1.29E+01} & 1.32E+01 & 1.37E+01 & 1.36E+01 & 1.47E+01 & 1.58E+01 \\
			30 & Schwefel & 4.21E+03 & 4.23E+03 & 4.17E+03 & 4.04E+03 & \textbf{4.03E+03} & 4.27E+03 & 4.56E+03 & 5.41E+03 \\
			\bottomrule
		\end{tabular}
	}
\end{table*}

\begin{table*}
	\caption{Simple regrets obtained by different batch approaches when $q=4$ are used}
	\label{table_batch_q4}
	\centering
	\renewcommand{\arraystretch}{1.3}
	\resizebox{1.0\textwidth}{!}{
		\begin{tabular}{c c c c c c c c c c c c c} 
			\toprule
			$d$  &   $f$   & RS & $q$EI & $q$logEI & F$q$EI & TuRBO & MSMR & MACE   & PEI & EIMI & KB & SSEI \\
			\midrule
			2 & GPMean & 6.07E-02 $+$  & 1.99E-04 $+$ & 1.98E-04 $=$ & 1.99E-04 $+$ & 6.17E-04 $+$ & 4.36E-04 $+$ & \textbf{1.98E-04} $-$ & 1.98E-04 $+$ & 2.02E-04 $+$ & 1.98E-04 $+$ & 1.98E-04 \\ 
			2 & SumSquares & 1.51E-01 $+$  & 8.59E-07 $+$ & 5.72E-07 $=$ & 2.95E-06 $+$ & 1.66E-04 $+$ & 6.83E-06 $+$ & \textbf{1.49E-09} $-$ & 7.06E-06 $+$ & 1.53E-05 $+$ & 5.44E-06 $+$ & 1.59E-07 \\ 
			2 & Rosenbrock & 3.76E+00 $+$  & 1.41E-02 $-$ & 2.38E-02 $-$ & 2.24E-04 $-$ & 2.86E-01 $+$ & 1.05E-03 $-$ & 3.22E-04 $-$ & \textbf{2.07E-04} $-$ & 1.71E-03 $-$ & 2.20E-04 $-$ & 7.81E-02 \\ 
			2 & DixonPrice & 5.89E-01 $+$  & 1.95E-03 $=$ & 2.12E-03 $=$ & 3.82E-05 $-$ & 6.05E-03 $+$ & 1.46E-04 $-$ & \textbf{1.91E-05} $-$ & 2.24E-05 $-$ & 4.66E-05 $-$ & 2.09E-05 $-$ & 2.27E-03 \\ 
			2 & Ackley & 6.46E+00 $+$  & 1.40E+00 $=$ & 1.27E+00 $=$ & 2.61E+00 $+$ & \textbf{1.92E-01} $-$ & 3.08E+00 $+$ & 1.81E+00 $+$ & 2.40E+00 $+$ & 3.56E+00 $+$ & 2.97E+00 $+$ & 1.19E+00 \\ 
			2 & Rastrigin & 3.06E+00 $+$  & 2.17E-01 $+$ & 2.27E-01 $+$ & 1.99E+00 $+$ & 6.88E-01 $+$ & 2.44E+00 $+$ & 3.30E+00 $+$ & 1.59E+00 $+$ & 2.70E+00 $+$ & 2.40E+00 $+$ & \textbf{1.83E-05 }\\ 
			2 & Griewank & 1.19E+00 $+$  & 4.96E-01 $+$ & 5.39E-01 $+$ & 6.77E-02 $-$ & 9.89E-02 $-$ & 1.03E-01 $-$ & 1.18E-01 $-$ & \textbf{6.60E-02} $-$ & 1.22E-01 $-$ & 1.33E-01 $-$ & 2.03E-01 \\ 
			2 & Levy & 1.11E-01 $+$  & 3.78E-03 $+$ & 5.00E-03 $+$ & 5.51E-02 $+$ & 1.73E-04 $+$ & 1.10E-01 $+$ & 1.92E-02 $+$ & 4.83E-02 $+$ & 2.03E-01 $+$ & 1.43E-01 $+$ & \textbf{7.56E-06} \\ 
			2 & Michalewicz & 2.46E-01 $+$  & 2.49E-05 $+$ & 1.31E-05 $+$ & 2.50E-03 $+$ & 2.04E-04 $+$ & 1.98E-04 $+$ & 1.82E-02 $+$ & 8.43E-04 $+$ & 7.47E-02 $+$ & 8.44E-03 $+$ & \textbf{4.04E-06 }\\ 
			2 & Schwefel & 1.06E+02 $+$  & 3.81E-03 $+$ & 2.58E-03 $+$ & 7.58E-01 $+$ & 5.74E+01 $+$ & 1.34E-01 $+$ & 2.74E+01 $+$ & 2.45E-01 $+$ & 1.53E+01 $+$ & 8.50E+00 $+$ & \textbf{3.99E-04} \\ 
			4 & GPMean & 8.25E-01 $+$  & 2.00E-02 $=$ & \textbf{1.30E-03} $=$ & 1.08E-01 $+$ & 3.50E-01 $+$ & 1.67E-01 $+$ & 1.62E-01 $+$ & 1.65E-02 $=$ & 1.31E-01 $+$ & 1.89E-01 $+$ & 6.19E-02 \\ 
			4 & SumSquares & 5.19E+00 $+$  & 3.28E-02 $+$ & 6.92E-05 $+$ & 8.78E-04 $+$ & 3.92E-03 $+$ & 8.13E-03 $+$ & 8.30E-03 $+$ & 6.24E-03 $+$ & 6.17E-04 $+$ & 3.84E-04 $+$ & \textbf{1.08E-05} \\ 
			4 & Rosenbrock & 8.06E+02 $+$  & 6.47E+00 $+$ & 8.01E+00 $+$ & 4.64E-01 $-$ & 1.07E+01 $=$ & 1.36E+01 $+$ & 2.91E-01 $-$ & 3.96E-01 $-$ & 4.65E-01 $-$ & \textbf{2.84E-01} $-$ & 3.95E+00 \\ 
			4 & DixonPrice & 1.75E+02 $+$  & 1.62E+00 $=$ & 1.74E+00 $=$ & 3.78E-02 $-$ & 5.76E-01 $-$ & 1.01E+01 $+$ & \textbf{1.29E-02} $-$ & 2.76E-02 $-$ & 2.21E-01 $-$ & 2.40E-02 $-$ & 4.33E+00 \\ 
			4 & Ackley & 1.46E+01 $+$  & 2.76E+00 $=$ & 3.19E+00 $=$ & 1.33E+01 $+$ & \textbf{1.02E+00} $-$ & 3.63E+00 $=$ & 1.50E+01 $+$ & 1.32E+01 $+$ & 1.55E+01 $+$ & 1.45E+01 $+$ & 6.17E+00 \\ 
			4 & Rastrigin & 1.96E+01 $+$  & 5.47E+00 $+$ & 5.51E+00 $+$ & 1.22E+01 $+$ & 6.67E+00 $+$ & 1.54E+01 $+$ & 1.72E+01 $+$ & 9.28E+00 $+$ & 2.11E+01 $+$ & 1.28E+01 $+$ & \textbf{2.16E+00 }\\
			4 & Griewank & 9.68E+00 $+$  & 1.04E+00 $+$ & 4.90E-01 $=$ & 6.44E-01 $+$ & \textbf{4.42E-01} $=$ & 9.01E-01 $+$ & 1.34E+00 $+$ & 6.22E-01 $+$ & 3.29E+00 $+$ & 3.06E+00 $+$ & 4.70E-01 \\ 
			4 & Levy & 2.13E+00 $+$  & 4.58E-01 $+$ & 5.21E-01 $+$ & 4.12E-01 $+$ & 1.47E-01 $+$ & 7.55E-01 $+$ & 1.28E+00 $+$ & 4.19E-01 $+$ & 2.30E+00 $+$ & 9.75E-01 $+$ & \textbf{8.72E-03 }\\ 
			4 & Michalewicz & 1.53E+00 $+$  & 1.72E-01 $+$ & 2.08E-01 $+$ & 1.28E+00 $+$ & 3.03E-01 $+$ & 4.23E-01 $+$ & 1.59E+00 $+$ & 1.05E+00 $+$ & 1.20E+00 $+$ & 1.45E+00 $+$ & \textbf{2.59E-02} \\ 
			4 & Schwefel & 5.82E+02 $+$  & 2.45E+02 $+$ & 2.08E+02 $+$ & 4.35E+02 $+$ & 3.29E+02 $+$ & 2.77E+02 $+$ & 5.03E+02 $+$ & 3.57E+02 $+$ & 5.27E+02 $+$ & 4.07E+02 $+$ & \textbf{1.59E+01} \\ 
			8 & GPMean & 2.07E+00 $+$  & 1.15E-01 $=$ & 5.81E-02 $=$ & 3.05E-01 $=$ & 2.29E-01 $=$ & 5.88E-01 $=$ & 2.95E-01 $=$ & \textbf{2.12E-03} $-$ & 2.27E-02 $=$ & 6.05E-01 $=$ & 5.11E-01 \\ 
			8 & SumSquares & 5.67E+01 $+$  & 1.08E+00 $+$ & 2.60E-03 $+$ & 1.34E-01 $+$ & 3.39E-02 $+$ & 1.37E-01 $+$ &\textbf{1.09E-04}$-$ & 1.25E-01 $+$ & 2.13E-01 $+$ & 2.41E-01 $+$ & 1.82E-04 \\ 
			8 & Rosenbrock & 1.44E+04 $+$  & 1.12E+03 $+$ & 9.50E+02 $+$ & 6.51E+02 $+$ & 8.12E+01 $=$ & 2.03E+03 $+$ & 5.69E+03 $+$ & 6.47E+02 $+$ & 7.93E+02 $+$ & 6.68E+02 $+$ & \textbf{7.95E+01} \\ 
			8 & DixonPrice & 9.67E+03 $+$  & 3.12E+02 $+$ & 3.14E+02 $+$ & 4.97E+02 $+$ & \textbf{6.68E+00} $-$ & 8.49E+02 $+$ & 2.42E+03 $+$ & 4.46E+02 $+$ & 5.30E+02 $+$ & 4.35E+02 $+$ & 2.65E+01 \\ 
			8 & Ackley & 1.77E+01 $+$  & 4.93E+00 $=$ & 3.06E+00 $=$ & 1.63E+01 $+$ & \textbf{1.77E+00} $=$ & 8.24E+00 $+$ & 1.78E+01 $+$ & 1.32E+01 $+$ & 1.57E+01 $+$ & 1.66E+01 $+$ & 3.37E+00 \\ 
			8 & Rastrigin & 6.37E+01 $+$  & 2.88E+01 $+$ & 2.93E+01 $+$ & 3.07E+01 $+$ & 1.64E+01 $=$ & 2.70E+01 $+$ & 3.11E+01 $+$ & 2.90E+01 $+$ & 3.53E+01 $+$ & 3.14E+01 $+$ & \textbf{1.35E+01} \\ 
			8 & Griewank & 5.10E+01 $+$  & 2.05E+00 $+$ & \textbf{5.25E-01} $=$ & 1.26E+00 $+$ & 9.40E-01 $+$ & 1.28E+00 $+$ & 7.27E-01 $+$ & 1.07E+00 $+$ & 1.42E+00 $+$ & 1.27E+00 $+$ & 5.48E-01 \\ 
			8 & Levy & 1.23E+01 $+$  & 1.27E+00 $+$ & 1.35E+00 $+$ & 3.22E+00 $+$ & 1.06E+00 $+$ & 5.34E+00 $+$ & 2.96E+00 $+$ & 2.57E+00 $+$ & 7.01E+00 $+$ & 2.33E+00 $+$ & \textbf{3.37E-01} \\ 
			8 & Michalewicz & 4.25E+00 $+$  & 2.49E+00 $+$ & 2.23E+00 $+$ & 3.34E+00 $+$ & 1.81E+00 $+$ & 2.64E+00 $+$ & 3.85E+00 $+$ & 2.97E+00 $+$ & 3.21E+00 $+$ & 3.48E+00 $+$ & \textbf{6.10E-01} \\ 
			8 & Schwefel & 1.67E+03 $+$  & 9.59E+02 $+$ & 8.65E+02 $+$ & 1.02E+03 $+$ & 7.14E+02 $+$ & 1.00E+03 $+$ & 1.14E+03 $+$ & 9.90E+02 $+$ & 1.23E+03 $+$ & 1.08E+03 $+$ & \textbf{3.61E+02} \\ 
			10 & GPMean & 1.92E+00 $+$  & 7.16E-01 $+$ & 3.53E-01 $-$ & 3.35E-01 $-$ & 4.72E-01 $=$ & 4.40E-01 $-$ & 5.77E-01 $=$ & \textbf{2.59E-01} $-$ & 3.23E-01 $-$ & 6.15E-01 $=$ & 6.24E-01 \\ 
			10 & SumSquares & 1.09E+02 $+$  & 3.16E+00 $+$ & 1.70E-02 $+$ & 4.12E-01 $+$ & 7.41E-02 $+$ & 5.14E-01 $+$ & \textbf{3.77E-04} $-$ & 5.42E-01 $+$ & 8.30E-01 $+$ & 6.58E-01 $+$ & 1.25E-03 \\ 
			10 & Rosenbrock & 4.14E+04 $+$  & 2.45E+03 $+$ & 2.37E+03 $+$ & 1.93E+03 $+$ & 2.41E+02 $=$ & 6.87E+03 $+$ & 1.17E+04 $+$ & 1.72E+03 $+$ & 2.04E+03 $+$ & 1.98E+03 $+$ & \textbf{2.12E+02} \\ 
			10 & DixonPrice & 2.75E+04 $+$  & 6.45E+02 $+$ & 6.56E+02 $+$ & 1.24E+03 $+$ & \textbf{2.19E+01} $-$ & 3.55E+03 $+$ & 4.71E+03 $+$ & 1.06E+03 $+$ & 1.08E+03 $+$ & 1.01E+03 $+$ & 8.01E+01 \\ 
			10 & Ackley & 1.88E+01 $+$  & 7.37E+00 $+$ & 3.35E+00 $+$ & 1.60E+01 $+$ & \textbf{1.78E+00 }$=$ & 1.26E+01 $+$ & 1.52E+01 $+$ & 1.17E+01 $+$ & 9.70E+00 $+$ & 1.68E+01 $+$ & 2.53E+00 \\ 
			10 & Rastrigin & 8.73E+01 $+$  & 4.56E+01 $+$ & 4.47E+01 $+$ & 2.43E+01 $=$ & 2.75E+01 $+$ & 3.98E+01 $+$ & 3.44E+01 $+$ & 1.96E+01 $=$ & 4.23E+01 $+$ & 3.75E+01 $+$ & \textbf{1.87E+01} \\ 
			10 & Griewank & 7.73E+01 $+$  & 3.23E+00 $+$ & 9.28E-01 $+$ & 1.47E+00 $+$ & 1.02E+00 $+$ & 1.37E+00 $+$ & 7.40E-01 $=$ & 1.28E+00 $+$ & 1.44E+00 $+$ & 1.34E+00 $+$ & \textbf{7.35E-01} \\ 
			10 & Levy & 2.03E+01 $+$  & 2.01E+00 $+$ & 1.71E+00 $+$ & 2.41E+00 $+$ & 2.52E+00 $+$ & 1.03E+01 $+$ & 3.46E+00 $+$ & 2.19E+00 $+$ & 7.08E+00 $+$ & 4.80E+00 $+$ & \textbf{6.02E-01} \\ 
			10 & Michalewicz & 5.79E+00 $+$  & 4.28E+00 $+$ & 3.85E+00 $+$ & 4.72E+00 $+$ & 2.76E+00 $+$ & 4.18E+00 $+$ & 5.26E+00 $+$ & 4.56E+00 $+$ & 4.92E+00 $+$ & 4.72E+00 $+$ & \textbf{1.17E+00} \\ 
			10 & Schwefel & 2.33E+03 $+$  & 1.69E+03 $+$ & 1.50E+03 $+$ & 1.33E+03 $+$ & 1.06E+03 $+$ & 1.46E+03 $+$ & 1.56E+03 $+$ & 1.34E+03 $+$ & 1.52E+03 $+$ & 1.45E+03 $+$ & \textbf{6.41E+02} \\ 
			20 & GPMean & 3.06E+00 $=$  & 3.06E+00 $+$ & 3.07E+00 $+$ & 2.96E+00 $=$ & \textbf{1.57E+00} $-$ & 1.80E+00 $-$ & 2.91E+00 $=$ & 3.06E+00 $+$ & 2.95E+00 $=$ & 2.97E+00 $=$ & 2.60E+00 \\ 
			20 & SumSquares & 6.71E+02 $+$  & 8.67E+01 $+$ & 2.47E-01 $+$ & 1.16E+01 $+$ & 7.10E-01 $+$ & 7.24E+00 $+$ & 1.56E-02 $=$ & 1.01E+01 $+$ & 9.51E+00 $+$ & 1.06E+01 $+$ & \textbf{1.39E-02 }\\ 
			20 & Rosenbrock & 2.65E+05 $+$  & 9.27E+03 $+$ & 8.42E+03 $+$ & 8.18E+03 $+$ & \textbf{7.26E+02} $=$ & 2.39E+04 $+$ & 4.19E+04 $+$ & 7.02E+03 $+$ & 7.59E+03 $+$ & 9.22E+03 $+$ & 8.32E+02 \\ 
			20 & DixonPrice & 2.75E+05 $+$  & 2.12E+03 $+$ & 2.91E+03 $+$ & 6.65E+03 $+$ & \textbf{1.69E+02} $-$ & 1.99E+04 $+$ & 3.17E+04 $+$ & 6.99E+03 $+$ & 6.23E+03 $+$ & 5.96E+03 $+$ & 4.15E+02 \\ 
			20 & Ackley & 1.98E+01 $+$  & 1.51E+01 $+$ & 3.91E+00 $+$ & 1.59E+01 $+$ & \textbf{2.34E+00} $=$ & 1.82E+01 $+$ & 1.41E+01 $+$ & 1.16E+01 $+$ & 1.22E+01 $+$ & 1.82E+01 $+$ & 2.45E+00 \\ 
			20 & Rastrigin & 2.31E+02 $+$  & 1.26E+02 $+$ & 1.36E+02 $+$ & \textbf{4.93E+01} $=$ & 5.98E+01 $=$ & 8.92E+01 $+$ & 8.94E+01 $+$ & 1.69E+02 $+$ & 6.53E+01 $=$ & 7.79E+01 $=$ & 5.51E+01 \\ 
			20 & Griewank & 2.60E+02 $+$  & 3.22E+01 $+$ & 1.07E+00 $+$ & 1.78E+00 $+$ & 1.18E+00 $+$ & 1.48E+00 $+$ & \textbf{9.00E-01} $=$ & 1.56E+00 $+$ & 1.60E+00 $+$ & 1.55E+00 $+$ & 9.39E-01 \\ 
			20 & Levy & 7.62E+01 $+$  & 3.99E+00 $+$ & 4.70E+00 $+$ & 6.88E+00 $+$ & 5.60E+00 $+$ & 3.27E+01 $+$ & 1.46E+01 $+$ & 5.72E+00 $+$ & 8.75E+00 $+$ & 7.67E+00 $+$ & \textbf{2.07E+00} \\ 
			20 & Michalewicz & 1.35E+01 $+$  & 1.24E+01 $+$ & 1.18E+01 $+$ & 1.35E+01 $+$ & 7.72E+00 $+$ & 1.16E+01 $+$ & 1.35E+01 $+$ & 1.33E+01 $+$ & 1.33E+01 $+$ & 1.34E+01 $+$ & \textbf{5.79E+00} \\ 
			20 & Schwefel & 5.53E+03 $+$  & 4.74E+03 $+$ & 4.40E+03 $+$ & 5.26E+03 $+$ & 2.44E+03 $+$ & 4.40E+03 $+$ & 4.72E+03 $+$ & 5.55E+03 $+$ & 4.00E+03 $+$ & 5.17E+03 $+$ & \textbf{1.98E+03} \\ 
			30 & GPMean & 2.41E+00 $+$  & 2.41E+00 $+$ & 2.41E+00 $+$ & 2.41E+00 $+$ & \textbf{1.27E+00} $-$ & 1.46E+00 $-$ & 2.41E+00 $+$ & 2.41E+00 $+$ & 2.41E+00 $+$ & 2.41E+00 $+$ & 2.41E+00 \\ 
			30 & SumSquares & 1.88E+03 $+$  & 4.48E+02 $+$ & 4.56E-01 $+$ & 3.33E+01 $+$ & 2.95E+00 $+$ & 1.65E+01 $+$ & \textbf{1.31E-01} $=$ & 3.25E+01 $+$ & 2.54E+01 $+$ & 2.88E+01 $+$ & 1.36E-01 \\ 
			30 & Rosenbrock & 7.26E+05 $+$  & 1.91E+04 $+$ & 1.60E+04 $+$ & 1.64E+04 $+$ & 4.22E+03 $+$ & 3.65E+04 $+$ & 8.62E+04 $+$ & 1.51E+04 $+$ & 1.90E+04 $+$ & 1.99E+04 $+$ & \textbf{2.25E+03}\\ 
			30 & DixonPrice & 9.98E+05 $+$  & 5.25E+03 $+$ & 6.85E+03 $+$ & 1.82E+04 $+$ & \textbf{7.20E+02} $-$ & 3.55E+04 $+$ & 8.89E+04 $+$ & 2.06E+04 $+$ & 1.77E+04 $+$ & 1.84E+04 $+$ & 1.29E+03 \\ 5
			30 & Ackley & 2.03E+01 $+$  & 1.78E+01 $+$ & 4.69E+00 $+$ & 1.49E+01 $+$ & 2.71E+00 $-$ & 1.86E+01 $+$ & 8.27E+00 $+$ & 1.15E+01 $+$ & 8.48E+00 $+$ & 1.82E+01 $+$ & 3.41E+00 \\ 
			30 & Rastrigin & 3.82E+02 $+$  & 2.38E+02 $+$ & 2.40E+02 $+$ & \textbf{1.07E+02} $-$ & 1.10E+02 $-$ & 2.19E+02 $+$ & 2.30E+02 $+$ & 2.92E+02 $+$ & 1.52E+02 $=$ & 1.54E+02 $=$ & 1.53E+02 \\ 
			30 & Griewank & 4.60E+02 $+$  & 9.16E+01 $+$ & 1.10E+00 $+$ & 2.23E+00 $+$ & 1.46E+00 $+$ & 1.52E+00 $+$ & \textbf{1.03E+00} $-$ & 1.66E+00 $+$ & 1.97E+00 $+$ & 1.95E+00 $+$ & 1.06E+00 \\ 
			30 & Levy & 1.51E+02 $+$  & 6.21E+00 $+$ & 6.88E+00 $+$ & 1.32E+01 $+$ & 1.01E+01 $+$ & 4.79E+01 $+$ & 3.84E+01 $+$ & 1.29E+01 $+$ & 1.38E+01 $+$ & 1.15E+01 $+$ & \textbf{2.38E+00} \\ 
			30 & Michalewicz & 2.14E+01 $+$  & 2.07E+01 $+$ & 2.02E+01 $+$ & 2.15E+01 $+$ & 1.29E+01 $=$ & 2.06E+01 $+$ & 2.14E+01 $+$ & 2.15E+01 $+$ & 2.15E+01 $+$ & 2.13E+01 $+$ & \textbf{1.29E+01} \\ 
			30 & Schwefel & 9.03E+03 $+$  & 7.97E+03 $+$ & 7.79E+03 $+$ & 8.57E+03 $+$ & \textbf{3.83E+03} $=$ & 7.76E+03 $+$ & 7.80E+03 $+$ & 9.06E+03 $+$ & 6.57E+03 $+$ & 8.83E+03 $+$ & 4.17E+03 \\ 
			\midrule
			\multicolumn{2}{c}{$+$/$\approx$/$-$}  &   59/1/0 &  52/7/1 &  47/11/ 2   & 49/4/7 &  34/14/12  & 52/2/6 &  43/7/10   &  51/2/7    &  50/4/6  &   50/5/5 &  N.A. \\
			\bottomrule
		\end{tabular}
	}
\end{table*}

\begin{table*}
	\caption{Simple regrets obtained by different batch approaches when $q=16$ are used}
	\label{table_batch_q16}
	\centering
	\renewcommand{\arraystretch}{1.3}
	\resizebox{1.0\textwidth}{!}{
		\begin{tabular}{c c c c c c c c c c c c c} 
			\toprule
			$d$  &   $f$   & RS & $q$EI & $q$logEI & F$q$EI & TuRBO & MSMR & MACE   & PEI & EIMI & KB & SSEI \\
			\midrule
			2 & GPMean & 8.35E-02 $+$  & 1.99E-04 $+$ & 1.98E-04 $=$ & 1.99E-04 $+$ & 1.31E-01 $+$ & 2.36E-04 $+$ & 1.98E-04 $=$ & 2.00E-04 $+$ & 2.36E-04 $+$ & 1.99E-04 $=$ & \textbf{1.98E-04} \\ 
			2 & SumSquares & 1.71E-01 $+$  & 4.03E-07 $+$ & 4.21E-07 $+$ & 4.92E-06 $+$ & 1.87E-04 $+$ & 1.42E-05 $+$ & \textbf{6.72E-09 }$-$ & 1.09E-05 $+$ & 1.40E-04 $+$ & 8.39E-06 $+$ & 7.43E-08 \\ 
			2 & Rosenbrock & 3.20E+00 $+$  & 1.45E-02 $-$ & 2.97E-02 $=$ & \textbf{3.21E-04} $-$ & 7.43E-01 $+$ & 4.79E-04 $-$ & 4.21E-03 $-$ & 6.84E-04 $-$ & 3.03E-02 $-$ & 3.62E-04 $-$ & 4.59E-02 \\ 
			2 & DixonPrice & 1.28E+00 $+$  & 1.05E-02 $=$ & 7.49E-03 $=$ & 5.53E-05 $-$ & 5.80E-03 $=$ & 6.42E-05 $-$ & 1.50E-04 $-$ & 8.32E-05 $-$ & 4.15E-04 $-$ & \textbf{5.26E-05} $-$ & 6.74E-03 \\ 
			2 & Ackley & 6.01E+00 $+$  & 1.96E+00 $=$ & 1.79E+00 $=$ & 3.42E+00 $+$ & \textbf{1.95E-01} $-$ & 3.50E+00 $+$ & 4.82E+00 $+$ & 2.53E+00 $+$ & 9.74E+00 $+$ & 4.66E+00 $+$ & 1.39E+00 \\ 
			2 & Rastrigin & 3.50E+00 $+$  & 5.92E-01 $+$ & 3.43E-01 $+$ & 4.52E+00 $+$ & 9.81E-01 $+$ & 3.15E+00 $+$ & 6.44E+00 $+$ & 2.56E+00 $+$ & 6.80E+00 $+$ & 6.32E+00 $+$ & \textbf{2.63E-05} \\ 
			2 & Griewank & 1.34E+00 $+$  & 5.76E-01 $+$ & 5.97E-01 $+$ & 9.70E-02 $-$ & 1.07E-01 $-$ & 2.11E-01 $=$ & 3.42E-01 $=$ & \textbf{7.41E-02} $-$ & 7.56E-01 $+$ & 1.43E-01 $-$ & 2.83E-01 \\ 
			2 & Levy & 1.47E-01 $+$  & 2.36E-02 $+$ & 3.48E-02 $+$ & 3.44E-01 $+$ & 3.39E-03 $+$ & 9.03E-02 $+$ & 4.80E-01 $+$ & 1.73E-01 $+$ & 6.25E-01 $+$ & 4.24E-01 $+$ & \textbf{1.47E-05} \\ 
			2 & Michalewicz & 2.47E-01 $+$  & 7.54E-05 $+$ & 2.07E-05 $+$ & 1.37E-01 $+$ & 1.63E-03 $+$ & 3.42E-03 $+$ & 5.19E-01 $+$ & 1.96E-03 $+$ & 3.63E-01 $+$ & 3.89E-01 $+$ & \textbf{8.32E-07} \\ 
			2 & Schwefel & 9.97E+01 $+$  & 8.37E-02 $+$ & 1.48E-01 $+$ & 2.36E+01 $+$ & 7.02E+01 $+$ & 2.15E+00 $+$ & 1.13E+02 $+$ & 2.84E+00 $+$ & 1.50E+02 $+$ & 1.16E+02 $+$ & \textbf{5.58E-05} \\ 
			4 & GPMean & 8.02E-01 $+$  & 7.64E-02 $=$ & \textbf{6.03E-02} $-$ & 2.36E-01 $=$ & 4.56E-01 $+$ & 1.86E-01 $=$ & 6.09E-01 $+$ & 1.23E-01 $=$ & 3.60E-01 $+$ & 2.44E-01 $=$ & 2.14E-01 \\ 
			4 & SumSquares & 5.54E+00 $+$  & 4.05E-02 $+$ & 3.12E-05 $+$ & 8.23E-05 $=$ & 9.17E-03 $+$ & 1.82E-03 $+$ & 5.79E-02 $+$ & 6.57E-05 $+$ & 3.69E-03 $+$ & 2.99E-04 $=$ & \textbf{2.10E-05 }\\ 
			4 & Rosenbrock & 6.50E+02 $+$  & 1.45E+01 $+$ & 1.90E+01 $+$ & 3.61E+00 $=$ & 7.56E+00 $=$ & 3.94E+01 $+$ & 2.32E+02 $+$ & \textbf{6.24E-01} $-$ & 1.46E+02 $+$ & 3.13E+00 $=$ & 4.11E+00 \\ 
			4 & DixonPrice & 1.94E+02 $+$  & 4.83E+00 $+$ & 5.45E+00 $+$ & 2.68E+00 $+$ & 3.51E-01 $-$ & 1.39E+01 $+$ & 1.32E+02 $+$ & \textbf{2.79E-01} $-$ & 7.75E+01 $+$ & 2.77E+00 $=$ & 9.30E-01 \\ 
			4 & Ackley & 1.37E+01 $+$  & 4.56E+00 $-$ & 3.52E+00 $-$ & 1.74E+01 $+$ & \textbf{1.22E+00} $-$ & 8.09E+00 $=$ & 1.84E+01 $+$ & 1.49E+01 $+$ & 1.64E+01 $+$ & 1.82E+01 $+$ & 9.41E+00 \\ 
			4 & Rastrigin & 2.04E+01 $+$  & 1.13E+01 $+$ & 1.08E+01 $+$ & 2.39E+01 $+$ & 9.36E+00 $+$ & 1.84E+01 $+$ & 2.73E+01 $+$ & 1.11E+01 $+$ & 2.46E+01 $+$ & 2.56E+01 $+$ & \textbf{3.50E+00}\\ 
			4 & Griewank & 8.26E+00 $+$  & 9.57E-01 $+$ & 6.59E-01 $=$ & 3.91E+00 $+$ & \textbf{5.26E-01} $-$ & 9.44E-01 $+$ & 2.82E+01 $+$ & 6.89E-01 $=$ & 1.32E+01 $+$ & 1.35E+01 $+$ & 6.58E-01 \\ 
			4 & Levy & 2.51E+00 $+$  & 1.15E+00 $+$ & 9.13E-01 $+$ & 2.56E+00 $+$ & 9.54E-01 $=$ & 1.08E+00 $+$ & 4.97E+00 $+$ & 5.00E-01 $+$ & 4.33E+00 $+$ & 3.24E+00 $+$ & \textbf{5.20E-02}\\ 
			4 & Michalewicz & 1.48E+00 $+$  & 5.61E-01 $+$ & 5.40E-01 $+$ & 1.71E+00 $+$ & 4.28E-01 $+$ & 5.97E-01 $+$ & 1.93E+00 $+$ & 1.17E+00 $+$ & 1.38E+00 $+$ & 1.88E+00 $+$ & \textbf{9.54E-02} \\ 
			4 & Schwefel & 5.86E+02 $+$  & 3.58E+02 $+$ & 2.96E+02 $+$ & 5.90E+02 $+$ & 4.30E+02 $+$ & 3.01E+02 $+$ & 6.89E+02 $+$ & 3.74E+02 $+$ & 6.94E+02 $+$ & 6.65E+02 $+$ & \textbf{7.18E+01} \\ 
			8 & GPMean & 2.10E+00 $+$  & 2.65E-01 $-$ & 1.90E-01 $-$ & 6.15E-01 $=$ & 4.31E-01 $=$ & 1.42E-01 $-$ & 6.50E-01 $=$ & \textbf{2.45E-03} $-$ & 2.05E-01 $-$ & 8.37E-01 $=$ & 6.12E-01 \\ 
			8 & SumSquares & 5.34E+01 $+$  & 9.27E-01 $+$ & 1.04E-03 $+$ & 4.58E-02 $+$ & 3.29E-02 $+$ & 7.69E-02 $+$ & 1.33E-03 $=$ & 5.96E-01 $+$ & 3.84E-01 $+$ & 3.91E-02 $+$ & \textbf{3.47E-04} \\ 
			8 & Rosenbrock & 1.48E+04 $+$  & 1.95E+03 $+$ & 1.94E+03 $+$ & 1.41E+03 $+$ & 1.18E+02 $=$ & 6.38E+02 $+$ & 9.72E+03 $+$ & 1.29E+03 $+$ & 2.13E+03 $+$ & 8.44E+02 $+$ & \textbf{8.95E+01} \\ 
			8 & DixonPrice & 8.32E+03 $+$  & 6.54E+02 $+$ & 5.87E+02 $+$ & 7.40E+02 $+$ & \textbf{9.12E+00} $-$ & 2.25E+02 $+$ & 4.57E+03 $+$ & 6.32E+02 $+$ & 1.38E+03 $+$ & 4.25E+02 $+$ & 4.64E+01 \\ 
			8 & Ackley & 1.79E+01 $+$  & 8.13E+00 $=$ & 3.75E+00 $-$ & 1.74E+01 $+$ & \textbf{1.39E+00} $-$ & 1.19E+01 $+$ & 1.90E+01 $+$ & 1.57E+01 $+$ & 1.44E+01 $+$ & 1.79E+01 $+$ & 6.87E+00 \\ 
			8 & Rastrigin & 6.68E+01 $+$  & 3.47E+01 $+$ & 3.72E+01 $+$ & 3.38E+01 $+$ & 2.11E+01 $+$ & 2.78E+01 $+$ & 4.79E+01 $+$ & 3.01E+01 $+$ & 4.12E+01 $+$ & 3.41E+01 $+$ & \textbf{1.40E+01} \\ 
			8 & Griewank & 5.09E+01 $+$  & 2.71E+00 $+$ & \textbf{2.94E-01} $-$ & 1.25E+00 $+$ & 9.27E-01 $+$ & 1.43E+00 $+$ & 8.53E-01 $+$ & 9.54E-01 $+$ & 1.44E+01 $+$ & 1.08E+01 $+$ & 6.16E-01 \\ 
			8 & Levy & 1.20E+01 $+$  & 2.61E+00 $+$ & 3.00E+00 $+$ & 4.44E+00 $+$ & 1.04E+00 $=$ & 3.17E+00 $+$ & 9.41E+00 $+$ & 3.43E+00 $+$ & 1.06E+01 $+$ & 7.61E+00 $+$ & \textbf{7.28E-01} \\ 
			8 & Michalewicz & 4.17E+00 $+$  & 3.18E+00 $+$ & 3.11E+00 $+$ & 3.66E+00 $+$ & 1.76E+00 $+$ & 2.66E+00 $+$ & 4.58E+00 $+$ & 3.04E+00 $+$ & 3.46E+00 $+$ & 3.76E+00 $+$ & \textbf{6.02E-01} \\ 
			8 & Schwefel & 1.70E+03 $+$  & 1.28E+03 $+$ & 1.27E+03 $+$ & 1.15E+03 $+$ & 8.77E+02 $+$ & 1.03E+03 $+$ & 1.46E+03 $+$ & 9.63E+02 $+$ & 1.33E+03 $+$ & 1.18E+03 $+$ & \textbf{4.33E+02 }\\ 
			10 & GPMean & 1.89E+00 $+$  & 1.30E+00 $+$ & 7.18E-01 $=$ & 6.04E-01 $=$ & 6.44E-01 $=$ & 3.09E-01 $-$ & 7.31E-01 $=$ & \textbf{2.85E-01} $-$ & 4.02E-01 $-$ & 7.12E-01 $=$ & 7.02E-01 \\ 
			10 & SumSquares & 1.09E+02 $+$  & 2.91E+00 $+$ & 1.22E-02 $+$ & 2.12E-01 $+$ & 8.33E-02 $+$ & 2.98E-01 $+$ & 2.41E-03 $+$ & 2.55E-01 $+$ & 1.15E+00 $+$ & 2.41E-01 $+$ & \textbf{1.84E-03} \\ 
			10 & Rosenbrock & 3.74E+04 $+$  & 4.64E+03 $+$ & 3.67E+03 $+$ & 3.80E+03 $+$ & 7.48E+02 $+$ & 1.87E+03 $+$ & 1.38E+04 $+$ & 2.90E+03 $+$ & 3.70E+03 $+$ & 2.24E+03 $+$ & \textbf{4.14E+02 }\\ 
			10 & DixonPrice & 2.78E+04 $+$  & 1.08E+03 $+$ & 1.34E+03 $+$ & 1.70E+03 $+$ & 1.37E+02 $=$ & 5.54E+02 $+$ & 4.44E+03 $+$ & 9.75E+02 $+$ & 1.72E+03 $+$ & 1.09E+03 $+$ & \textbf{1.23E+02} \\ 
			10 & Ackley & 1.86E+01 $+$  & 1.09E+01 $=$ & 4.24E+00 $-$ & 1.50E+01 $+$ & \textbf{1.53E+00} $-$ & 1.28E+01 $+$ & 1.34E+01 $+$ & 1.04E+01 $=$ & 1.00E+01 $=$ & 1.69E+01 $+$ & 8.90E+00 \\ 
			10 & Rastrigin & 9.42E+01 $+$  & 5.74E+01 $+$ & 5.37E+01 $+$ & 3.73E+01 $+$ & 2.64E+01 $=$ & 3.70E+01 $+$ & 5.50E+01 $+$ & \textbf{2.02E+01} $=$ & 4.55E+01 $+$ & 4.23E+01 $+$ & 2.26E+01 \\ 
			10 & Griewank & 8.12E+01 $+$  & 3.79E+00 $+$ & 9.79E-01 $+$ & 1.49E+00 $+$ & 1.03E+00 $+$ & 1.53E+00 $+$ & 8.21E-01 $=$ & 1.10E+00 $+$ & 1.83E+01 $+$ & 1.91E+01 $+$ & \textbf{7.63E-01} \\ 
			10 & Levy & 2.18E+01 $+$  & 4.07E+00 $+$ & 3.42E+00 $+$ & 5.10E+00 $+$ & 2.17E+00 $+$ & 6.34E+00 $+$ & 9.80E+00 $+$ & 2.93E+00 $+$ & 8.93E+00 $+$ & 7.72E+00 $+$ & \textbf{1.52E+00} \\ 
			10 & Michalewicz & 5.80E+00 $+$  & 4.75E+00 $+$ & 4.88E+00 $+$ & 4.89E+00 $+$ & 3.22E+00 $+$ & 4.27E+00 $+$ & 5.70E+00 $+$ & 4.59E+00 $+$ & 4.59E+00 $+$ & 5.15E+00 $+$ & \textbf{1.31E+00} \\ 
			10 & Schwefel & 2.36E+03 $+$  & 1.98E+03 $+$ & 1.92E+03 $+$ & 1.49E+03 $+$ & 1.16E+03 $+$ & 1.55E+03 $+$ & 2.23E+03 $+$ & 1.52E+03 $+$ & 1.86E+03 $+$ & 1.51E+03 $+$ & \textbf{7.01E+02} \\ 
			20 & GPMean & 3.06E+00 $+$  & 3.06E+00 $+$ & 3.03E+00 $+$ & 2.98E+00 $=$ & 1.69E+00 $-$ & \textbf{1.38E+00} $-$ & 2.92E+00 $=$ & 3.06E+00 $+$ & 2.93E+00 $=$ & 2.99E+00 $=$ & 2.49E+00 \\ 
			20 & SumSquares & 6.79E+02 $+$  & 9.19E+01 $+$ & 2.18E-01 $+$ & 1.05E+01 $+$ & 6.49E-01 $+$ & 6.02E+00 $+$ & 3.41E-02 $=$ & 8.69E+00 $+$ & 8.46E+00 $+$ & 9.68E+00 $+$ & \textbf{1.86E-02} \\ 
			20 & Rosenbrock & 2.85E+05 $+$  & 1.41E+04 $+$ & 1.39E+04 $+$ & 9.92E+03 $+$ & 2.92E+03 $+$ & 1.05E+04 $+$ & 4.26E+04 $+$ & 8.41E+03 $+$ & 8.65E+03 $+$ & 8.17E+03 $+$ & \textbf{1.04E+03} \\ 
			20 & DixonPrice & 2.77E+05 $+$  & 3.73E+03 $+$ & 3.73E+03 $+$ & 8.39E+03 $+$ & 7.08E+02 $+$ & 2.89E+03 $+$ & 3.00E+04 $+$ & 4.34E+03 $+$ & 6.62E+03 $+$ & 6.41E+03 $+$ & \textbf{4.87E+02}\\ 
			20 & Ackley & 1.99E+01 $+$  & 1.63E+01 $+$ & 4.68E+00 $+$ & 1.77E+01 $+$ & \textbf{2.13E+00} $-$ & 1.77E+01 $+$ & 1.80E+01 $+$ & 1.28E+01 $+$ & 9.97E+00 $+$ & 1.83E+01 $+$ & 2.72E+00 \\ 
			20 & Rastrigin & 2.30E+02 $+$  & 1.36E+02 $+$ & 1.46E+02 $+$ & \textbf{4.59E+01} $-$ & 6.59E+01 $=$ & 1.09E+02 $+$ & 8.66E+01 $+$ & 1.81E+02 $+$ & 6.57E+01 $=$ & 8.37E+01 $=$ & 6.50E+01 \\ 
			20 & Griewank & 2.38E+02 $+$  & 3.79E+01 $+$ & 1.14E+00 $+$ & 1.30E+00 $+$ & 1.18E+00 $+$ & 1.35E+00 $+$ & \textbf{9.20E-01} $-$ & 1.53E+00 $+$ & 1.76E+00 $+$ & 1.74E+00 $+$ & 9.62E-01 \\ 
			20 & Levy & 7.63E+01 $+$  & 7.68E+00 $+$ & 8.68E+00 $+$ & 8.28E+00 $+$ & 5.99E+00 $+$ & 1.06E+01 $+$ & 9.79E+00 $+$ & \textbf{3.39E+00} $=$ & 1.36E+01 $+$ & 1.19E+01 $+$ & 3.56E+00 \\ 
			20 & Michalewicz & 1.32E+01 $+$  & 1.23E+01 $+$ & 1.22E+01 $+$ & 1.35E+01 $+$ & 7.78E+00 $+$ & 1.20E+01 $+$ & 1.35E+01 $+$ & 1.31E+01 $+$ & 1.31E+01 $+$ & 1.34E+01 $+$ & \textbf{5.90E+00} \\ 
			20 & Schwefel & 5.46E+03 $+$  & 4.83E+03 $+$ & 4.80E+03 $+$ & 3.54E+03 $+$ & 2.48E+03 $+$ & 4.28E+03 $+$ & 4.71E+03 $+$ & 5.47E+03 $+$ & 4.00E+03 $+$ & 5.09E+03 $+$ & \textbf{1.95E+03} \\ 
			30 & GPMean & 2.41E+00 $+$  & 2.41E+00 $+$ & 2.41E+00 $+$ & 2.41E+00 $+$ & \textbf{2.19E+00} $-$ & 2.34E+00 $-$ & 2.41E+00 $+$ & 2.37E+00 $-$ & 2.41E+00 $+$ & 2.41E+00 $+$ & 2.40E+00 \\ 
			30 & SumSquares & 1.86E+03 $+$  & 4.68E+02 $+$ & 4.55E-01 $+$ & 3.56E+01 $+$ & 6.61E+00 $+$ & 5.62E+01 $+$ & \textbf{1.60E-01} $=$ & 3.32E+01 $+$ & 2.63E+01 $+$ & 2.55E+01 $+$ & 2.23E-01 \\ 
			30 & Rosenbrock & 7.86E+05 $+$  & 2.58E+04 $+$ & 2.43E+04 $+$ & 1.85E+04 $+$ & 7.95E+03 $+$ & 3.76E+04 $+$ & 1.02E+05 $+$ & 1.56E+04 $+$ & 1.79E+04 $+$ & 1.96E+04 $+$ & \textbf{3.27E+03} \\ 
			30 & DixonPrice & 1.01E+06 $+$  & 9.41E+03 $+$ & 1.01E+04 $+$ & 2.14E+04 $+$ & 2.48E+03 $=$ & 4.04E+04 $+$ & 1.01E+05 $+$ & 1.70E+04 $+$ & 1.62E+04 $+$ & 1.74E+04 $+$ & \textbf{1.88E+03} \\ 
			30 & Ackley & 2.03E+01 $+$  & 1.81E+01 $+$ & 6.22E+00 $+$ & 1.69E+01 $+$ & \textbf{3.31E+00} $-$ & 1.85E+01 $+$ & 1.59E+01 $+$ & 8.37E+00 $+$ & 8.78E+00 $+$ & 1.83E+01 $+$ & 3.78E+00 \\ 
			30 & Rastrigin & 3.79E+02 $+$  & 2.44E+02 $+$ & 2.51E+02 $+$ & 1.01E+02 $-$ & \textbf{9.92E+01} $-$ & 2.29E+02 $+$ & 1.96E+02 $+$ & 2.78E+02 $+$ & 1.18E+02 $-$ & 1.24E+02 $-$ & 1.60E+02 \\ 
			30 & Griewank & 4.56E+02 $+$  & 1.23E+02 $+$ & 1.17E+00 $+$ & 1.77E+00 $+$ & 2.50E+00 $+$ & 1.35E+01 $+$ & \textbf{1.05E+00} $=$ & 2.11E+00 $+$ & 1.84E+00 $+$ & 2.02E+00 $+$ & 1.08E+00 \\ 
			30 & Levy & 1.50E+02 $+$  & 1.07E+01 $+$ & 1.37E+01 $+$ & 1.43E+01 $+$ & 1.03E+01 $+$ & 5.45E+01 $+$ & 2.51E+01 $+$ & 6.89E+00 $+$ & 1.39E+01 $+$ & 1.16E+01 $+$ & \textbf{4.63E+00 }\\ 
			30 & Michalewicz & 2.15E+01 $+$  & 2.04E+01 $+$ & 2.03E+01 $+$ & 2.15E+01 $+$ & \textbf{1.36E+01} $=$ & 2.06E+01 $+$ & 2.13E+01 $+$ & 2.15E+01 $+$ & 2.15E+01 $+$ & 2.15E+01 $+$ & 1.37E+01 \\ 
			30 & Schwefel & 9.26E+03 $+$  & 8.13E+03 $+$ & 8.21E+03 $+$ & 6.91E+03 $+$ & \textbf{3.62E+03} $-$ & 8.32E+03 $+$ & 7.09E+03 $+$ & 9.01E+03 $+$ & 6.62E+03 $+$ & 8.39E+03 $+$ & 4.03E+03 \\ 
			\midrule
			\multicolumn{2}{c}{$+$/$\approx$/$-$}  &   60/0/0  &  52/5/3 &   48/6/6  &  49/6/5   &  34/12/14 &    51/3/6  &  46/10/4  &  47/5/8    &52/3/5   & 47/9/4 &  N.A. \\
			\bottomrule
		\end{tabular}
	}
\end{table*}

\begin{table*}
	\caption{Simple regrets obtained by different batch approaches when $q=128$ are used}
	\label{table_batch_q128}
	\centering
	\renewcommand{\arraystretch}{1.3}
	\resizebox{1.0\textwidth}{!}{
		\begin{tabular}{c c c c c c c c c c c c c} 
			\toprule
			$d$  &   $f$   & RS & $q$EI & $q$logEI & F$q$EI & TuRBO & MSMR & MACE   & PEI & EIMI & KB & SSEI \\
			\midrule
			2 & GPMean & 9.42E-02 $+$  & 3.18E-04 $=$ & 2.31E-04 $=$ & 7.45E-03 $+$ & 8.79E-02 $+$ & 1.65E-03 $+$ & 4.66E-01 $+$ & \textbf{2.28E-04} $=$ & 1.84E-02 $+$ & 2.48E-03 $+$ & 2.72E-04 \\ 
			2 & SumSquares & 1.91E-01 $+$  & \textbf{2.31E-06} $-$ & 3.24E-06 $-$ & 3.08E-04 $+$ & 7.36E-02 $+$ & 1.76E-04 $+$ & 9.43E-02 $+$ & 9.06E-05 $=$ & 1.64E-02 $+$ & 7.45E-02 $+$ & 8.63E-06 \\ 
			2 & Rosenbrock & 2.14E+00 $+$  & 1.56E-01 $=$ & 1.49E-01 $=$ & 4.08E+00 $=$ & 1.15E+00 $+$ & 2.13E+00 $+$ & 1.88E+02 $+$ & \textbf{1.47E-01} $-$ & 1.90E+02 $+$ & 4.63E+01 $+$ & 2.49E-01 \\ 
			2 & DixonPrice & 1.02E+00 $+$  & \textbf{5.29E-02 }$=$ & 5.85E-02 $=$ & 6.22E-01 $=$ & 3.28E-01 $+$ & 6.48E-01 $+$ & 7.99E+01 $+$ & 1.61E-01 $+$ & 5.00E+01 $+$ & 4.49E+00 $+$ & 7.82E-02 \\ 
			2 & Ackley & 6.26E+00 $+$  & 3.76E+00 $=$ & 3.88E+00 $=$ & 1.02E+01 $+$ & 4.70E+00 $=$ & 7.65E+00 $+$ & 1.63E+01 $+$ & \textbf{2.93E+00} $-$ & 8.51E+00 $+$ & 3.37E+00 $-$ & 4.84E+00 \\ 
			2 & Rastrigin & 3.05E+00 $+$  & 3.11E+00 $+$ & 2.21E+00 $+$ & 6.29E+00 $+$ & 2.65E+00 $+$ & 4.64E+00 $+$ & 1.12E+01 $+$ & 4.36E+00 $+$ & 1.11E+01 $+$ & 3.39E+00 $+$ & \textbf{1.09E+00} \\ 
			2 & Griewank & 1.40E+00 $+$  & 7.97E-01 $=$ & 8.58E-01 $=$ & 2.72E+00 $=$ & 9.03E-01 $=$ & 1.01E+00 $+$ & 1.52E+01 $+$ & \textbf{5.10E-01}$=$ & 1.39E+01 $+$ & 2.07E+00 $+$ & 6.88E-01 \\ 
			2 & Levy & 1.40E-01 $+$  & 1.10E-01 $+$ & 1.15E-01 $+$ & 6.48E-01 $+$ & \textbf{3.94E-02} $=$ & 1.18E-01 $+$ & 1.94E+00 $+$ & 1.84E-01 $+$ & 1.80E+00 $+$ & 2.61E-01 $+$ & 3.98E-02 \\ 
			2 & Michalewicz & 2.86E-01 $+$  & 3.43E-02 $+$ & 3.39E-02 $+$ & 6.59E-01 $+$ & 1.12E-01 $+$ & 3.97E-01 $+$ & 8.93E-01 $+$ & 4.44E-02 $+$ & 3.17E-01 $+$ & 1.11E-01 $+$ & \textbf{1.18E-02} \\ 
			2 & Schwefel & 8.80E+01 $+$  & 3.29E+01 $+$ & 3.66E+01 $=$ & 2.27E+02 $+$ & 1.16E+02 $+$ & 1.29E+02 $+$ & 3.13E+02 $+$ & 1.31E+02 $+$ & 3.17E+02 $+$ & 1.41E+02 $+$ & \textbf{1.54E+01} \\ 
			4 & GPMean & 8.46E-01 $+$  & \textbf{4.22E-01} $-$ & 5.07E-01 $=$ & 1.24E+00 $+$ & 7.53E-01 $+$ & 6.51E-01 $=$ & 1.61E+00 $+$ & 5.32E-01 $=$ & 8.77E-01 $+$ & 9.41E-01 $+$ & 5.60E-01 \\ 
			4 & SumSquares & 5.20E+00 $+$  & \textbf{6.22E-04} $-$ & 7.82E-04 $-$ & 5.59E-03 $+$ & 2.13E+00 $+$ & 7.42E-03 $+$ & 1.05E-01 $+$ & 7.44E-03 $+$ & 2.55E-01 $=$ & 4.39E-02 $+$ & 4.90E-03 \\ 
			4 & Rosenbrock & 8.24E+02 $+$  & 6.92E+02 $+$ & 5.82E+02 $+$ & 3.08E+03 $+$ & \textbf{1.64E+02} $=$ & 7.87E+02 $+$ & 1.13E+04 $+$ & 6.72E+02 $=$ & 1.13E+04 $+$ & 4.32E+03 $+$ & 2.83E+02 \\ 
			4 & DixonPrice & 2.14E+02 $+$  & 1.72E+02 $+$ & 1.39E+02 $=$ & 5.53E+02 $+$ & \textbf{4.18E+01} $=$ & 2.77E+02 $+$ & 5.90E+03 $+$ & 8.06E+01 $=$ & 3.40E+03 $+$ & 3.96E+02 $+$ & 5.50E+01 \\
			4 & Ackley & 1.31E+01 $+$  & \textbf{9.55E+00} $=$ & 9.94E+00 $=$ & 1.64E+01 $+$ & 1.04E+01 $=$ & 1.59E+01 $+$ & 1.82E+01 $+$ & 1.50E+01 $+$ & 1.70E+01 $+$ & 1.60E+01 $+$ & 1.06E+01 \\ 
			4 & Rastrigin & 2.19E+01 $+$  & 1.66E+01 $+$ & 1.62E+01 $+$ & 2.74E+01 $+$ & 1.78E+01 $+$ & 2.17E+01 $+$ & 3.26E+01 $+$ & 2.24E+01 $+$ & 2.97E+01 $+$ & 2.44E+01 $+$ & \textbf{1.04E+01 }\\ 
			4 & Griewank & 9.57E+00 $+$  & 9.71E-01 $-$ & \textbf{9.61E-01} $-$ & 2.22E+01 $+$ & 4.03E+00 $+$ & 1.34E+01 $+$ & 3.96E+01 $+$ & 1.40E+00 $=$ & 3.39E+01 $+$ & 1.97E+01 $+$ & 1.36E+00 \\ 
			4 & Levy & 2.14E+00 $+$  & 1.95E+00 $+$ & 2.15E+00 $+$ & 4.17E+00 $+$ & 9.54E-01 $+$ & 2.12E+00 $+$ & 6.39E+00 $+$ & 2.20E+00 $+$ & 5.61E+00 $+$ & 2.88E+00 $+$ & \textbf{6.55E-01} \\ 
			4 & Michalewicz & 1.46E+00 $+$  & 1.27E+00 $+$ & 1.26E+00 $+$ & 1.85E+00 $+$ & 1.26E+00 $+$ & 1.60E+00 $+$ & 2.23E+00 $+$ & 1.40E+00 $+$ & 1.62E+00 $+$ & 1.61E+00 $+$ & \textbf{8.03E-01} \\ 
			4 & Schwefel & 6.02E+02 $+$  & 6.00E+02 $+$ & 5.53E+02 $+$ & 7.54E+02 $+$ & 5.90E+02 $+$ & 4.74E+02 $+$ & 8.25E+02 $+$ & 5.44E+02 $+$ & 8.83E+02 $+$ & 6.47E+02 $+$ & \textbf{4.00E+02} \\ 
			8 & GPMean & 2.06E+00 $+$  & \textbf{3.46E-01} $-$ & 4.66E-01 $-$ & 1.21E+00 $=$ & 1.08E+00 $=$ & 6.42E-01 $-$ & 1.85E+00 $+$ & 5.08E-01 $-$ & 1.16E+00 $=$ & 1.26E+00 $=$ & 1.14E+00 \\ 
			8 & SumSquares & 5.59E+01 $+$  & 3.37E-01 $+$ & 6.05E-03 $-$ & 1.90E-01 $=$ & 1.47E+01 $+$ & \textbf{4.20E-03} $-$ & 9.36E-01 $=$ & 2.84E-01 $+$ & 4.02E-01 $=$ & 6.35E-02 $-$ & 1.26E-01 \\ 
			8 & Rosenbrock & 1.30E+04 $+$  & 6.76E+03 $+$ & 7.36E+03 $+$ & 8.60E+03 $+$ & \textbf{1.51E+03} $=$ & 2.93E+03 $+$ & 3.51E+04 $+$ & 5.37E+03 $+$ & 7.31E+03 $+$ & 5.75E+03 $+$ & 1.80E+03 \\ 
			8 & DixonPrice & 8.85E+03 $+$  & 1.93E+03 $+$ & 2.43E+03 $+$ & 3.99E+03 $=$ & \textbf{5.96E+02} $=$ & 7.19E+02 $=$ & 1.60E+04 $+$ & 9.89E+02 $=$ & 2.67E+03 $+$ & 1.78E+03 $+$ & 7.07E+02 \\ 
			8 & Ackley & 1.80E+01 $+$  & \textbf{9.65E+00} $-$ & 1.06E+01 $-$ & 1.80E+01 $+$ & 1.19E+01 $-$ & 1.75E+01 $+$ & 1.69E+01 $+$ & 1.80E+01 $+$ & 1.77E+01 $+$ & 1.82E+01 $+$ & 1.58E+01 \\ 
			8 & Rastrigin & 6.44E+01 $+$  & 4.62E+01 $+$ & 4.76E+01 $+$ & 5.35E+01 $+$ & 4.48E+01 $+$ & 4.33E+01 $+$ & 6.83E+01 $+$ & 4.55E+01 $+$ & 5.23E+01 $+$ & 4.48E+01 $+$ & \textbf{3.12E+01} \\ 
			8 & Griewank & 4.78E+01 $+$  & 1.45E+00 $+$ & \textbf{1.01E+00} $-$ & 1.22E+01 $=$ & 1.46E+01 $+$ & 6.82E+00 $+$ & 4.12E+01 $+$ & 1.42E+00 $=$ & 1.47E+01 $=$ & 1.66E+01 $=$ & 1.08E+00 \\ 
			8 & Levy & 1.27E+01 $+$  & 6.41E+00 $=$ & 5.63E+00 $=$ & 1.09E+01 $+$ & \textbf{3.96E+00} $=$ & 5.83E+00 $=$ & 1.18E+01 $+$ & 8.40E+00 $+$ & 1.29E+01 $+$ & 1.02E+01 $+$ & 5.14E+00 \\ 
			8 & Michalewicz & 4.20E+00 $+$  & 3.81E+00 $+$ & 3.84E+00 $+$ & 3.88E+00 $+$ & 3.83E+00 $+$ & 3.79E+00 $+$ & 4.68E+00 $+$ & 3.36E+00 $+$ & 3.78E+00 $+$ & 3.76E+00 $+$ & \textbf{2.41E+00} \\ 
			8 & Schwefel & 1.72E+03 $+$  & 1.70E+03 $+$ & 1.67E+03 $+$ & 1.37E+03 $+$ & 1.53E+03 $+$ & 1.22E+03 $+$ & 1.84E+03 $+$ & 1.21E+03 $+$ & 1.57E+03 $+$ & 1.30E+03 $+$ & \textbf{9.55E+02} \\ 
			10 & GPMean & 1.82E+00 $+$  & 1.35E+00 $+$ & 1.30E+00 $+$ & 1.65E+00 $+$ & 1.43E+00 $+$ & \textbf{6.50E-01} $-$ & 1.47E+00 $+$ & 1.10E+00 $=$ & 1.47E+00 $+$ & 1.15E+00 $=$ & 9.36E-01 \\ 
			10 & SumSquares & 9.70E+01 $+$  & 1.21E+00 $+$ & 9.23E-02 $-$ & 1.87E-01 $-$ & 2.90E+01 $+$ & 1.14E-02 $-$ & 3.88E-01 $-$ & \textbf{1.26E-03} $-$ & 6.97E-01 $=$ & 3.39E-01 $-$ & 4.32E-01 \\ 
			10 & Rosenbrock & 3.83E+04 $+$  & 7.73E+03 $+$ & 7.83E+03 $+$ & 1.47E+04 $+$ & 5.08E+03 $+$ & 3.57E+03 $=$ & 3.18E+04 $+$ & 5.47E+03 $+$ & 1.18E+04 $+$ & 5.40E+03 $+$ & \textbf{3.12E+03} \\ 
			10 & DixonPrice & 2.91E+04 $+$  & 2.24E+03 $+$ & 2.43E+03 $+$ & 5.87E+03 $=$ & 1.81E+03 $=$ & 1.64E+03 $=$ & 7.88E+03 $+$ & \textbf{8.21E+02} $-$ & 4.69E+03 $=$ & 2.06E+03 $=$ & 1.12E+03 \\ 
			10 & Ackley & 1.87E+01 $+$  & \textbf{1.09E+01} $-$ & 1.17E+01 $-$ & 1.51E+01 $=$ & 1.35E+01 $-$ & 1.76E+01 $+$ & 1.59E+01 $=$ & 1.36E+01 $-$ & 1.34E+01 $-$ & 1.68E+01 $=$ & 1.65E+01 \\ 
			10 & Rastrigin & 9.14E+01 $+$  & 6.52E+01 $+$ & 6.72E+01 $+$ & 6.04E+01 $+$ & 6.75E+01 $+$ & \textbf{4.56E+01} $=$ & 7.15E+01 $+$ & 7.70E+01 $+$ & 8.04E+01 $+$ & 7.84E+01 $+$ & 5.04E+01 \\ 
			10 & Griewank & 8.25E+01 $+$  & 1.45E+00 $+$ & \textbf{1.12E+00} $-$ & 2.90E+01 $+$ & 2.44E+01 $+$ & 2.17E+00 $=$ & 1.91E+01 $+$ & 2.92E+00 $=$ & 3.89E+01 $+$ & 4.07E+01 $+$ & 1.28E+00 \\ 
			10 & Levy & 2.12E+01 $+$  & 8.09E+00 $=$ & 7.79E+00 $=$ & 7.20E+00 $=$ & 6.96E+00 $=$ & 1.04E+01 $+$ & 1.17E+01 $=$ & \textbf{4.23E+00} $-$ & 1.34E+01 $=$ & 1.24E+01 $=$ & 7.02E+00 \\ 
			10 & Michalewicz & 5.85E+00 $+$  & 5.45E+00 $+$ & 5.39E+00 $+$ & 5.38E+00 $+$ & 5.35E+00 $+$ & 5.19E+00 $+$ & 5.82E+00 $+$ & 5.29E+00 $+$ & 5.24E+00 $+$ & 5.18E+00 $+$ & \textbf{3.33E+00} \\ 
			10 & Schwefel & 2.28E+03 $+$  & 2.35E+03 $+$ & 2.38E+03 $+$ & 2.20E+03 $+$ & 2.22E+03 $+$ & 1.83E+03 $+$ & 2.42E+03 $+$ & 2.04E+03 $+$ & 2.27E+03 $+$ & 2.07E+03 $+$ & \textbf{1.49E+03} \\ 
			20 & GPMean & 3.06E+00 $+$  & 3.06E+00 $+$ & 3.05E+00 $+$ & 2.99E+00 $+$ & 2.68E+00 $=$ & \textbf{1.98E+00} $-$ & 2.75E+00 $=$ & 3.06E+00 $+$ & 3.06E+00 $+$ & 3.03E+00 $+$ & 2.81E+00 \\ 
			20 & SumSquares & 6.66E+02 $+$  & 2.51E+01 $+$ & 7.16E-01 $=$ & 4.96E+00 $+$ & 1.10E+02 $+$ & 2.40E+00 $+$ & \textbf{2.59E-01} $-$ & 1.22E+00 $=$ & 2.08E+00 $+$ & 4.65E+00 $+$ & 8.52E-01 \\ 
			20 & Rosenbrock & 2.66E+05 $+$  & 1.04E+04 $+$ & 1.10E+04 $+$ & 2.85E+04 $+$ & 3.30E+04 $+$ & 4.45E+03 $+$ & 6.18E+04 $+$ & 1.45E+04 $+$ & 2.05E+04 $+$ & 1.01E+04 $+$ & \textbf{3.37E+03} \\ 
			20 & DixonPrice & 2.81E+05 $+$  & 7.26E+03 $+$ & 8.16E+03 $+$ & 1.49E+04 $+$ & 1.63E+04 $+$ & \textbf{1.82E+03} $=$ & 6.09E+04 $+$ & 7.09E+03 $+$ & 1.97E+04 $+$ & 8.72E+03 $+$ & 2.21E+03 \\ 
			20 & Ackley & 1.99E+01 $+$  & 1.59E+01 $+$ & \textbf{9.65E+00} $-$ & 1.63E+01 $+$ & 1.39E+01 $=$ & 1.68E+01 $+$ & 1.66E+01 $+$ & 1.32E+01 $=$ & 1.38E+01 $=$ & 1.68E+01 $+$ & 1.31E+01 \\ 
			20 & Rastrigin & 2.30E+02 $+$  & 1.47E+02 $+$ & 1.56E+02 $+$ & \textbf{8.22E+01} $=$ & 1.50E+02 $+$ & 1.22E+02 $+$ & 1.61E+02 $+$ & 1.70E+02 $+$ & 1.34E+02 $+$ & 1.01E+02 $=$ & 9.39E+01 \\ 
			20 & Griewank & 2.53E+02 $+$  & 3.17E+00 $+$ & 1.19E+00 $-$ & 1.21E+00 $-$ & 4.61E+01 $+$ & \textbf{1.12E+00} $-$ & 1.17E+00 $-$ & 1.18E+00 $-$ & 1.66E+00 $=$ & 3.97E+00 $+$ & 1.50E+00 \\ 
			20 & Levy & 7.29E+01 $+$  & 2.06E+01 $+$ & 1.72E+01 $+$ & 1.76E+01 $=$ & 1.95E+01 $+$ & 1.22E+01 $=$ & 2.70E+01 $+$ & 1.92E+01 $+$ & 1.88E+01 $+$ & 1.62E+01 $=$ & \textbf{1.17E+01} \\ 
			20 & Michalewicz & 1.34E+01 $+$  & 1.25E+01 $+$ & 1.28E+01 $+$ & 1.33E+01 $+$ & 1.17E+01 $+$ & 1.22E+01 $+$ & 1.36E+01 $+$ & 1.34E+01 $+$ & 1.33E+01 $+$ & 1.34E+01 $+$ & \textbf{7.74E+00} \\ 
			20 & Schwefel & 5.57E+03 $+$  & 5.48E+03 $+$ & 5.55E+03 $+$ & 5.06E+03 $+$ & 4.24E+03 $+$ & 5.35E+03 $+$ & 5.58E+03 $+$ & 5.52E+03 $+$ & 5.24E+03 $+$ & 5.16E+03 $+$ & \textbf{3.06E+03 }\\ 
			30 & GPMean & 2.41E+00 $+$  & 2.41E+00 $+$ & 2.41E+00 $+$ & 2.35E+00 $=$ & 2.41E+00 $+$ & 2.41E+00 $+$ & 2.22E+00 $-$ & 2.36E+00 $-$ & 2.41E+00 $+$ & \textbf{2.22E+00} $=$ & 2.41E+00 \\ 
			30 & SumSquares & 1.90E+03 $+$  & 1.03E+02 $+$ & \textbf{1.44E+00} $-$ & 4.17E+01 $+$ & 3.47E+02 $+$ & 2.67E+01 $+$ & 4.48E+00 $-$ & 1.72E+01 $+$ & 1.81E+01 $+$ & 2.82E+01 $+$ & 9.00E+00 \\ 
			30 & Rosenbrock & 8.50E+05 $+$  & 2.66E+04 $+$ & 2.49E+04 $+$ & 3.87E+04 $+$ & 9.59E+04 $+$ &\textbf{ 7.45E+03} $-$ & 9.23E+04 $+$ & 2.54E+04 $+$ & 3.02E+04 $+$ & 2.14E+04 $+$ & 1.08E+04 \\ 
			30 & DixonPrice & 1.01E+06 $+$  & 2.08E+04 $+$ & 2.23E+04 $+$ & 3.20E+04 $+$ & 1.13E+05 $+$ & \textbf{1.65E+03} $-$ & 1.21E+05 $+$ & 1.74E+04 $+$ & 2.84E+04 $+$ & 2.15E+04 $+$ & 6.36E+03 \\ 
			30 & Ackley & 2.03E+01 $+$  & 1.79E+01 $+$ & 1.21E+01 $=$ & 1.76E+01 $+$ & 1.58E+01 $+$ & 1.63E+01 $+$ & 1.80E+01 $+$ & \textbf{1.07E+01} $=$ & 1.51E+01 $+$ & 1.75E+01 $+$ & 1.22E+01 \\ 
			30 & Rastrigin & 3.77E+02 $+$  & 2.45E+02 $+$ & 2.69E+02 $+$ & \textbf{1.13E+02} $-$ & 2.40E+02 $+$ & 2.49E+02 $+$ & 2.78E+02 $+$ & 2.65E+02 $+$ & 1.98E+02 $=$ & 1.82E+02 $=$ & 1.91E+02 \\ 
			30 & Griewank & 4.58E+02 $+$  & 2.10E+01 $+$ & \textbf{1.22E+00} $-$ & 1.69E+00 $-$ & 1.08E+02 $+$ & 4.38E+00 $=$ & 1.32E+00 $-$ & 1.52E+00 $-$ & 1.90E+00 $-$ & 2.40E+00 $-$ & 3.87E+00 \\ 
			30 & Levy & 1.48E+02 $+$  & 3.34E+01 $+$ & 3.34E+01 $+$ & 1.82E+01 $-$ & 4.99E+01 $+$ & \textbf{9.92E+00} $-$ & 4.12E+01 $+$ & 2.89E+01 $+$ & 3.70E+01 $+$ & 3.02E+01 $=$ & 2.39E+01 \\ 
			30 & Michalewicz & 2.14E+01 $+$  & 2.06E+01 $+$ & 2.07E+01 $+$ & 2.10E+01 $+$ & 1.78E+01 $+$ & 2.12E+01 $+$ & 2.12E+01 $+$ & 2.14E+01 $+$ & 2.11E+01 $+$ & 2.09E+01 $+$ & \textbf{1.58E+01} \\ 
			30 & Schwefel & 9.15E+03 $+$  & 9.04E+03 $+$ & 9.09E+03 $+$ & 7.35E+03 $+$ & 6.89E+03 $+$ & 8.97E+03 $+$ & 8.51E+03 $+$ & 9.16E+03 $+$ & 8.19E+03 $+$ & 8.18E+03 $+$ & \textbf{5.41E+03} \\ 
			\midrule
			\multicolumn{2}{c}{$+$/$\approx$/$-$}  &  60/0/0  &    45/8/7   &   33/13/14  &    42/13/5 &    44/14/2 &    41/10/9  &    50/4/6   &   36/14/10  &    48/10/2  &    45/11/4 &    N.A. \\
			\bottomrule
		\end{tabular}
	}
\end{table*}

%
%
%
%
%

\end{document}